\documentclass[journal]{IEEEtran}
\ifCLASSINFOpdf
\else
\fi
\usepackage[stretch=10]{microtype}
\usepackage[utf8]{inputenc}
\usepackage[T1]{fontenc}
\PassOptionsToPackage{hyphens}{url}
\usepackage{times,multirow,float}
\usepackage{graphicx}
\usepackage{booktabs}
\usepackage{epsfig,xspace,layout}
\usepackage{color}
\usepackage{amsfonts}
\usepackage{times}
\usepackage{amssymb}
\usepackage{amsthm}
\usepackage{hyperref}
\usepackage{amsmath,bm}
\usepackage{rotating}
\usepackage{mathrsfs}
\usepackage{mathtools}
\usepackage{makeidx}
\usepackage{stmaryrd}
\usepackage{scalerel}
\newcommand{\bigllbracket}{\mathopen{\scalerel*{\llbracket}{\sum}}}
\newcommand{\bigrrbracket}{\mathclose{\scalerel*{\rrbracket}{\sum}}}
\usepackage[]{threeparttable}
\usepackage{dsfont}
\usepackage{cite}
\usepackage{cuted}
\usepackage{flushend}
\AfterEndEnvironment{strip}{\leavevmode}
\usepackage{xcolor}
\usepackage{cleveref}
\usepackage[ruled, lined, linesnumbered, commentsnumbered, longend]{algorithm2e}
\usepackage{tikz}
\usetikzlibrary{shapes,arrows,positioning,calc}
\usepackage{siunitx}
\usepackage{multicol,lipsum}
\usepackage{subcaption}
\usepackage[font=footnotesize]{caption}
\usepackage{booktabs}
\usepackage{lineno}
\usepackage{hhline}
\usepackage[Symbol]{upgreek}
\usepackage{float}
\newtheorem{remark}{Remark}
\newtheorem{lemma}{Lemma}
\newtheorem{theorem}{Theorem}
\SetKwComment{Comment}{$\triangleright$\ }{}
\SetKwInput{Return}{Return}
\makeatletter
\renewcommand{\Indentp}[1]{%
  \advance\leftskip by #1
  \advance\skiptext by -#1
  \advance\skiprule by #1}%
\renewcommand{\Indp}{\algocf@adjustskipindent\Indentp{\algoskipindent}}
\renewcommand{\Indm}{\algocf@adjustskipindent\Indentp{-\algoskipindent}}
\makeatother

\begin{document}
% \captionsetup[table]{name=TABLE,labelsep=newline,textfont=sc}
\title{DiffCoord: Differentiable Coordination for Distributed Multi-Agent Trajectory Optimization}
%Neural Moving Horizon Estimation\\ for Robotic Control

\author{Bingheng Wang, Yichao Gao, Tianchen Sun, Shanker Ajay, and Lin Zhao% <-this % stops a space

% \thanks{The videos and source code of this work are available at \url{https://github.com/RCL-NUS/NeuroMHE}.} 
\thanks{The authors are with the Department of Electrical and Computer Engineering,
        National University of Singapore, Singapore 117583, Singapore (email: 
        {wangbingheng@u.nus.edu}, {yichao\_gao@u.nus.edu}, {tianc.s@nus.edu.sg}, {ajay\_shanker@u.nus.edu},
        {elezhli@nus.edu.sg}).}
        }

        % <-this % stops a space
% \thanks{J. Doe and J. Doe are with Anonymous University.}% <-this % stops a space
% \thanks{Manuscript received July 11, 2022}}

\newcommand{\Lin}[1]{\textcolor{blue}{[#1]}}

\markboth{IEEE XXX}%
{B. Wang \MakeLowercase{\textit{et al.}}: DiffCoord}



\maketitle
\begin{abstract} 

Integrating the Alternating Direction Method of Multipliers (ADMM) with Differential Dynamic Programming (DDP) provides a scalable framework for distributed multi-agent trajectory optimization. In practice, ADMM is typically truncated for computational efficiency, tightly coupling parameters that would otherwise separately govern coordination quality and task performance. In this paper, we propose Differentiable Coordination (DiffCoord), a unified framework that jointly meta-learns these coupled parameters for the truncated ADMM-DDP pipeline. These parameters are generated by agent-wise neural networks for task adaptation, and the same networks are shared among isomorphic agents to enable scalability to varying agent counts. We achieve efficient meta-learning by differentiating the ADMM-DDP pipeline end-to-end. Notably, this yields an auxiliary ADMM-LQR distributed gradient solver that computes and coordinates meta-gradients with respect to these parameters. This solver inherits the computational structure of the pipeline, enabling reuse of key computation results and efficient parallelization over agents and along trajectory horizons. We validate DiffCoord through numerical and physical experiments on a cooperative aerial transport system, where it reconfigures quadrotor formations for safe 6-DoF load manipulation in tight spaces. It adapts robustly to varying team sizes and load dynamics, while reducing per-agent gradient computation time by up to $70\%$ compared with state-of-the-art trajectory-gradient methods.

\end{abstract}


% \fi
\IEEEpeerreviewmaketitle

% \section*{Supplementary Material}
% The videos and source code of this work are available at \url{https://github.com/RCL-NUS/NeuroMHE}.
\section*{Supplementary Material}
The source code of this work is available at \url{https://github.com/BinghengNUS/DiffCoord/tree/main}.

\section{Introduction}
\label{sec:intro}

\IEEEPARstart{T}{rajectory} optimization is critical in multi-agent coordination, providing reference and feedforward trajectories for downstream controllers. In tasks such as aerial transport of a cable-suspended load~\cite{10328685}, multi-arm manipulation~\cite{1192151}, and legged locomotion~\cite{10341987}, trajectory optimization must ensure that each agent respects its own dynamics while coordinating with others through inter-agent constraints. These include managing cable tensions, synchronizing contact forces, and maintaining formations. Effective multi-agent trajectory optimization demands fast computation, scalability with the number of agents, and adaptability to varying tasks, while minimizing manual design and tuning effort.

\begin{figure}
    \centering
    \includegraphics[width=0.8\linewidth]{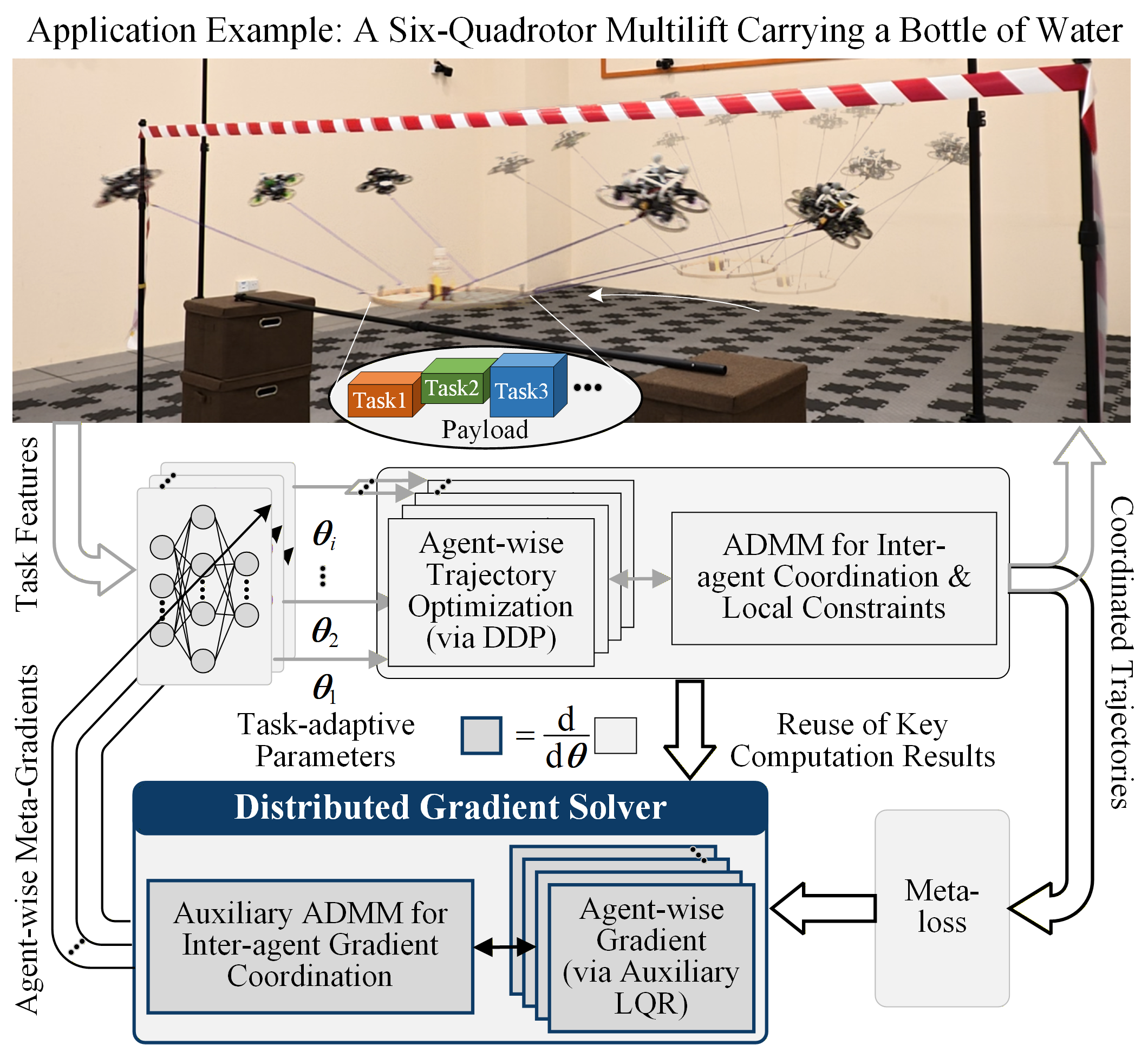}
    \caption{\footnotesize Overview of the DiffCoord framework. We integrate an ADMM-DDP pipeline with agent-wise neural networks that map task features to task-adaptive parameters. The same networks are shared among isomorphic agents, enabling scalability to varying agent counts. Central to our framework is a distributed gradient solver that computes and coordinates meta-gradients with respect to these parameters. This solver inherits the dynamic-static computational structure of the ADMM-DDP pipeline, enabling parallelization and reuse of key computation results for efficient meta-gradient computation.}
    \label{fig:diagram}
\end{figure}

Centralized trajectory optimization provides fully coordinated solutions~\cite{1184275}, but scales poorly with the number of agents. Distributed methods address this limitation by decomposing the problem into agent-wise local subproblems that are coupled through coordination constraints. The Alternating Direction Method of Multipliers (ADMM) is a well-established framework for enforcing such coupling~\cite{boyd2011distributed}. It alternates between solving local subproblems in parallel and updating shared dual variables, iteratively driving the agents towards coordinated solutions, and has been successfully applied in a range of distributed optimization problems~\cite{7353543,9993352}.

When ADMM is applied to multi-agent planning and control~\cite{9736627,8814732}, each local subproblem typically becomes a dynamic optimization problem governed by the corresponding agent’s nonlinear dynamics. Differential Dynamic Programming (DDP) is a particularly effective solver for such subproblems. It iteratively applies Bellman's principle with second-order approximations to compute locally optimal trajectories and feedback control policies, with computational complexity linear in the horizon length and well-studied convergence properties~\cite{MAYNE01011966,86943,liao1992advantages,6225174,9561530,9981586,9981476}. Recent works have combined ADMM with DDP for distributed multi-agent trajectory optimization~\cite{9485100,8894001,10288223}. In particular, Saravanos et al.~\cite{10288223} improved the scalability of this framework to systems with thousands of agents by developing decentralized architectures. In their ADMM-DDP pipeline, DDP optimizes each agent’s dynamically feasible trajectory independently. Safe copy variables are introduced to satisfy local and inter-agent constraints at each time step, and ADMM iterations drive the trajectories and their safe copies toward consensus.

A key practical challenge in the ADMM-DDP pipeline is the high computational cost of ADMM convergence. Although adaptive penalty strategies~\cite{he2000alternating,song2016fast,xu2017adaptive,saravanosdeep} can improve the practical convergence behavior, ADMM often still requires many iterations to reach acceptable residual tolerances. This cost becomes more pronounced in long-horizon multi-agent planning, where each ADMM iteration involves solving dynamic trajectory subproblems. For practical use, it is often preferable to truncate the ADMM iterations and trade exact consensus for computational efficiency.

This truncation changes the role of parameters within the pipeline. When ADMM-DDP fully converges and satisfies the inter-agent coupling constraints, it returns coordinated trajectories governed mainly by the problem formulation. We refer to parameters defining this formulation, such as cost weights, references, dynamics-model parameters, and constraint parameters, as problem-level parameters. In this ideal regime, the ADMM penalty parameters, referred to as solver-level parameters, primarily affect the convergence behavior but not the final trajectories. A natural strategy is to tune the solver-level parameters for fast and stable convergence, and then tune the problem-level parameters for task performance. However, this separation breaks down under truncation. The truncated ADMM-DDP solution depends jointly on both classes of parameters, which interact in a nonlinear and task-dependent manner. The existing ADMM-DDP-based methods rely largely on manual tuning of these parameters~\cite{9485100,8894001,10288223}, a laborious process that demands substantial expert intuition and generalizes poorly across tasks. This difficulty is further compounded under a limited ADMM iteration budget, where the solver-level penalty parameters must simultaneously shape the dynamic DDP updates, the static safe-copy feasibility updates, and the residuals passed to subsequent ADMM iterations.

In this paper, we propose Differentiable Coordination (DiffCoord), a unified meta-learning framework built on the truncated ADMM-DDP pipeline for distributed multi-agent trajectory optimization (see Fig.~\ref{fig:diagram}). DiffCoord employs agent-wise neural networks to generate task-adaptive solver-level and problem-level parameters from diverse task features. The same networks are shared among isomorphic agents, enabling generalization across varying team sizes without retraining. To guide efficient learning, we design a meta-loss that encodes both agent-wise task performance and inter-agent coordination, with the latter measured by ADMM residuals. This residual-aware objective promotes coordination under truncation without requiring pre-solved centralized solutions, unlike~\cite{saravanosdeep}, which relies on a centralized solution dataset. Our approach thus combines the representational flexibility of neural networks with the interpretability and constraint-handling capability of first-principles distributed optimal control. In contrast to model-free multi-agent reinforcement learning~\cite{9201368,9623508,GU2023103905,s23073625, zengdecentralized}, DiffCoord exploits the differentiable structure of the truncated ADMM-DDP pipeline to compute analytical meta-gradients, enabling data-efficient training and optimization-structured task adaptation. Recent works have explored the integration of ADMM-based optimization and learning~\cite{pmlr-v97-xie19c,saravanosdeep,sun2023alternating,Andrew2023,11119192}. Most of these methods focus on static or convex optimization problems, learning either solver-level parameters for improved solver behavior or problem-level parameters for improved solution quality. DiffCoord contributes a complementary direction: a learning-empowered, structure-exploiting differentiable coordination framework built on the truncated ADMM-DDP pipeline for efficient, scalable, and distributed multi-agent trajectory optimization.

At the core of our approach is the end-to-end differentiation of the truncated ADMM-DDP pipeline with respect to (w.r.t.) its parameters, which yields an auxiliary ADMM-LQR distributed gradient solver. This requires not only differentiating the agent-wise dynamic optimization problems solved by DDP, but also coordinating their trajectory gradients through the ADMM structure. In single-agent problems, a growing body of work has explored the differentiation of dynamic optimization problems either by unrolling the solver for automatic differentiation~\cite{okada2017path,LandryDifferentiable}, or by applying the implicit function theorem to the optimality conditions~\cite{10.5555/3327757.3327922,dinev2022differentiable,Alexdiffrobust,Angel2025actorcriticmpc,NEURIPS2020_5a7b238b,NEURIPS2021_85ea6fd7,liang2025onlinecontrolinformedlearning,11214470}. Among the latter, vector-Jacobian product (VJP) methods~\cite{10.5555/3327757.3327922,dinev2022differentiable,Alexdiffrobust,Angel2025actorcriticmpc} efficiently compute scalar-loss gradients w.r.t. problem-level parameters without explicitly forming full trajectory Jacobians, i.e., matrix-valued trajectory gradients. Full-Jacobian methods, such as Pontryagin Differential Programming (PDP) and its variants~\cite{NEURIPS2020_5a7b238b,NEURIPS2021_85ea6fd7,liang2025onlinecontrolinformedlearning,11214470}, instead compute these trajectory gradients recursively, but at higher computational and memory costs. In ADMM-based multi-agent problems, such trajectory gradients are essential for gradient coordination across agents, a requirement not addressed by the VJP methods.

Interestingly, we show that differentiating the ADMM-DDP pipeline reveals a two-level structural mirroring. At the dynamic optimization level, each agent's trajectory gradients correspond to the optimal solution of an auxiliary matrix-valued Linear Quadratic Regulator (LQR). Its recursive solution produces the same trajectory gradients as the full-Jacobian methods, yet achieves higher efficiency by exploiting its structural overlap with the DDP solve. This overlap enables DDP quantities, such as Riccati recursions and feedback gains, to be reused rather than recomputed. At the coordination level, these auxiliary LQR problems are coupled by an auxiliary ADMM pipeline that mirrors the dynamic–static update pattern of the original ADMM-DDP pipeline: agent-wise LQR gradient subproblems correspond to the dynamic DDP updates, while auxiliary safe-copy gradient updates correspond to the static coordination subproblems. The auxiliary pipeline thus reuses the same ADMM penalty parameters and iteration budget, rather than requiring separately tuned ADMM penalties for the auxiliary gradient solver. This two-level mirroring makes the distributed gradient computation structure-exploiting: much of the computational effort is shared with the ADMM-DDP solve. Under mild assumptions, the auxiliary problem is convex, which facilitates theoretical analysis. Through ADMM-enabled parallelization, the distributed gradient solver scales efficiently with both agent count and trajectory horizon length.

We validate DiffCoord on the multilift system~\cite{michael2011cooperative,sreenath2013dynamics,7759262,7843619,8567931,geng2020cooperative,10492665,wang2024auto}, where multiple quadrotors cooperatively transport a cable-suspended payload (see top of Fig.~\ref{fig:diagram}). The cables induce tight dynamic coupling between the payload and each quadrotor through tension forces, and impose kinematic constraints that limit each quadrotor's workspace. The system is therefore highly sensitive to both local trajectories and inter-agent coordination, making it a demanding benchmark for distributed trajectory optimization. Prior trajectory optimization approaches in this domain include centralized methods~\cite{doi:10.1177/0278364910382803,geng2022load,sun2023nonlinear,wahba2024kinodynamic,11455369,sun2025agile,11123816,CUI2026107037} and non-ADMM distributed methods~\cite{9341541,9007450,agarwal2025decentralized}. These approaches often rely on manually specified objectives and parameters, fixed team sizes, conservative formation shapes, or hand-crafted formation reconfigurations to pass through narrow passages. These designs limit adaptability and generalization, often requiring substantial redesign when the team size, task objectives, or load dynamics change. 

In contrast, DiffCoord meta-learns adaptive ADMM-DDP parameters that generalize across varying team sizes and task conditions, including shifts in the payload center of mass that require different tension distributions and formation shapes. Training is highly efficient. The DDP-level mirroring yields up to $70\%$ faster per-agent gradient computation than the existing full-Jacobian methods~\cite{NEURIPS2020_5a7b238b,11214470}, while the ADMM-level parallelization enables the overall gradient computation to scale efficiently with agent count and trajectory horizon. In real-flight experiments, we train the multilift system with 4 quadrotors and deploy the meta-learned networks to the ADMM-DDP pipelines for two teams of 3 and 6 quadrotors carrying a bottle of water, without extra tuning. The system generates dynamically feasible and safe trajectories through unseen obstacles, including narrow passages and horizontally oriented gaps. The meta-learned DDP feedback gains further improve payload tracking by closing the loop against model uncertainties such as liquid sloshing and downwash airflow.

We summarize the main contributions as follows.
\begin{enumerate}
\item We propose DiffCoord, a unified framework for distributed meta-trajectory optimization that jointly meta-learns solver-level and problem-level parameters of the truncated ADMM-DDP pipeline. The parameters are modeled by agent-wise neural networks shared among isomorphic agents for scalability and task adaptation. Central to DiffCoord is a two-level structural mirroring revealed by end-to-end differentiation of the truncated ADMM-DDP pipeline.
\item At the dynamic optimization level, we show that each agent’s trajectory gradients correspond to the optimal solution of an auxiliary matrix-valued LQR system. Its structural overlap with the DDP solve enables reuse of Riccati recursions and feedback gains, achieving significant acceleration over the existing full-Jacobian methods in long-horizon settings.
\item At the coordination level, we show that the agent-wise LQR gradient problems are coordinated by an auxiliary ADMM framework. It inherits the dynamic–static coordination structure of the ADMM-DDP pipeline, reuses the penalty parameters and iteration budget, admits a convex formulation under mild conditions, and yields Lipschitz-bounded truncation errors. Its ADMM-enabled parallelization allows the overall gradient computation to scale efficiently with team size and trajectory horizon.
\item We validate DiffCoord on the multilift task through real-flight experiments. Trained with 4 quadrotors, the meta-learned networks are deployed to the 3- and 6-quadrotor ADMM-DDP pipelines without extra tuning, demonstrating scalability, obstacle adaptation, and robust payload tracking under model uncertainties.
\end{enumerate}

The rest of the paper is organized as follows: Section~\ref{section:preliminaries} reviews the ADMM-DDP pipeline. Section~\ref{section:DiffCoord} presents the proposed DiffCoord framework, and Section~\ref{section:analytical grad} derives the analytical gradients and auxiliary ADMM coordination. Section~\ref{section:multilift} applies DiffCoord to multilift systems, with experiments in Section~\ref{section:experiments}. Section~\ref{section:conclusion} concludes the paper.

\section{Preliminaries}\label{section:preliminaries}
% \subsection{Notation}
%     In this paper, the 2-norm of $\bm{x}\in \mathbb{R}^n$ is $\|\bm{x}\|_2$ or $\|\bm{x}\|$, and the norm weighted by a positive definite matrix $\bm{W}$ is denoted with $\|\bm{x}\|_{\bm W}\coloneqq \sqrt{\bm{x}^{\top}\bm{W}\bm{x}}$. For a differentiable scalar function $f : \mathbb{R}^{{\rm{dim}}(\bm{x})} \to \mathbb{R}$, we denote its gradient and Hessian as $\nabla_{\bm{x}} f$ and $\nabla^2_{\bm{x}} f$. For a function $f$ with multiple arguments, we use $\frac{\partial f}{\partial (\cdot)}$ and $\frac{\mathrm{d} f}{\mathrm{d} (\cdot)}$ respectively to denote the partial and total derivatives. Vectorization of a matrix is denoted with $\operatorname{vec}(\cdot)$. Finally, let $\mathcal{I}_{\mathcal{C}}(\bm{x})$ denote an indicator function such that $\mathcal{I}_c(\bm{x})=0$ if $\bm{x}\in \mathcal{C}$, otherwise $\mathcal{I}_c(\bm{x})=+\infty$.

\subsection{Centralized multi-agent Trajectory Optimization}
Consider $n$ robots collaborating on a task such as cooperative transport or manipulation. Their trajectories are found by minimizing the sum of local costs subject to dynamics, local constraints, and inter-agent constraints, yielding the centralized trajectory optimization problem:
\begin{subequations}
    \begin{align}
        \min_{\{\bm{x}_i, \bm{u}_i\}_{i=1}^n} & \sum_{i=1}^n J_i(\bm{x}_i,\bm{u}_i) \label{eq:total cost} \\
        \text{s.t.} \quad 
        & \bm{x}_{i,k+1} = \bm{f}_i(\bm{x}_{i,k}, \bm{u}_{i,k}), \ \bm{x}_{i,0}:\text{given}, \label{eq:robot dynamics}\\
        & g_{i,k}(\bm{x}_{i,k},\bm{u}_{i,k}) \le 0, \label{eq:local constraint} \\
        & h_{ij,k}(\bm{x}_{i,k}, \bm{x}_{j,k}, \bm{u}_{i,k}, \bm{u}_{j,k})=0,\label{eq:inter-robot equality constraint} \\
        & g_{ij,k}(\bm{x}_{i,k}, \bm{x}_{j,k}, \bm{u}_{i,k}, \bm{u}_{j,k}) \le 0, \label{eq:inter-robot inequality constraint} 
    \end{align}
    \label{eq:centralized optimization}%
\end{subequations}
where $\bm{x}_{i,k} \in \mathbb{R}^{n_i}$ and $\bm{u}_{i,k} \in \mathbb{R}^{m_i}$ denote the state and control of robot $i$ at time step $k$, with the corresponding trajectories $\bm{x}_i = [\bm{x}_{i,0},\dots,\bm{x}_{i,N}]$ and $\bm{u}_i = [\bm{u}_{i,0},\dots,\bm{u}_{i,N-1}]$ over the horizon length $N$. The local cost for each robot is given by:
\begin{equation}
    J_i(\bm{x}_i,\bm{u}_i)=\sum_{k=0}^{N-1}\ell_{i,k}(\bm{x}_{i,k},\bm{u}_{i,k}) + \ell_{i,N}(\bm{x}_{i,N}),
    \label{eq:local cost}
\end{equation}
where $\ell_{i,k}: \mathbb{R}^{n_i} \times \mathbb{R}^{m_i} \to \mathbb{R}$ and $\ell_{i,N}: \mathbb{R}^{n_i} \to \mathbb{R}$ are the stage and terminal costs. In~\eqref{eq:centralized optimization}, $\bm{f}_i$ denotes the $i$th robot's discrete-time dynamics, \eqref{eq:local constraint} is the local state and control constraint, \eqref{eq:inter-robot equality constraint} and \eqref{eq:inter-robot inequality constraint} are the inter-agent constraints (e.g., load sharing and collision avoidance).

\subsection{Distributed ADMM-DDP Pipeline}
The centralized problem \eqref{eq:centralized optimization} becomes computationally prohibitive for large-scale systems. ADMM offers a scalable alternative by decomposing computation and enforcing coordination iteratively. ADMM typically handles problems with separable objectives and affine constraints:
\begin{equation}
    \min_{\bm{x},\bm{z}} f(\bm{x})+g(\bm{z}) \quad \text{s.t.} \ \bm{A}\bm{x}+\bm{B}\bm{z}=\bm{c}.
    \label{eq:standard admm}
\end{equation}
% with decision variables $\bm{x}\in\mathbb{R}^{p}$, $\bm{z}\in\mathbb{R}^{q}$, separable cost functions $f:\mathbb{R}^{p}\to\mathbb{R}$, $g:\mathbb{R}^{q}\to\mathbb{R}$, and affine constraints defined by $\bm{A}\in\mathbb{R}^{r\times p}$, $\bm{B}\in\mathbb{R}^{r\times q}$, $\bm{c}\in\mathbb{R}^{r}$. 
The classical ADMM solves \eqref{eq:standard admm} via:
\begin{equation}
    \begin{aligned}
        \bm{x}^{a}&=\arg\min_{\bm{x}}\mathcal{L}(\bm{x},\bm{z}^{a-1},\bm{\lambda}^{a-1}),\\
        \bm{z}^{a}&=\arg\min_{\bm{z}}\mathcal{L}(\bm{x}^{a},\bm{z},\bm{\lambda}^{a-1}),\\
\bm{\lambda}^{a}&=\bm{\lambda}^{a-1}+\rho(\bm{A}\bm{x}^{a}+\bm{B}\bm{z}^{a}-\bm{c}),
    \end{aligned}
    \label{eq:classical admm update}
\end{equation}
where the superscript $a\in \mathbb{N}_+$ denotes the iteration index, $\bm{\lambda}$ is the dual variable for the constraint, and $\hat{\mathcal{L}}$ is the augmented Lagrangian given by
\begin{equation}
    \begin{aligned}
        \hat{\mathcal{L}}(\bm{x},\bm{z},\bm{\lambda})&=f(\bm{x})+g(\bm{z})+\bm{\lambda}^{\top}(\bm{A}\bm{x}+\bm{B}\bm{z}-\bm{c})\\
        &\quad  + \frac{\rho}{2}\|\bm{A}\bm{x}+\bm{B}\bm{z}-\bm{c}\|_2^2,
    \end{aligned}
    \nonumber
\end{equation}
with $\rho >0$ as the ADMM penalty parameter. These updates enforce the constraint iteratively by penalizing violations while updating the dual variable (see~\cite{boyd2011distributed} for details).

However, applying ADMM directly to multi-agent trajectory optimization is challenging due to nonlinear, nonconvex inter-agent constraints~\eqref{eq:inter-robot equality constraint} and~\eqref{eq:inter-robot inequality constraint}. A recent work \cite{10288223} addresses this by introducing~\textit{safe-copy variables}. These variables satisfy both local and inter-agent constraints, and enable affine consensus constraints, aligning the resulting problem with the standard ADMM form. 

Adopting this strategy, we define safe-copy variables $\bm{\Tilde{x}}_{i,k}\in \mathbb{R}^{n_i}$ and $\bm{\Tilde{u}}_{i,k}\in \mathbb{R}^{m_i}$ for each agent's state and control. They remain consensus with the original variables via:
\begin{equation}
    \bm{x}_{i,k}=\bm{\Tilde{x}}_{i,k},\ \bm{u}_{i,k}=\bm{\Tilde{u}}_{i,k}, \quad \forall i\in \{1,\ldots,n\}.
    \label{eq:consensus constraint}
\end{equation}
The corresponding augmented Lagrangian is:
\begin{equation}
\begin{aligned}
    \hat{\mathcal{L}}_{\text{c}}&= \sum_{i=1}^n J_i(\bm{x}_i,\bm{u}_i)+\mathcal{I}_{\bm{f}_i}(\bm{x}_i,\bm{u}_i)+\mathcal{I}_{g_i}(\bm{\Tilde{x}}_i,\bm{\Tilde{u}}_i) \\
    &\quad+ \mathcal{I}_{h_{ij}}(\bm{\Tilde{x}}_i,\bm{\Tilde{x}}_j,\bm{\Tilde{u}}_i,\bm{\Tilde{u}}_j)+\mathcal{I}_{g_{ij}}(\bm{\Tilde{x}}_i,\bm{\Tilde{x}}_j,\bm{\Tilde{u}}_i,\bm{\Tilde{u}}_j)\\
    &\quad+\bm{\nu}^{\top}_i(\bm{x}_i-\bm{\Tilde{x}}_i)+\bm{\xi}^{\top}_i(\bm{u}_i-\bm{\Tilde{u}}_i)\\
    &\quad+\frac{\rho_{i}}{2}\|\bm{x}_i-\bm{\Tilde{x}}_i \|_2^2+\frac{\sigma_i}{2}\|\bm{u}_i-\bm{\Tilde{u}}_i \|_2^2,
\end{aligned}
\label{eq:augmented lagrangian for our problem}
\end{equation}
where $\mathcal{I}_{\mathcal{C}}(\bm{x})$ denote an indicator function such that $\mathcal{I}_c(\bm{x})=0$ if $\bm{x}\in \mathcal{C}$, otherwise $\mathcal{I}_c(\bm{x})=+\infty$, $\bm{\nu}_i\in\mathbb{R}^{n_i}$ and $\bm{\xi}_i\in\mathbb{R}^{m_i}$ are the dual variables for the state and control consensus constraints of agent $i$, $\rho_i$ and $\sigma_i$ are the ADMM penalty parameters for the state and control, respectively. The distributed formulation proceeds in the following three subproblems.

\subsubsection{Subproblem 1}\label{subsubp1} Each agent independently minimizes~\eqref{eq:augmented lagrangian for our problem} w.r.t. its local state and control trajectories. This yields the following $n$ dynamic optimization subproblems:
\begin{equation}
    \begin{aligned}
        \{\bm{x}_i, \bm{u}_i\}^{a}&=\arg\min_{\bm{x}_i,\bm{u}_i}\sum_{k=0}^{N-1}\hat{\ell}_{i,k}^{a} + \hat{\ell}_{i,N}^{a}\\
        \text{s.t.}\quad
        &  \bm{x}_{i,k+1}^{a} = \bm{f}_{i,k}^{a}(\bm{x}_{i,k}^{a}, \bm{u}_{i,k}^{a}), \ \bm{x}_{i,0}:\text{given},
    \end{aligned}
    \label{eq:subproblem1}
\end{equation}
where 
\begin{equation}
    \begin{aligned}
        \hat{\ell}_{i,k}^{a}&=\ell_{i,k}^{a}(\bm{x}_{i,k}^{a},\bm{u}_{i,k}^{a})+\frac{\rho_i^a}{2}\left\|\bm{x}_{i,k}^{a}-\bm{\Tilde{x}}_{i,k}^{a-1}+ \frac{\bm{\nu}_{i,k}^{a-1}}{\rho_i^a} \right\|_2^2\\
        &\quad + \frac{\sigma_i^a}{2}\left\|\bm{u}_{i,k}^{a}-\bm{\Tilde{u}}_{i,k}^{a-1}+ \frac{\bm{\xi}_{i,k}^{a-1}}{\sigma_i^a} \right\|_2^2,\\
        \hat{\ell}_{i,N}^{a}&=\ell_{i,N}^{a}(\bm{x}_{i,N}^{a})+\frac{\rho_i^a}{2}\left\|\bm{x}_{i,N}^{a}-\bm{\Tilde{x}}_{i,N}^{a-1}+ \frac{\bm{\nu}_{i,N}^{a-1}}{\rho_i^a} \right\|_2^2.
    \end{aligned}
    \label{eq:DPP cost augmented by ADMM}
\end{equation}

\subsubsection{Subproblem 2}\label{subsubp2} Given $\{\bm{x}_i,\bm{u}_i\}^a$ from Subproblem 1, all agents then minimize~\eqref{eq:augmented lagrangian for our problem} w.r.t. their safe-copy variables. The resulting problem is centralized because these variables are coupled via the inter-agent constraints~\eqref{eq:inter-robot equality constraint} and~\eqref{eq:inter-robot inequality constraint}. However, since the agent dynamics do not depend on the safe-copy variables, this subproblem reduces to a static optimization that can be decomposed temporally. Together with Subproblem 1, this static safe-copy update yields the dynamic–static decomposition of the ADMM-DDP pipeline. The step-wise static optimization subproblem at $k\in [0,N-1]$ is defined by:
\begin{equation}
    \begin{aligned}
        \{\bm{\Tilde{x}}_k, \bm{\Tilde{u}}_k\}^{a}
        &= \mathop{\arg\min}_{\{\bm{\Tilde{x}}_k,\, \bm{\Tilde{u}}_k\}^{a}}\!
        \sum_{i=1}^n \left( e_{x_i,k}^a + e_{u_i,k}^a\right) \\
        \text{s.t.}&\quad \bm{g}_{i,k}^a(\bm{\Tilde{x}}_{i,k}^{a},\bm{\Tilde{u}}_{i,k}^{a}) \le 0, \\
        &\quad \bm{h}_{ij,k}^a(\bm{\Tilde{x}}_{i,k}^{a}, \bm{\Tilde{x}}_{j,k}^{a}, \bm{\Tilde{u}}_{i,k}^{a}, \bm{\Tilde{u}}_{j,k}^{a})=0,\\
        &\quad \bm{g}_{ij,k}^a(\bm{\Tilde{x}}_{i,k}^{a}, \bm{\Tilde{x}}_{j,k}^{a}, \bm{\Tilde{u}}_{i,k}^{a}, \bm{\Tilde{u}}_{j,k}^{a}) \le 0.
    \end{aligned}
    \label{eq:stage subproblem 2}
\end{equation}
where $e_{x_i,k}^a=\frac{\rho_i^a}{2} \left\| \bm{x}_{i,k}^{a} - \bm{\Tilde{x}}_{i,k}^a + \bm{\nu}_{i,k}^{a-1}/\rho_i^a \right\|_2^2 $ and $e_{u_i,k}^a=\frac{\sigma_i^a}{2} \left\| \bm{u}_{i,k}^{a} - \bm{\Tilde{u}}_{i,k}^a + \bm{\xi}_{i,k}^{a-1}/\sigma_i^a \right\|_2^2$. Similarly, the terminal static optimization subproblem at $k=N$ is defined by:
\begin{equation}
    \begin{aligned}
        \bm{\Tilde{x}}_N^{a} &=\mathop{\arg\min}_{\bm{\Tilde{x}}_N}\!\sum_{i=1}^n e_{x,i,N}^a\\
        \text{s.t.}&\quad \bm{g}_{i,N}^a(\bm{\Tilde{x}}_{i,N}^{a}) \le 0, \ \bm{h}_{ij,N}^a(\bm{\Tilde{x}}_{i,N}^{a}, \bm{\Tilde{x}}_{j,N}^{a}) = 0,\\
        &\quad \bm{g}_{ij,N}^a(\bm{\Tilde{x}}_{i,N}^{a}, \bm{\Tilde{x}}_{j,N}^{a}) \le 0.
    \end{aligned}
    \label{eq:terminal subproblem 2}
\end{equation}

\begin{remark}\label{remark:decentralized subproblem2}
    These static subproblems can be further decentralized by augmenting each agent’s safe-copy variables with local copies of neighboring agents’ variables. Let $\bm{\Tilde{x}}_{i,k}^{\mathrm{a}} = [\bm{\Tilde{x}}_{j,k}^{i}]_{j\in \mathcal{N}_i}$ denote the augmented safe-copy state of agent $i$, where $\mathcal{N}_i$ denotes the neighborhood set of agent $i$ (including itself), $\bm{\Tilde{x}}_{i,k}^{i} = \bm{\Tilde{x}}_{i,k}$ is the self-copy, and $\bm{\Tilde{x}}_{j,k}^{i}$, $j \ne i$, are local copies of neighboring agents’ states; the same applies to the controls. This yields fully decentralized subproblems consistent with the merged distributed structure in~\cite{10288223}. We develop DiffCoord using the above centralized formulation to minimize the number of safe-copy variables, but it naturally extends to the fully decentralized case via the augmented safe-copy formulation.
\end{remark}

\subsubsection{Subproblem 3}\label{subsubp3} Given $\{\bm{x}_i,\bm{u}_i\}^a$ from Subproblem 1 and $\{\bm{\Tilde{x}}_k, \bm{\Tilde{u}}_k\}^{a}$ from Subproblem 2, each agent updates its dual variables independently at each time step according to:
\begin{subequations}
    \begin{align}
        \bm{\nu}_{i,k}^{a}&=\bm{\nu}_{i,k}^{a-1}+\rho_i^a(\bm{x}_{i,k}^{a}-\bm{\Tilde{x}}_{i,k}^{a}),\label{eq:dual_state}\\
        \bm{\xi}_{i,k}^{a}&=\bm{\xi}_{i,k}^{a-1}+\sigma_i^a(\bm{u}_{i,k}^{a}-\bm{\Tilde{u}}_{i,k}^{a})\label{eq:dual_control}.
    \end{align}
    \label{eq:subproblem 3}%
\end{subequations}

The dynamic optimization problem~\eqref{eq:subproblem1} can be efficiently solved via DDP, a second order trajectory optimization method based on Bellman’s principle~\cite{bellman1966dynamic}. It states that any remaining segment of an optimal trajectory is itself optimal, enabling decomposition into smaller subproblems. This principle leads to a recursive formulation of the value function (the optimal cost-to-go). For agent $i$, the value function at ADMM iteration $a$ and time step $k$ is defined as
\begin{equation}
    \hat{V}_{i,k}^{a}(\bm{x}_{i,k}^a)=\min_{\bm{u}_{i,k}^a}\hat{Q}_{i,k}^{a}(\bm{x}_{i,k}^a,\bm{u}_{i,k}^a),
    \label{eq:value function}
\end{equation}
where the Q-function (the cost-to-go) is defined using the augmented stage cost \eqref{eq:DPP cost augmented by ADMM} as
\begin{equation}
    \begin{aligned}
        \hat{Q}_{i,k}^{a}(\bm{x}_{i,k}^a,\bm{u}_{i,k}^a)&=\hat{\ell}_{i,k}^{a}(\bm{x}_{i,k}^a,\bm{u}_{i,k}^a,\bm{\Tilde{x}}_{i,k}^{a-1},\bm{\Tilde{u}}_{i,k}^{a-1},\bm{\nu}_{i,k}^{a-1},\bm{\xi}_{i,k}^{a-1}) \\
        &\quad + \hat{V}_{i,k+1}^{a}(\bm{f}_{i,k}^a(\bm{x}_{i,k}^a,\bm{u}_{i,k}^a)), 
    \end{aligned}
    \label{eq:Q function}
\end{equation}
with terminal condition
\begin{equation}
    \hat{V}_{i,N}^{a}(\bm{x}_{i,N}^a)=\hat{\ell}_{i,N}^{a}(\bm{x}_{i,N}^a,\bm{\Tilde{x}}_{i,N}^{a-1},\bm{\nu}_{i,N}^{a-1}).
    \label{eq:terminal Q function}
\end{equation}

For nonlinear systems, the Q-function is generally complex and does not admit an analytical solution. DDP approximates it to second order around nominal trajectories $\{ \bm{\bar{x}}_{i},\bm{\bar{u}}_{i}\}^a$. Let $\delta\bm{x}_{i,k}^a=\bm{x}_{i,k}^a-\bm{\bar{x}}_{i,k}^a$ and $\delta\bm{u}_{i,k}^a=\bm{u}_{i,k}^a-\bm{\bar{u}}_{i,k}^a$ denote the deviations. The quadratic approximation of $\hat{Q}_{i,k}^{a}$ is given by
\begin{equation}
    \begin{aligned}
        \hat{Q}_{i,k}^{a}(\bm{x}_{i,k}^a,\bm{u}_{i,k}^a) &\approx \hat{Q}_{i,k}^{a}(\bm{\bar{x}}_{i,k}^a,\bm{\bar{u}}_{i,k}^a) +(\delta\bm{x}_{i,k}^a)^{\top}\hat{Q}_{i,k}^{xu,a}\delta\bm{u}_{i,k}^a \\
        &\quad + (\hat{Q}_{i,k}^{x,a})^{\top}\delta\bm{x}_{i,k}^a+ \frac{1}{2}(\delta\bm{x}_{i,k}^a)^{\top}\hat{Q}_{i,k}^{xx,a}\delta\bm{x}_{i,k}^a\\
        &\quad + (\hat{Q}_{i,k}^{u,a})^{\top}\delta\bm{u}_{i,k}^a + \frac{1}{2}(\delta\bm{u}_{i,k}^a)^{\top}\hat{Q}_{i,k}^{uu,a}\delta\bm{u}_{i,k}^a,
    \end{aligned}
    \label{eq:approximated value function}
\end{equation}
with
\begin{subequations}
    \begin{align}
        \hat{Q}_{i,k}^{x,a}&=\hat{\ell}_{i,k}^{x,a}+({f}_{i,k}^{x,a})^{\top}\hat{V}_{i,k+1}^{x,a},\label{eq:Q_x}\\
        \hat{Q}_{i,k}^{u,a}&=\hat{\ell}_{i,k}^{u,a}+({f}_{i,k}^{u,a})^{\top}\hat{V}_{i,k+1}^{x,a},\label{eq:Q_u}\\
        \begin{split}
            \hat{Q}_{i,k}^{xx,a}&=\hat{\ell}_{i,k}^{xx,a}+({f}_{i,k}^{x,a})^{\top}\hat{V}_{i,k+1}^{xx,a}{f}_{i,k}^{x,a} + \hat{V}_{i,k+1}^{x,a} \cdot {f}_{i,k}^{xx,a}
        \end{split}\label{eq:Q_xx}\\
        \begin{split}
            \hat{Q}_{i,k}^{xu,a}&=\hat{\ell}_{i,k}^{xu,a}+({f}_{i,k}^{x,a})^{\top}\hat{V}_{i,k+1}^{xx,a}{f}_{i,k}^{u,a} + \hat{V}_{i,k+1}^{x,a}\cdot{f}_{i,k}^{xu,a}
        \end{split}\label{eq:Q_xu}\\
        \begin{split}
            \hat{Q}_{i,k}^{uu,a}&=\hat{\ell}_{i,k}^{uu,a}+({f}_{i,k}^{u,a})^{\top}\hat{V}_{i,k+1}^{xx,a}{f}_{i,k}^{u,a} + \hat{V}_{i,k+1}^{x,a}\cdot{f}_{i,k}^{uu,a}
        \end{split}\label{eq:Q_uu}
    \end{align}
    \label{eq:derivatives in approx Q}%
\end{subequations}
where $(\cdot)^{x} \coloneqq \nabla_{\bm{x}}(\cdot)$, $(\cdot)^{u} \coloneqq \nabla_{\bm{u}}(\cdot)$, $(\cdot)^{xx}\coloneqq \nabla_{\bm{x}}^2(\cdot)$, $(\cdot)^{xu}\coloneqq \frac{\partial^2 (\cdot)}{\partial \bm{x} \partial\bm{u}}$, and $(\cdot)^{uu}\coloneqq \nabla_{\bm{u}}^2(\cdot)$ denote the first and second partial derivatives, they are evaluated at the nominal trajectories. The last terms in \eqref{eq:Q_xx}-\eqref{eq:Q_uu} involve second-order derivatives of system dynamics. If omitted, DDP reduces to iterative LQR (iLQR) \cite{9766194}. The optimal control deviation is obtained by minimizing \eqref{eq:approximated value function}:
\begin{equation}
    \delta\bm{u}_{i,k}^a =\underbrace{-(\hat{Q}_{i,k}^{uu,a})^{-1}(\hat{Q}_{i,k}^{xu,a})^{\top}}_{\hat{\bm{K}}_{i,k}^{a}}\delta\bm{x}_{i,k}^a\underbrace{-(\hat{Q}_{i,k}^{uu,a})^{-1}\hat{Q}_{i,k}^{u,a}}_{\hat{\bm{k}}_{i,k}^{a}}, \label{eq:DDP control law}
\end{equation}
where $\hat{\bm{K}}_{i,k}^{a}$ and $\hat{\bm{k}}_{i,k}^{a}$ are the feedback and feedforward gains. In practice, $\hat{\bm{K}}_{i,k}^{a}$ can be used online to design a feedback control law that improves robustness of the optimal trajectories to model uncertainties, as shown in~\cite{10288223} and in our experiments. Plugging $\delta\bm{u}_{i,k}^a$ into \eqref{eq:approximated value function} and matching the resulting approximation of $\hat{Q}_{i,k}^{  a}$ with the second-order Taylor expansion of $\hat{V}_{i,k}^{a}$ yields the following recursive updates:
\begin{subequations}
    \begin{align}
        \hat{V}_{i,k}^{x,a}&=\hat{Q}_{i,k}^{x,a}-\hat{Q}_{i,k}^{xu,a}(\hat{Q}_{i,k}^{uu,a})^{-1}\hat{Q}_{i,k}^{u,a},\label{eq:V_x}\\
        \hat{V}_{i,k}^{xx,a}&=\hat{Q}_{i,k}^{xx,a}-\hat{Q}_{i,k}^{xu,a}(\hat{Q}_{i,k}^{uu,a})^{-1}(\hat{Q}_{i,k}^{xu,a})^{\top}, \label{eq:V_xx}
    \end{align}
    \label{eq:derivatives of V}%
\end{subequations}
with the terminal derivatives as $\hat{V}_{i,N}^{x,a}=\hat{\ell}_{i,N}^{x,a}$ and $\hat{V}_{i,N}^{xx,a}=\hat{\ell}_{i,N}^{xx,a}$. After computing these derivatives \eqref{eq:Q_x}-\eqref{eq:V_xx} backward in time, the nominal trajectories $\{\bm{\bar{x}}_i,\bm{\bar{u}}_i \}^a$ are updated through the following forward iterations:
\begin{subequations}
    \begin{align}
        \bm{u}_{i,k}^a&=\bm{\bar{u}}_{i,k}^a+\hat{\bm{K}}_{i,k}^{a}(\bm{x}_{i,k}^a-\bm{\bar{x}}_{i,k}^a)+\hat{\bm{k}}_{i,k}^{a},\label{eq:new control}\\
        \bm{x}_{i,k+1}^a&=\bm{f}_{i,k}^a(\bm{x}_{i,k}^a,\bm{u}_{i,k}^a),\ \bm{x}_{i,0}^a:\text{given},\label{eq:new state}\\
        \bm{\bar{x}}_{i,k}^a&\gets\bm{x}_{i,k}^a,\ \bm{\bar{u}}_{i,k}^a\gets\bm{u}_{i,k}^a.
    \end{align}
    \label{eq:new DDP trajectories}%
\end{subequations}
The above backward and forward iterations are repeated until convergence, typically determined by a stopping condition such as $\|\delta\bm{u}_{i,k}^a \|_2 \le \epsilon$, where $\epsilon > 0$ is a predefined tolerance.

Finally, the distributed ADMM-DDP pipeline is summarized in Algorithm \ref{alg: ADMM-DDP forward pass} where $a_{\rm{f}}$ denotes the total ADMM iterations. Subproblem 2 can be solved using a generic nonlinear programming (NLP) method (e.g., \texttt{ipopt}~\cite{wachter2002interior}). 

\begin{algorithm}
\caption{Distributed ADMM-DDP Pipeline}
\label{alg: ADMM-DDP forward pass}
\SetKwInput{Input}{Input}
\SetKwInput{Output}{Output}
\SetKwInput{Initialization}{Initialization}
\SetKwComment{Comment}{$\triangleright$\ }{}
\SetNoFillComment
\DontPrintSemicolon
\Initialization{$\{\bm{\Tilde{x}}_{i}^0,\bm{\Tilde{u}}_{i}^0\}_{i=1}^{n}$, $\{\bm{\nu}_{i}^0,\bm{\xi}_i^0\}_{i=1}^{n}$.}
\For {$a \leftarrow 1$ \KwTo $a_{\rm f}$}{
\tcc*[l]{Subproblem 1 (via DDP)}
\For {$i \leftarrow 1$ \KwTo $n$}{
$\{\bm{x}_i,\bm{u}_i \}^{a}\gets$ Solve \eqref{eq:subproblem1} with the cost in~\eqref{eq:DPP cost augmented by ADMM}; 
}
\tcc*[l]{Subproblem 2 (via NLP)}
\For {$k \leftarrow 0$ \KwTo $N$}{
$\{\bm{\Tilde{x}}_k,\bm{\Tilde{u}}_k \}^{a}\gets$ Solve \eqref{eq:stage subproblem 2}; \Comment{$k<N$}
$\bm{\Tilde{x}}_N^{a}\gets$ Solve \eqref{eq:terminal subproblem 2};
}
\tcc*[l]{Subproblem 3}
\For {$i \leftarrow 1$ \KwTo $n$}
{\For {$k \leftarrow 0$ \KwTo $N$}
{$\{\bm{\nu}_{i,k},\bm{\xi}_{i,k}\}^{a}\gets$ Solve \eqref{eq:subproblem 3}; \Comment{$k<N$}
$\bm{\nu}_{i,N}^{a}\gets$ Solve \eqref{eq:dual_state};}
}
}
\Output{$\{\bm{x}_i^{  a_{\rm f}},\bm{u}_i^{  a_{\rm f}}\}_{i=1}^{n}$ and $\{\bm{\Tilde{x}}_i^{  a_{\rm f}},\bm{\Tilde{u}}_i^{  a_{\rm f}}\}_{i=1}^{n}$.}
\end{algorithm}

\section{Formulation of DiffCoord}\label{section:DiffCoord}

\begin{figure*}[h]
    \centering
    \includegraphics[width=0.8\linewidth]{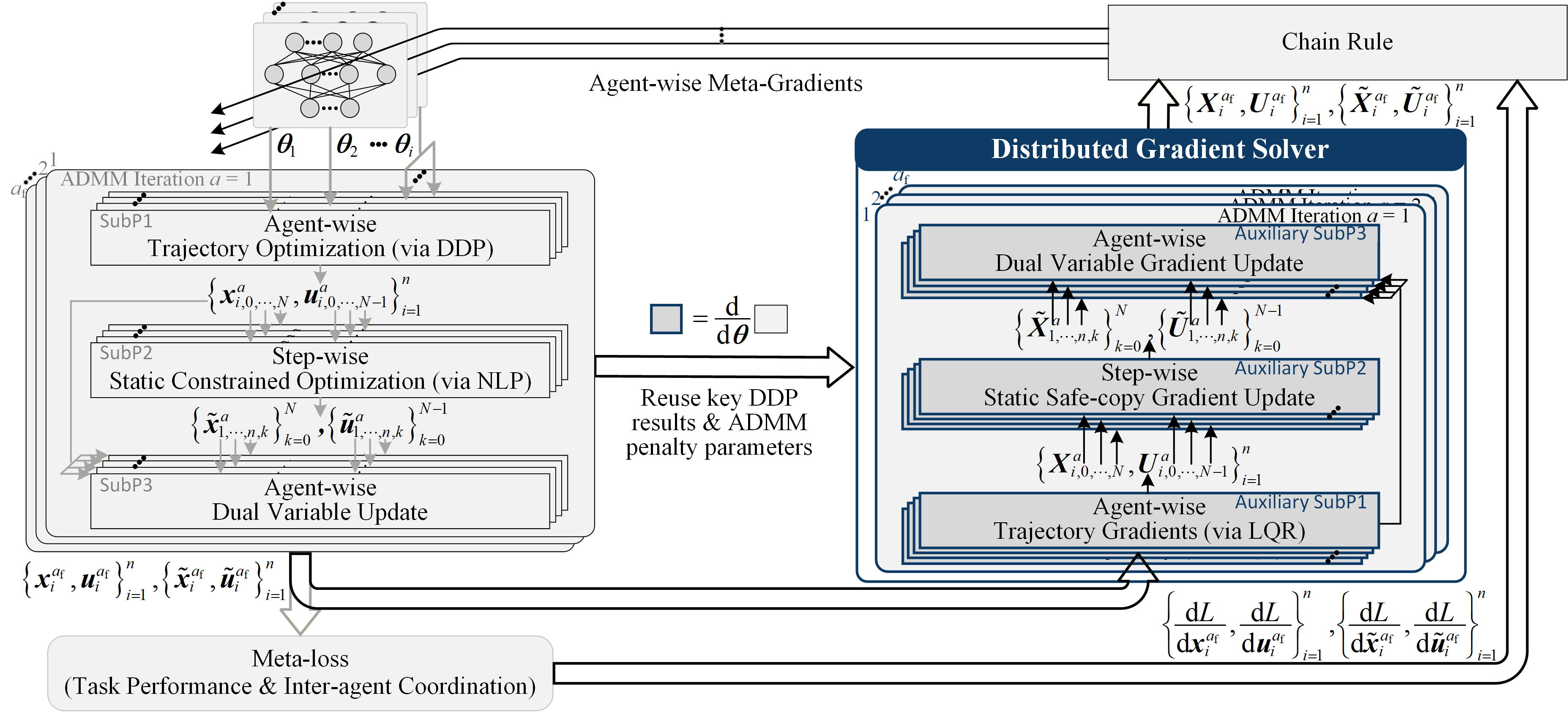}
    \caption{\footnotesize The forward (gray) and backward (black) passes of the DiffCoord framework. In the forward pass, the optimal trajectories are generated via the ADMM-DDP pipeline (Algorithm~\ref{alg: ADMM-DDP forward pass}) whose parameters are modeled by the agent-wise neural networks~\eqref{eq:agent_wise network} for task adaptation. In the backward pass, central to DiffCoord is the ADMM-LQR distributed gradient solver, which is constructed by differentiating the forward pass end-to-end to compute and coordinate the analytical meta-gradients. A key insight is that this solver mirrors the dynamic–static structure of the ADMM-DDP pipeline.}
    \label{fig:learning pipeline}
\end{figure*}

\subsection{Problem Statement}\label{subsec:problem statement}
For the ADMM-DDP pipeline (e.g., Algorithm~\ref{alg: ADMM-DDP forward pass}), when the ADMM fully converges (i.e., $a_{\rm f}\to \infty$), its solution is a fixed point of the ADMM update operator:
\begin{equation}
    \{\bm{\tau}_{i}^{\ast},\Tilde{\bm{\tau}}_{i}^{\ast}\}^n_{i=1}={\text{ADMM-DDP}}\left(\{\bm{\tau}_{i}^{\ast},\Tilde{\bm{\tau}}_{i}^{\ast}\}^n_{i=1}, \bm{\theta}^{\rm{p}}\right),
    \label{eq:fixed point admm}
\end{equation}
where $\{{\bm{\tau}_{i}^{\ast}}\}_{i=1}^{n}= \{\bm{x}_{i}^{\ast},\bm{u}_{i}^{\ast}\}_{i=1}^{n}$ and $\{\bm{\Tilde{\tau}}_{i}^{\ast}\}_{i=1}^{n}= \{\bm{\Tilde{x}}_{i}^{\ast},\bm{\Tilde{u}}_{i}^{\ast}\}_{i=1}^{n}$ are the optimal trajectories of the original and copied variables for all agents, and $\bm{\theta}^{\rm{p}}$ denotes the problem-level parameters. For example, consider a widely used quadratic local cost:
\begin{equation}
    \ell_{i,k}=\frac{1}{2} \left\lVert \bm{x}_{i,k}-\bm{x}_{i,k}^{\text{ref}} \right\rVert _{\bm{Q}_i}^2+\frac{1}{2}\left\lVert\bm{u}_{i,k}-\bm{u}_{i,k}^{\text{ref}}\right\rVert_{\bm{R}_i}^2,
    \label{eq:quadratic local cost}
\end{equation}
where $\bm{Q}_i  \succ 0$ and $\bm{R}_i \succ 0$ are positive definite weight matrices. The trajectories are shaped by the references $\bm{\theta}_i^{\text{r}}\coloneqq [\bm{x}_{i,k}^{\text{ref}}]_{k=0}^{N}$ and the weights $\bm{\theta}_i^{\text{w}}\coloneqq[\operatorname{vec}(\bm{Q}_i),\operatorname{vec}(\bm{R}_i)]$ that trade off reference tracking accuracy and control effort, where $\operatorname{vec} (\cdot)$ denotes the vectorization of a matrix. These are components of $\bm{\theta}^{\rm{p}}$. Other components may include system model parameters $\bm{\theta}^{\rm m}_i$ and constraint parameters $\bm{\theta}^{\rm c}_i$. In this fully converged regime, the trajectories are completely coordinated and governed mainly by $\bm{\theta}^{\rm{p}}$. Therefore, a natural tuning strategy is to tune the solver-level parameters $\bm{\theta}^{\rm{s}}$ (e.g., the ADMM penalty parameters) to ensure stable convergence, and subsequently tune $\bm{\theta}^{\rm{p}}$ to improve task performance. However, the ADMM penalty parameters that yield good convergence behavior may still vary across tasks, making them difficult to determine in advance and complicating such a separate tuning strategy.

Achieving full convergence of ADMM is often impractical under limited computational budgets. In practice, we typically truncate $a_{\rm f}$ to a finite number, which may be further reduced in long-horizon planning problems to maintain efficiency. In this case, $\bm{\theta}^{\rm{p}}$ and $\bm{\theta}^{\rm{s}}$ become tightly coupled, and the trajectories depend recursively on both through the following iterations:
\begin{equation}
    \{\bm{\tau}_{i}^a,\Tilde{\bm{\tau}}_{i}^a\}^n_{i=1}={\text{ADMM-DDP}}\left(\{\bm{\tau}_{i}^{a-1},\Tilde{\bm{\tau}}_{i}^{a-1}\}^n_{i=1}, \bm{\theta}^{\rm{p}},\bm{\theta}^{\rm{s}}\right), 
    \label{eq:truncated admm}
\end{equation}
where $a\in[1,a_{\rm f}]$. As shown in~\eqref{eq:DPP cost augmented by ADMM}, large $\rho_i^a$ and $\sigma_i^a$ can enforce the consensus but also degrade the agent-wise performance if not properly balanced against $\bm{\theta}_i^{\text{w}}$. This phenomenon has been observed in~\cite{10288223}, where the ADMM penalty parameters are manually initialized proportional to the DDP cost weights. This heuristic scaling reveals an intrinsic coupling between $\bm{\theta}^{\rm{s}}$ and $\bm{\theta}^{\rm{p}}$. This coupling makes the parameter tuning increasingly challenging as the number of agents grows or inter-agent interactions become stronger. In practice, manual tuning often relies on trial and error and tends to be inefficient and task-specific. Because system dynamics and environments vary across tasks, the parameters must adapt accordingly. To address this challenge, we propose DiffCoord, a unified framework that jointly meta-learns $\bm{\theta}^{\rm{p}}$ and $\bm{\theta}^{\rm{s}}$ to adapt across tasks.

\subsection{Neural Adaptive Parameters}\label{subsec:neural adaptive parameters}
We aggregate each agent's parameters into the vector $\bm{\theta}_i=[\bm{\theta}^{\rm{p}}_i,\bm{\theta}^{\rm{s}}_i]\in\mathbb{R}^{p_i}$. Let $\bm{\theta}=[\bm{\theta}_1,\dots,\bm{\theta}_{n}]\in\mathbb{R}^{\sum{p_i}}$ denote the total parameter vector for the entire multi-agent system. To efficiently realize the task-adaptation, we employ agent-wise neural networks to learn the task-to-parameter mapping, which is difficult to model from first principles:
\begin{equation}
    \bm{\theta}_i=\bm{\pi}_i(\bm{\chi}_i), \label{eq:agent_wise network}
\end{equation}
where $\bm{\chi}_i$ denotes task-specific features. Compared with a single large network for $\bm{\theta}$, the agent-wise design scales better as the number of agents increases. The same networks can also be shared among isomorphic agents to further improve scalability to varying agent counts.

In this work, we consider an open-loop ADMM penalty policy that adapts only to task variations and ADMM iterations. Open-loop policies have shown robust performance comparable to closed-loop policies that adapt to ADMM residuals~\cite{saravanosdeep}. Although closed-loop policies may perform better with more ADMM iterations, their dependence on residuals complicates the computational graph and the learning pipeline. In contrast, open-loop policies preserve important problem structure in the learning framework and enable the reuse of many computational results, improving learning efficiency.

% discuss open loop penalty policy here

\subsection{Meta-Learning Via Gradient Descent}\label{subsec:meta_learning via gradient}

The tuning problem becomes meta-learning of the network parameters $\{\bm{\varpi}_i\}_{i=1}^{n}$ that perform well on average across tasks. This yields the bi-level optimization problem:
\begin{subequations}
    \begin{align}
        \min_{\{\bm{\varpi}_i\}_{i=1}^{n}} & L_{\rm m}= \frac{1}{M}\sum_{t=1}^{M}L_{t}(\{\bm{\tau}_{i,t}^{a_{\rm f}},\bm{\Tilde{\tau}}_{i,t}^{a_{\rm f}}\}_{i=1}^{n})\label{eq:high_level loss}\\
        \text{s.t.}\quad
        & \bm{\theta}_{i,t}=\bm{\pi}_i(\bm{\chi}_{i,t}),\ \forall i\in\{1,\dots,n\}, \label{eq:neural adaptive parameters}\\
        & \{\bm{\tau}_{i,t}^{a_{\rm f}},\bm{\Tilde{\tau}}_{i,t}^{ a_{\rm f}}\}_{i=1}^{n}\ \text{generated by Algorithm \ref{alg: ADMM-DDP forward pass}},
    \end{align}
    \label{eq:bilevel optimization}%
\end{subequations}
where $(\cdot)_t$ denotes task $t$. The lower level solves a task-specific ADMM-DDP pipeline using Algorithm~\ref{alg: ADMM-DDP forward pass} with the corresponding parameters to generate agent trajectories, while the upper level optimizes a meta-loss $L_{\rm m}$ across tasks to learn the task-adaptive parameters~\footnote{This bi-level structure falls into the paradigm of \textit{control-oriented meta-learning}~\cite{richards2023control}, where the task adaptation is achieved through controller optimization rather than through online updates of network parameters as in traditional meta-learning frameworks.}. 

The task-specific loss $L_{t}$ captures both agent-wise performance and inter-agent coordination, where the latter is measured by ADMM residuals, i.e., $\left\| \bm{\tau}^{a_{\mathrm{f}}}_{i,t}-\Tilde{\bm{\tau}}^{a_{\mathrm{f}}}_{i,t}\right\|^2_2$. Minimizing the ADMM residuals in $L_t$ drives the ADMM-DDP pipeline towards its centralized counterpart, without requiring the centralized problem to be solved in advance. This differs from the training approach in~\cite{saravanosdeep}, which relies on a dataset of pre-solved centralized problems.

We solve problem \eqref{eq:bilevel optimization} via gradient descent. We compute the meta-gradients w.r.t. $\bm{\theta}$ using chain rule and then extract the relevant block for each agent $j\in\{1,\dots,n\}$:
\begin{equation}
    \frac{\mathrm{d} L_{\rm m}}{\mathrm{d} \bm{\varpi}_j}= \frac{1}{M}\sum_{t=1}^{M}\bigllbracket\sum_{i=1}^{n}\left( \frac{\partial L_t}{\partial \bm{\tau}_{i,t}^{a_{\rm f}}} \frac{\mathrm{d} \bm{\tau}_{i,t}^{a_{\rm f}}}{\mathrm{d} \bm{\theta}_t} + \frac{\partial L_t}{\partial \bm{\Tilde{\tau} }_{i,t}^{a_{\rm f}}} \frac{\mathrm{d} \bm{\Tilde{\tau} }_{i,t}^{a_{\rm f}}}{\mathrm{d} \bm{\theta}_t} \right) \bigrrbracket_j  \frac{\partial \bm{\theta}_{j,t}}{\partial \bm{\varpi}_j},
\label{eq:chain rule loss gradient2}
\end{equation}
where $\bigllbracket \cdot \bigrrbracket_j$ denotes $[\ldots, \bm{I}_{p_j}, \ldots]$ with $\bm{I}_{p_j} \in \mathbb{R}^{p_j \times p_j}$. In~\eqref{eq:chain rule loss gradient2}, $\{\frac{\partial L_{t}}{\partial \bm{\tau}_{i,t}^{a_{\rm f}}}, \frac{\partial L_{t}}{\partial \bm{\Tilde{\tau}}_{i,t}^{a_{\rm f}}}\}_{i=1}^n$ can be computed directly since $L_{t}$ is explicit in $\{\bm{\tau}_{i,t}^{a_{\rm f}},\bm{\Tilde{\tau}}_{i,t}^{a_{\rm f}}\}_{i=1}^n$. Computing $\{\frac{\partial \bm{\theta}_{j,t}}{\partial \bm{\varpi}_j}\}_{j=1}^n$ is standard and handled by automatic differentiation. The main challenge lies in $\{\frac{\mathrm{d}\bm{\tau}_{i,t}^{a_{\rm f}}}{\mathrm{d}\bm{\theta}_t}, \frac{\mathrm{d}\bm{\Tilde{\tau}}_{i,t}^{a_{\rm f}}}{\mathrm{d}\bm{\theta}_t}\}_{i=1}^n$. Because the trajectories are recursively updated through~\eqref{eq:truncated admm}, their gradients are obtained accordingly by differentiating through the iterations. This yields an end-to-end differentiable ADMM computational graph that naturally gives rise to a recursive gradient-computation scheme. Unlike single-agent settings~\cite{10.5555/3327757.3327922,NEURIPS2020_5a7b238b}, these trajectory gradients must also be coordinated through the ADMM structure, making their computation more challenging. The meta-learning pipeline of  DiffCoord is illustrated in Fig.~\ref{fig:learning pipeline}, where the forward pass solves the lower-level problem and the backward pass computes the meta-gradients for solving the upper-level problem.

\section{ADMM-LQR Distributed Gradient Solver}\label{section:analytical grad}
We aim to efficiently compute $\{\tfrac{\mathrm{d}\bm{\tau}_{i}^{  a_{\rm f}}}{\mathrm{d}\bm{\theta}}, \tfrac{\mathrm{d}\bm{\Tilde{\tau}}_{i}^{  a_{\rm f}}}{\mathrm{d}\bm{\theta}}\}_{i=1}^n$ (hereafter we drop $(\cdot)_t$ for simplicity). To this end, we differentiate the ADMM-DDP pipeline in Section~\ref{section:preliminaries} end-to-end, obtaining an auxiliary ADMM-LQR distributed gradient solver that inherits the computational structure, ADMM penalty parameters, and iteration budget of the forward pass (see Fig.~\ref{fig:learning pipeline}). This structural mirroring enables reuse of key computation results from the forward pass and parallelization over agents and along trajectory horizons.

\subsection{Efficient Gradient Computation Via DDP Reuse}\label{subsec:gradient DDP}
% We begin with Subsystem 1 (\ref{subsubp1}) to solve for the DDP gradients.
At ADMM iteration $a$, once DDP converges, each agent’s trajectory $\bm{\tau}_i^{  a}$ satisfies the first-order optimality conditions parameterized by $\bm{\theta}$. These conditions can be differentiated using the implicit function theorem (IFT) (Theorem 1B.1, \cite{dontchev2009implicit}). A direct IFT application to the full trajectory scales poorly with the horizon length $N$ and is impractical for long-horizon planning. The state-of-the-art methods such as PDP~\cite{NEURIPS2020_5a7b238b} and its variants~\cite{NEURIPS2021_85ea6fd7,liang2025onlinecontrolinformedlearning} instead compute the gradients recursively with linear complexity $\mathcal{O}(N)$. Our method preserves this $\mathcal{O}(N)$ complexity while further reducing runtime by exploiting computations already available from the DDP solve. Specifically, we show that the agent-wise trajectory gradients correspond to the optimal solution of an auxiliary matrix-valued LQR system. Its recursive solution overlaps structurally with the DDP solve, enabling reuse of key DDP computation results such as Riccati recursions, feedback gains, and related Hessian matrices.

When DDP converges to a local optimum, the following optimality condition holds theoretically:
\begin{equation}
    \mathbf{0} = \hat{Q}_{i,k}^{u,a}(\bm{x}_{i,k}^{  a},\bm{u}_{i,k}^{  a},\bm{\Tilde{u}}_{i,k}^{  a-1},\bm{\xi}_{i,k}^{a-1},\bm{\theta}).\label{eq:first order optimal cond_u}
\end{equation}
It follows from the standard DDP stopping criterion $\|\delta\bm{u}_{i,k}\| \le \epsilon$ (Section~\ref{section:preliminaries}), and can also be derived by differentiating~\eqref{eq:value function} w.r.t. $\bm{u}_{i,k}^{  a}$. Differentiating \eqref{eq:value function} w.r.t. $\bm{x}_{i,k}^{  a}$ yields the other optimality condition:
\begin{equation}
    \hat{V}_{i,k}^{x,a}=\hat{Q}_{i,k}^{x,a}(\bm{x}_{i,k}^{  a},\bm{u}_{i,k}^{  a},\bm{\Tilde{x}}_{i,k}^{  a-1},\bm{\nu}_{i,k}^{a-1},\bm{\theta}). \label{eq:first order optimal cond_x}
\end{equation}
Finally, the optimal state and control input satisfy the agent's dynamics~\eqref{eq:robot dynamics}:
\begin{equation}
    \bm{x}_{i,k+1}^{  a}=\bm{f}_{i}^{a}(\bm{x}_{i,k}^{  a},\bm{u}_{i,k}^{  a},\bm{\theta}). \label{eq:optimal dynamics}
\end{equation}
The equations~\eqref{eq:first order optimal cond_u},~\eqref{eq:first order optimal cond_x}, and \eqref{eq:optimal dynamics} constitute the first-order optimality conditions for Subsystem 1~\ref{subsubp1}. 

To solve for $\frac{\mathrm{d} \bm{\tau}_i^{  a}}{\mathrm{d} \bm{\theta}}$, we are motivated to implicitly differentiate both sides of \eqref{eq:first order optimal cond_x}, \eqref{eq:first order optimal cond_u}, and \eqref{eq:optimal dynamics} w.r.t. $\bm{\theta}$ using chain rule. This yields the following differential optimality conditions:

\begin{subequations}
    \begin{align}
        \frac{\mathrm{d} \hat{V}_{i,k}^{x,a}}{\mathrm{d} \bm{\theta}} &=\hat{Q}_{i,k}^{xx,a}\frac{\mathrm{d} {\bm{x}_{i,k}^{  a}}}{\mathrm{d} \bm{\theta}}+\hat{Q}_{i,k}^{xu,a}\frac{\mathrm{d} {\bm{u}_{i,k}^{  a}}}{\mathrm{d} \bm{\theta}}+\hat{Q}_{i,k}^{x\theta,a}, \label{eq:differential condition on x}\\
        \bm{0}& = \hat{Q}_{i,k}^{ux,a}\frac{\mathrm{d} {\bm{x}_{i,k}^{  a}}}{\mathrm{d} \bm{\theta}}+\hat{Q}_{i,k}^{uu,a}\frac{\mathrm{d} {\bm{u}_{i,k}^{  a}}}{\mathrm{d} \bm{\theta}}+\hat{Q}_{i,k}^{u\theta,a}, \label{eq:differential condition on u}\\
        \frac{\mathrm{d} \bm{x}_{i,k+1}^{  a}}{\mathrm{d} \bm{\theta}} &=f_{i,k}^{x,a}\frac{\mathrm{d} {\bm{x}_{i,k}^{  a}}}{\mathrm{d} \bm{\theta}}+f_{i,k}^{u,a}\frac{\mathrm{d} {\bm{u}_{i,k}^{  a}}}{\mathrm{d} \bm{\theta}}+f_{i,k}^{\theta,a},\label{eq:differential dynamics}
    \end{align}
    \label{eq:differential optimality conditions}%
\end{subequations}
where $\hat{Q}_{i,k}^{ux,a} = (\hat{Q}_{i,k}^{xu,a})^{\top}$ and $f_{i,k}^{\theta,a}\coloneqq\nabla_{\bm{\theta}}\bm{f}_{i,k}^{a}$. Note that $\hat{Q}_{i,k}^{x\theta,a}$ and $\hat{Q}_{i,k}^{u\theta,a}$ not only represent the explicit partial derivatives of $\hat{Q}_{i,k}^{x,a}$ and $\hat{Q}_{i,k}^{u,a}$ w.r.t. $\bm{\theta}$, but also include the gradients of $\bm{\Tilde{x}}_{i,k}^{  a-1}$, $\bm{\Tilde{u}}_{i,k}^{a-1}$, $\bm{\nu}_{i,k}^{a-1}$, and $\bm{\xi}_{i,k}^{a-1}$ w.r.t. $\bm{\theta}$. This leads to
\begin{subequations}
    \begin{align}
        \hat{Q}_{i,k}^{x\theta,a} &= \nabla_{\bm{\theta}}\hat{Q}_{i,k}^{x,a}-\rho_i\frac{\mathrm{d} {\bm{\Tilde{x}}_{i,k}^{  a-1}}}{\mathrm{d} \bm{\theta}}+\frac{\mathrm{d} {\bm{\nu}_{i,k}^{a-1}}}{\mathrm{d} \bm{\theta}},\label{eq:differential x lambda}\\
        \hat{Q}_{i,k}^{u\theta,a} &= \nabla_{\bm{\theta}}\hat{Q}_{i,k}^{u,a}-\sigma_i\frac{\mathrm{d} {\bm{\Tilde{u}}_{i,k}^{  a-1}}}{\mathrm{d} \bm{\theta}}+\frac{\mathrm{d} {\bm{\xi}_{i,k}^{a-1}}}{\mathrm{d} \bm{\theta}}.\label{eq:differential u xi}
    \end{align}
    \label{eq:gradient couplings in subsystem1}%
\end{subequations}
The gradients $ \frac{\mathrm{d} {\bm{\Tilde{\tau}}_{i}^{  a-1}}}{\mathrm{d} \bm{\theta}}$ and $\{\frac{\mathrm{d} {\bm{\nu}_{i,0:N}^{a-1}}}{\mathrm{d} \bm{\theta}}, \frac{\mathrm{d} {\bm{\xi}_{i,0:N-1}^{a-1}}}{\mathrm{d} \bm{\theta}}\}$ are computed in Subsystems 2 and 3 at the previous ADMM iteration. Hence, the gradient computations across subsystems are inherently coupled, posing a major challenge and distinguishing the multi-agent case from its single-agent counterpart~\footnote{Moreover, these gradient coupling terms (shown in~\eqref{eq:gradient couplings in subsystem1} and later in~\eqref{eq:gradient couplings auxiliary subproblem2}) prevent the use of VJP methods, which do not provide such matrix-valued trajectory gradients required for distributed coordination across agents.}.

We refer to the conditions \eqref{eq:differential condition on x}, \eqref{eq:differential condition on u}, and \eqref{eq:differential dynamics} as the differential Bellman's principle (DBP) conditions. We show that the DBP conditions are equivalent to the differential PMP conditions used to construct an auxiliary LQR system for recursive gradient computation (Lemma 5.1 in~\cite{NEURIPS2020_5a7b238b}). Moreover, the same LQR system can also be derived from DBP, as formalized in the following lemma.

\begin{lemma}
    \label{lm:DBP and PDP equivalence} 
    Let $\hat{\mathcal{H}}_{i,k}^{a}=\hat{\ell}_{i,k}^{a}+(\hat{\bm{\lambda}}_{i,k+1}^a)^{\top}\bm{f}_{i,k}^{a}$ define the augmented Hamiltonian of~\eqref{eq:subproblem1}, where $\hat{\bm{\lambda}}_{i,k}^{a}\in\mathbb{R}^{n}$ is the co-state. Under the relation $\hat{\bm{\lambda}}_{i,k}^a =\hat{V}_{i,k}^{x,a}$, these DBP conditions~\eqref{eq:differential optimality conditions} are equivalent to the differential PMP conditions:
    
    \begin{subequations}
        \begin{align}
            \begin{split}
                \frac{\mathrm{d}\hat{\bm{\lambda}}_{i,k}^{  a}}{\mathrm{d}\hat{\bm{\theta}}}&=\hat{H}_{i,k}^{xx,a}\frac{\mathrm{d} {\bm{x}_{i,k}^{  a}}}{\mathrm{d} \bm{\theta}}+\hat{H}_{i,k}^{xu,a}\frac{\mathrm{d} {\bm{u}_{i,k}^{  a}}}{\mathrm{d} \bm{\theta}}+\hat{H}_{i,k}^{x\theta,a}\\
                &\quad +(f_{i,k}^{x,a})^{\top}\frac{\mathrm{d}\hat{\bm{\lambda}}_{i,k+1}^{  a}}{\mathrm{d}\bm{\theta}},
            \end{split}
            \label{eq:differential PMP on x lemma1}\\
            \begin{split}
                \bm{0}&=\hat{H}_{i,k}^{ux,a}\frac{\mathrm{d} {\bm{x}_{i,k}^{  a}}}{\mathrm{d} \bm{\theta}}+\hat{H}_{i,k}^{uu,a}\frac{\mathrm{d} {\bm{u}_{i,k}^{  a}}}{\mathrm{d} \bm{\theta}}+\hat{H}_{i,k}^{u\theta,a}\\
                &\quad +(f_{i,k}^{u,a})^{\top}\frac{\mathrm{d}\hat{\bm{\lambda}}_{i,k+1}^{  a}}{\mathrm{d}\bm{\theta}},
            \end{split}
            \label{eq:differential PMP on u lemma1}\\
            \frac{\mathrm{d} \bm{x}_{i,k+1}^{  a}}{\mathrm{d} \bm{\theta}} &=f_{i,k}^{x,a}\frac{\mathrm{d} {\bm{x}_{i,k}^{  a}}}{\mathrm{d} \bm{\theta}}+f_{i,k}^{u,a}\frac{\mathrm{d} {\bm{u}_{i,k}^{  a}}}{\mathrm{d} \bm{\theta}}+f_{i,k}^{\theta,a}.\label{eq:differential dynamics PMP lemma1}
        \end{align}
        \label{eq:differential PMP conditions lemma1}%
    \end{subequations}
    % Hence, the DBP and differential PMP conditions yield identical gradients for the optimal trajectories of Problem~\eqref{eq:a general dynamic optimization in lemma1}. 
    where $\hat{H}^{(\cdot)(\cdot),a}_{i,k}\coloneqq \frac{\partial^2\hat{\cal H}^a_{i,k}}{\partial(\cdot)\partial(\cdot)}$ denotes the Hessian of the Hamiltonian, and
    \begin{subequations}
    \begin{align}
        \hat{H}_{i,k}^{x\theta,a} &= \nabla_{\bm{\theta}}\hat{H}_{i,k}^{x,a}-\rho_i\frac{\mathrm{d} {\bm{\Tilde{x}}_{i,k}^{  a-1}}}{\mathrm{d} \bm{\theta}}+\frac{\mathrm{d} {\bm{\nu}_{i,k}^{a-1}}}{\mathrm{d} \bm{\theta}},\label{eq:differential x lambda}\\
        \hat{H}_{i,k}^{u\theta,a} &= \nabla_{\bm{\theta}}\hat{H}_{i,k}^{u,a}-\sigma_i\frac{\mathrm{d} {\bm{\Tilde{u}}_{i,k}^{  a-1}}}{\mathrm{d} \bm{\theta}}+\frac{\mathrm{d} {\bm{\xi}_{i,k}^{a-1}}}{\mathrm{d} \bm{\theta}}.\label{eq:differential u xi}
    \end{align}
    \label{eq:gradient couplings in differential PMP}%
\end{subequations}
From~\eqref{eq:differential PMP conditions lemma1}, the following matrix-valued LQR system can be constructed:
    
    \begin{equation}
        \begin{split}
            \min_{\{ \bm{X}_i,\bm{U}_i\}^{a}}& \sum_{k=0}^{N-1}\operatorname{tr} \left(\frac{1}{2}\begin{bmatrix}
                \bm{X}_{i,k}^{a}\\
                \bm{U}_{i,k}^{a}
            \end{bmatrix}^{\top}
            \underbrace{\begin{bmatrix}
                \hat{H}_{i,k}^{xx,a},\hat{H}_{i,k}^{xu,a}\\
                \hat{H}_{i,k}^{ux,a},\hat{H}_{i,k}^{uu,a}
            \end{bmatrix}}_{\hat{H}^{\tau\tau,a}_{i,k}}
            \begin{bmatrix}
                \bm{X}_{i,k}^{a}\\
                \bm{U}_{i,k}^{a}
            \end{bmatrix} \right.\\
            &\qquad \qquad \quad \left. +\begin{bmatrix}
                \hat{H}_{i,k}^{x\theta,a}\\
                \hat{H}_{i,k}^{u\theta,a}
            \end{bmatrix}^{\top}\begin{bmatrix}
                \bm{X}_{i,k}^{a}\\
                \bm{U}_{i,k}^{a}
            \end{bmatrix} \right)\\
            &\ +\operatorname{tr}\left(\frac{1}{2}\left(\bm{X}_{i,N}^{a}\right)^{\top}\hat{H}_{i,N}^{xx,a}\bm{X}_{i,N}^{a}+\left(\hat{H}_{i,N}^{x\theta,a}\right)^{\top}\bm{X}_{i,N}^{a} \right)\\
            \text{s.t.}\quad
            & \bm{X}_{i,k+1}^{a}=f_{i,k}^{x,a}\bm{X}_{i,k}^{a}+f_{i,k}^{u,a}\bm{U}_{i,k}^{a}+f_{i,k}^{\theta,a},\ \bm{X}_{i,0}^{a}=\bm{0},
        \end{split}
        \label{eq:same LQR system in lemma1}
    \end{equation}
    where $\bm{X}_{i,k}^{a}\in\mathbb{R}^{n\times p}$ and $\bm{U}_{i,k}^{a}\in \mathbb{R}^{m\times p}$ are the matrix-valued state and control, respectively, and $\operatorname{tr}(\cdot)$ is the matrix trace operator. Let $\{\bm{X}_{i,0:N}^{a},\bm{U}_{i,0:N-1}^{a} \}$ be the optimal solution to the LQR system~\eqref{eq:same LQR system in lemma1}. They satisfy Bellman's principle of~\eqref{eq:same LQR system in lemma1}, which coincides with the DBP conditions~\eqref{eq:differential optimality conditions}. Consequently,  
    \begin{equation}
        \{\bm{X}_{i,0:N}^{a},\bm{U}_{i,0:N-1}^{a}\}= \left\{ \frac{\mathrm{d}\bm{x}_{i,0:N}^{a}}{\mathrm{d}\bm{\theta}}, \frac{\mathrm{d}\bm{u}_{i,0:N-1}^{a}}{\mathrm{d}\bm{\theta}}\right\}.
        \label{eq:equivalence between the optimal solutions and the gradients}
    \end{equation}
\end{lemma}
\begin{proof}
    See Appendix-\ref{appendix:proof of lemma1}.
\end{proof}

Lemma~\ref{lm:DBP and PDP equivalence} shows that the gradients of DDP trajectories w.r.t. $\bm\theta$ can be obtained analytically by solving~\eqref{eq:same LQR system in lemma1}. Importantly, the Hessians $\{\hat{H}^{xx,a}_{i,k}\}_{k=0}^{N}$, $\{\hat{H}^{xu,a}_{i,k}\}_{k=0}^{N-1}$, and $\{\hat{H}^{uu,a}_{i,k}\}_{k=0}^{N-1}$, together with the Jacobians $\{f^{x,a}_{i,k},f^{u,a}_{i,k}\}_{k=0}^{N-1}$, are already available from the DDP derivatives~\eqref{eq:derivatives in approx Q} and can be reused to construct~\eqref{eq:same LQR system in lemma1}. While PDP~\cite{NEURIPS2020_5a7b238b} applies the Pontryagin Maximum Principle (PMP) to obtain a recursive analytical solution of~\eqref{eq:same LQR system in lemma1}, we derive the recursive solution from Bellman's principle, which aligns with the DDP solve. This retains the $\mathcal{O}(N)$ complexity while enabling the reuse of key DDP results, as formalized in the following lemma.

% Unlike PDP~\cite{NEURIPS2020_5a7b238b} that solves~\eqref{eq:same LQR system in lemma1} via the Pontryagin Maximum Principle (PMP), we exploit Bellman’s principle to preserve $\mathcal{O}(N)$. Moreover, by reusing key DDP results, our approach achieves higher efficiency in the gradient computation, as formalized in the following lemma.

\begin{lemma}
\label{lm:DDP gradient solver}
    Let $\{\hat{Q}^{uu,a}_{i,k},\hat{Q}^{xu,a}_{i,k},f^{x,a}_{i,k},f^{u,a}_{k},\hat{\bm{K}}^{a}_{i,k},\hat{V}^{xx,a}_{i,k+1}\}_{k=0}^{N-1}$ denote the quantities computed during the DDP solve for Problem~\eqref{eq:subproblem1}. The following Riccati equation can be solved efficiently backward in time from $k=N-1$ to $k=0$:
    \begin{equation}
            \hat{V}^{x\theta,a}_{i,k}=\hat{Q}^{x\theta}_{i,k}-\hat{Q}^{xu,a}_{i,k}(\hat{Q}^{uu,a}_{i,k})^{-1}\hat{Q}^{u\theta}_{i,k},
        \label{eq:Riccati equation in lemma2}
    \end{equation}
    with the terminal condition $V^{x\theta,a}_{i,N}=\hat{H}^{x\theta,a}_{i,N}=\hat{\ell}^{x\theta,a}_{i,N}$, where $\hat{Q}^{x\theta}_{i,k}=\hat{H}^{x\theta,a}_{i,k}+(f_{i,k}^{x,a})^{\top}\hat{V}^{x\theta,a}_{i,k+1}+(f_{i,k}^{x,a})^{\top}\hat{V}^{xx,a}_{i,k+1}f^{\theta,a}_{i,k}$ and $\hat{Q}^{u\theta}_{i,k}=\hat{H}^{u\theta,a}_{i,k}+(f^{u,a}_{i,k})^{  \top}\hat{V}^{x\theta,a}_{i,k+1}+(f^{u,a}_{i,k})^{\top}\hat{V}^{xx,a}_{i,k+1}f^{\theta,a}_{i,k}$. Define the feedforward gain
    \begin{equation}
            \hat{\bm{K}}_{i,k}^{\mathrm{ff},a}\coloneqq -(\hat{Q}^{uu,a}_{i,k})^{-1}\hat{Q}^{u\theta}_{i,k}.
    \label{eq:feedforward gain in gradient computation}
    \end{equation}
    Then, the gradient trajectories $\{\bm{X}_{i,0:N},\bm{U}_{i,0:N-1}\}^a$ can be computed by the following forward recursion from $k=0$ to $k=N-1$, with the initial condition $\bm{X}_{i,0}=\bm{0}$:
     \begin{subequations}
         \begin{align}
             \bm{U}_{i,k}^{a}&=\hat{\bm{K}}_{i,k}^a\bm{X}_{i,k}^{a}+\hat{\bm{K}}_{i,k}^{\mathrm{ff},a}, \label{eq:optimal control law in lemma2}\\
             \bm{X}_{i,k+1}^{a}&=f_{i,k}^{x,a}\bm{X}_{i,k}^{a}+f_{i,k}^{u,a}\bm{U}_{i,k}^{a}+f_{i,k}^{\theta,a}.
         \end{align}
         \label{eq:gradient dynamics lemma2}
     \end{subequations}
\end{lemma}
\begin{proof}
    See Appendix-\ref{appendix:proof of lemma2}.
\end{proof}

The matrix-valued control law~\eqref{eq:optimal control law in lemma2} shares a similar structure with the DDP control law~\eqref{eq:DDP control law}, where the feedback gain $\hat{\bm{K}}_{i,k}^a$ can contribute to stabilizing the gradient computation. Moreover, the inverse of $\hat{Q}^{uu,a}_{i,k}$ in~\eqref{eq:feedforward gain in gradient computation} is already computed during the DDP solve and can therefore be efficiently reused, avoiding redundant and costly matrix inversions. 

\begin{remark}
    By simply dropping the cross-subsystem gradient contributions in~\eqref{eq:gradient couplings in differential PMP}, Lemma~\ref{lm:DBP and PDP equivalence} and~\ref{lm:DDP gradient solver} reduce to the single-agent case and naturally recover the analytical gradient computation for a general dynamic optimization problem.
\end{remark}

\subsection{Gradient Coordination Via ADMM}\label{subsec:gradient ADMM}
This subsection constructs an auxiliary ADMM pipeline to enforce the gradient-matching constraints $\{\frac{\mathrm{d} \bm{\tau}_i^a}{\mathrm{d} \bm{\theta}}=\frac{\mathrm{d} \bm{\Tilde{\tau}}_i^a}{\mathrm{d}\bm{\theta}}\}_{i=1}^n$, arising from the \textit{safe-copy-variable} strategy~\eqref{eq:consensus constraint}. Each ADMM iteration mirrors the forward-pass pipeline and consists of three subsystems that update and coordinate $ \frac{\mathrm{d} \bm{\tau}_i^{  a}}{\mathrm{d} \bm{\theta}}$, $ \frac{\mathrm{d} \bm{\Tilde{\tau}}_i^{  a}}{\mathrm{d} \bm{\theta}}$, $\frac{\mathrm{d} \bm{\nu}_i^{a}}{\mathrm{d} \bm{\theta}}$, and $\frac{\mathrm{d} \bm{\xi}_i^{a}}{\mathrm{d} \bm{\theta}}$.

\subsubsection{Auxiliary Subsystem 1} Each agent independently obtains its trajectory gradients $\{\bm{X}_{i}^{a},\bm{U}_{i}^{a} \}$  by solving the matrix-valued LQR subproblem~\eqref{eq:same LQR system in lemma1} using Lemma~\ref{lm:DDP gradient solver}.

\subsubsection{Auxiliary Subsystem 2} All agents compute the gradients of their safe-copy variables $\{\frac{\mathrm{d}\bm{\Tilde{x}}_{k}^a}{\mathrm{d}\bm{\theta}},\frac{\mathrm{d}\bm{\Tilde{u}}_{k}^a}{\mathrm{d}\bm{\theta}} \}$ by implicitly differentiating the static optimization problems~\eqref{eq:stage subproblem 2} and~\eqref{eq:terminal subproblem 2}. However, applying the IFT to constrained nonlinear optimization is nontrivial for two reasons: 1) identifying active inequalities (i.e., $g_{i,k}^a=0$, $g_{ij,k}^a=0$) is difficult; and 2) discontinuous switching between active and inactive constraints can destabilize gradient computation. To address this challenge, we approximate all the constraints in Subsystem 2 as soft constraints using logarithmic barrier functions, rendering the resulting optimality conditions differentiable. This technique, explored in Safe-PDP, has shown stable and effective learning in MPC~\cite{NEURIPS2021_85ea6fd7}. The approximation is used only for the gradient computation.

Define the approximated Lagrangian $\hat{\mathcal{L}}_k$ of Subproblem~\eqref{eq:stage subproblem 2} under the soft-constraint formulation as
\begin{equation}
    \begin{split}
        \hat{\mathcal{L}}_k^a&=\sum_{i=1}^n \left(e_{x_i,k}^a + e_{u_i,k}^a\right)+ \frac{1}{2\mu}\sum_{i=1}^n\sum_{j\ne i}^n(h_{ij,k}^a)^2  \\
        &\quad  -\mu\sum_{i=1}^n \Bigl[\ln \left(-g_{i,k}^a \right) + \sum_{j\ne i}^n\ln\left( -g_{ij,k}^a\right) \Bigl] ,
    \end{split}
    \label{eq:approximated Lagrangian of subproblem 2}
\end{equation}
where $\mu\in\mathbb{R}_+$ is the barrier parameter. After applying the IFT to~\eqref{eq:approximated Lagrangian of subproblem 2}, $\{\frac{\mathrm{d}\bm{\Tilde{x}}_{k}}{\mathrm{d}\bm{\theta}},\frac{\mathrm{d}\bm{\Tilde{u}}_{k}}{\mathrm{d}\bm{\theta}} \}^a$ can be approximated by the optimal solutions $\Tilde{\bm{Z}}^a_k=[\bm{\Tilde{X}}_k^{a};\bm{\Tilde{U}}_k^{a} ]$ of the following unconstrained matrix-valued static optimization subproblems. The stage-wise subproblem at $k\in[0,N-1]$ is defined by:
\begin{equation}
        \min_{\Tilde{\bm{Z}}^a_k} \operatorname{tr} \Bigl(\frac{1}{2}( \Tilde{\bm{Z}}^a_k)^{\top}
        \underbrace{\begin{bmatrix}
                \hat{L}^{{\Tilde{x}}{\Tilde{x}},a}_{k},\hat{L}^{\Tilde{x}\Tilde{u},a}_{k}\\
                \hat{L}^{\Tilde{u}\Tilde{x},a}_{k},\hat{L}^{\Tilde{u}\Tilde{u},a}_{k}
        \end{bmatrix}}_{\hat{L}^{\hat{\tau}\hat{\tau},a}_{k}}
        \Tilde{\bm{Z}}^a_k  +\begin{bmatrix}
                \hat{L}^{\Tilde{x}\theta,a}_{k}\\
                \hat{L}^{\Tilde{u}\theta,a}_{k}
        \end{bmatrix}^{\top}\Tilde{\bm{Z}}^a_k \Bigl),
    \label{eq:matrix-valued static optimization subproblem2}
\end{equation}
where $\hat{L}^{(\cdot)(\cdot),a}_{k}\coloneqq \frac{\partial^2\hat{\mathcal{L}}_k^a}{\partial(\cdot)\partial(\cdot)}$ denotes the Hessians of $\hat{\mathcal{L}}_k^a$. Accordingly, the terminal subproblem at $k=N$ is defined by:
\begin{equation}
    \min_{\bm{\Tilde{X}}_{N}^a} \operatorname{tr} \Bigl ( \frac{1}{2}(\bm{\Tilde{X}}_{N}^a)^{\top}\hat{L}^{\Tilde{x}\Tilde{x},a}_{N}\bm{\Tilde{X}}_{N}^a + \left(\hat{L}^{\Tilde{x}\theta,a}_{N}\right)^{\top}\bm{\Tilde{X}}_{N}^a\Bigl).\label{eq:ternimal matrix-valued static optimization subproblem2}
\end{equation}
In~\eqref{eq:matrix-valued static optimization subproblem2} and~\eqref{eq:ternimal matrix-valued static optimization subproblem2}, the coefficients of the linear terms account for the gradient couplings and are defined by:
\begin{subequations}
    \begin{align} 
    \hat{L}^{\Tilde{x}\theta,a}_{k}&=\nabla_{\bm{\theta}}\hat{L}^{\Tilde{x},a}_{k}-\operatorname{Blkdiag} \Bigl[\rho_i^a\bm{I}_{n_i} \Bigl]_{i=1}^n \bm{X}_k^a-\frac{\mathrm{d}\bm{\nu}_k^{a-1}}{\mathrm{d}\bm{\theta}},
    \label{eq:L_x_theta auxiliary subproblem2}\\
        \hat{L}^{\Tilde{u}{\theta},a}_{k}&=\nabla_{\bm{\theta}}\hat{L}^{{\Tilde{u}},a}_{k}-\operatorname{Blkdiag} \Bigl[\sigma_i^a\bm{I}_{m_i}\Bigl]_{i=1}^n\bm{U}_k^a-\frac{\mathrm{d}\bm{\xi}_k^{a-1}}{\mathrm{d}\bm{\theta}}.\label{eq:L_u_theta auxiliary subproblem2}
    \end{align}
    \label{eq:gradient couplings auxiliary subproblem2}%
\end{subequations}
As $\mu \to 0$, we can recover $\bm{\Tilde{X}}_{k}^{  a}\to \frac{\mathrm{d}\bm{\Tilde{x}}_{k}^{  a}}{\mathrm{d}\bm{\theta}}$ and $\bm{\Tilde{U}}_{k}^{  a}\to \frac{\mathrm{d}\bm{\Tilde{u}}_{k}^{  a}}{\mathrm{d}\bm{\theta}}$. 

% (Theorem 2 in~\cite{NEURIPS2021_85ea6fd7}).
\begin{remark}\label{remark:decentralized auxiliary subproblem2}
    At each step, solving Auxiliary Subproblem 2 incurs up to $\mathcal{O}(n^3)$ computational complexity due to the inversion of $\hat{L}^{\hat{\tau}\hat{\tau},a}_{k}$, whose dimensions scale linearly with the total number of agents $n$. Because these step-wise optimizations are static and can be fully parallelized across the planning horizon $N$, this parallel implementation bypasses sequential wall-clock bottlenecks and yields massive speedups over centralized gradient solvers for dynamic optimization. Furthermore, the $\mathcal{O}(n^3)$ inversion bottleneck can be eliminated by adopting the fully decentralized local safe-copy formulation detailed in Remark~\ref{remark:decentralized subproblem2}. In that case, each agent only needs to copy the variables of a fixed, bounded number of local neighbors, and the sizes of the resulting local quadratic coefficient matrices become independent of $n$.
\end{remark}

\subsubsection{Auxiliary Subsystem 3} At each step, each agent independently updates the gradients of its dual variables, $\frac{\mathrm{d}\bm{\nu}_{i,k}^{a}}{\mathrm{d}\bm{\theta}}$ and $\frac{\mathrm{d}\bm{\xi}_{i,k}^{a}}{\mathrm{d}\bm{\theta}}$. Differentiating both sides of~\eqref{eq:subproblem 3} w.r.t. $\bm{\theta}$ yields:
\begin{subequations}
    \begin{align}
        \frac{\mathrm{d}\bm{\nu}_{i,k}^a}{\mathrm{d}\bm{\theta}}&=\frac{\mathrm{d}\bm{\nu}_{i,k}^{a-1}}{\mathrm{d}\bm{\theta}}+\rho_i^a\left(\bm{X}_{i,k}^a-\bm{\Tilde{X}}_{i,k}^a\right) + \nabla_{\bm{\theta}}\bm{\nu}_{i,k}^a, \label{eq:gradient dual1 subsystem 3}\\
        \frac{\mathrm{d}\bm{\xi}_{i,k}^a}{\mathrm{d}\bm{\theta}}&=\frac{\mathrm{d}\bm{\xi}_{i,k}^{a-1}}{\mathrm{d}\bm{\theta}}+\sigma_i^a\left(\bm{U}_{i,k}^a-\bm{\Tilde{U}}_{i,k}^a\right)  + \nabla_{\bm{\theta}} \bm{\xi}_{i,k}^a,
       \label{eq:gradient dual2 subsystem 3}
    \end{align}
    \label{eq:dual gradient subsystem3}%
\end{subequations}
where $\nabla_{\bm{\theta}}\bm{\nu}_{i,k}^a$ and $\nabla_{\bm{\theta}} \bm{\xi}_{i,k}^a$ represent the explicit partial derivatives of~\eqref{eq:subproblem 3} w.r.t. $\bm \theta$~\footnote{They can be interpreted as perturbations to the dual updates. Similar perturbed dual forms appear in ADMM pipelines, e.g., for enhancing privacy~\cite{8772211}. In~\eqref{eq:dual gradient subsystem3}, these perturbations are proportional to the ADMM residuals:
$\nabla_{\bm{\theta}}\bm{\nu}_{i,k}^a=\nabla_{\bm{\theta}}\!\left[ \rho_i^a(\bm{x}_{i,k}^a-\bm{\Tilde{x}}_{i,k}^a)\right]$ and 
$\nabla_{\bm{\theta}} \bm{\xi}_{i,k}^a=\nabla_{\bm{\theta}}\!\left[ \sigma_i^a(\bm{u}_{i,k}^a-\bm{\Tilde{u}}_{i,k}^a)\right]$.
Hence, they remain bounded as the ADMM iterations progress.}. 

Finally, Algorithm~\ref{alg: ADMM-DDP backward pass} summarizes the auxiliary ADMM-LQR system for the distributed gradient computation, which mirrors the three-subsystem structure of the forward ADMM-DDP pipeline, including its dynamic–static decomposition, and preserves the same iteration budget $a_{\rm f}$.

\begin{algorithm}
\caption{Distributed Gradient Solver}
\label{alg: ADMM-DDP backward pass}
\SetKwInput{Input}{Input}
\SetKwInput{Output}{Output}
\SetKwComment{Comment}{$\triangleright$\ }{}
\SetNoFillComment
\DontPrintSemicolon
\Input{$\left\{  \{{\rho}_i^{a},{{\sigma}}_i^{a}\}_{i=1}^{n} \right\}_{a=1}^{a_{\rm{f}}}$,$\{\bm{\Tilde{X}}_i^{0}=\bm 0,\bm{\Tilde{U}}_i^{  0} =\bm 0\}_{i=1}^n$, $\{\frac{\mathrm{d}\bm{\nu}_i^{0}}{\mathrm{d}\bm{\theta}}=\bm 0,\frac{\mathrm{d}\bm{\xi}_i^{0}}{\mathrm{d}\bm{\theta}}=\bm 0\}_{i=1}^{n}$,$\left\{  \{\bm{\tau}_i^{a},\bm{\Tilde{\tau}}_i^{a}\}_{i=1}^{n} \right\}_{a=1}^{a_{\rm{f}}}$, $\left\{  \{\bm{\nu}_i^{a},\bm{\xi}_i^{a}\}_{i=1}^{n} \right\}_{a=1}^{a_{\rm{f}}}$, and the DDP components.} 

\For{$a \leftarrow 1$ \KwTo $a_{\rm f}$}{
\tcc*[l]{Auxiliary Subproblem 1}
\For(\Comment*[h]{in parallel}) {$i \leftarrow 1$ \KwTo $n$}{
$\{ \hat{H}_{i}^{x\theta,a},\hat{H}_{i}^{u\theta,a}\}\gets$ Solve~\eqref{eq:gradient couplings in differential PMP};\\
$\{\bm{X}_i^{a},\bm{U}_i^{a} \}\gets$ Solve~\eqref{eq:same LQR system in lemma1} using Lemma~\ref{lm:DDP gradient solver}; 
}
\tcc*[l]{Auxiliary Subproblem 2}
\For(\Comment*[h]{in parallel})  {$k \leftarrow 0$ \KwTo $N$}{
$\{\hat{L}^{\Tilde{x}\theta,a}_{k},\hat{L}^{\Tilde{u}\theta,a}_{k}\}\gets$ Solve~\eqref{eq:gradient couplings auxiliary subproblem2} and compute $\hat{L}^{{\Tilde{x}}{\Tilde{x}},a}_{k}$, $\hat{L}^{{\Tilde{x}}{\Tilde{u}},a}_{k}$, and $\hat{L}^{{\Tilde{u}}{\Tilde{u}},a}_{k}$;\\
$\{\bm{\Tilde{X}}_k^{a},\bm{\Tilde{U}}_k^{a} \}\gets$ Solve~\eqref{eq:matrix-valued static optimization subproblem2}; \Comment{$k<N$}
$\hat{L}^{\Tilde{x}\theta,a}_{N}\gets$ Solve~\eqref{eq:L_x_theta auxiliary subproblem2} and compute $\hat{L}^{\Tilde{x}\Tilde{x},a}_{N}$;\\
$\bm{\Tilde{X}}_N^{  a}\gets$ Solve \eqref{eq:ternimal matrix-valued static optimization subproblem2};}

\tcc*[l]{Auxiliary Subproblem 3}
\For(\Comment*[h]{in parallel})  {$i \leftarrow 1$ \KwTo $n$}
{\For(\Comment*[h]{in parallel})  {$k \leftarrow 0$ \KwTo $N$}
{$\{\frac{\mathrm{d}\bm{\nu}_{i,k}^{a}}{\mathrm{d}\bm{\theta}},\frac{\mathrm{d}\bm{\xi}_{i,k}^{a}}{\mathrm{d}\bm{\theta}}\}\gets$ Solve~\eqref{eq:dual gradient subsystem3}; \Comment{$k<N$}
$\frac{\mathrm{d}\bm{\nu}_{i,N}^{a}}{\mathrm{d}\bm{\theta}}\gets$ Solve \eqref{eq:gradient dual1 subsystem 3};
}
}
}
\Output{ $\{\bm{X}_i^{a_{\mathrm{f}}},\bm{U}_i^{a_{\mathrm{f}}}\}_{i=1}^{n}$ and $\{\bm{\Tilde{X}}_i^{a_{\mathrm{f}}},\bm{\Tilde{U}}_i^{a_{\mathrm{f}}}\}_{i=1}^{n}$.}
\end{algorithm}

\subsection{Theoretical Guarantees}\label{subsec theoretical guarantee}
The distributed gradient solver admits several theoretical guarantees that guide practical implementation. 

\textit{Centralized View of the Distributed Solver.} At each ADMM iteration, Algorithm~\ref{alg: ADMM-DDP backward pass} can be interpreted as a distributed approximation to a centralized auxiliary optimal control system:
    \begin{equation}
        \begin{split}
            \min_{\{\bm{X},\bm{U},\bm{\Tilde{X}},\bm{\Tilde{U}}\}^a} & \sum_{i=1}^{n}\bar{J}_i\left(\bm{X}_i^a,\bm{U}_i^a \right) + \sum_{k=0}^{N-1}\bar{J}_k(\bm{\Tilde{X}}_k^a,\bm{\Tilde{U}}_k^a ) +\bar{J}_N(\bm{\Tilde{X}}_N^a ) \\
            \text{s.t.}\quad
            & \bm{X}_{i,k+1}^a=f^{x,a}_{i,k}\bm{X}_{i,k}^a+f^{u,a}_{i,k}\bm{U}_{i,k}^a+f^{\theta,a}_{i,k},\\
            & \bm{X}_{i,0}^a=\bm{0},\ \bm{X}_{i,k}^a=\bm{\Tilde{X}}_{i,k}^a, \ \bm{U}_{i,k}^a=\bm{\Tilde{U}}_{i,k}^a.
        \end{split}
        \label{eq:auxiliary centralized qp}
    \end{equation}
To see this, observe that the augmented Hessians admit the decompositions $\hat{H}^{xx,a}_{i,k}=H^{xx,a}_{i,k}+\rho^a_i\bm{I}_{n_i}$, $\hat{H}^{uu,a}_{i,k}=H^{uu,a}_{i,k}+\sigma^a_i\bm{I}_{m_i}$, and similarly for the augmented Lagrangian Hessians $\hat{L}^{\Tilde{x}\Tilde{x},a}_{i,k}=L^{\Tilde{x}\Tilde{x},a}_{i,k}+\operatorname{Blkdiag} [ \rho^a_i \bm{I}_{n_i}]_{i=1}^n$ and $\hat{L}^{\Tilde{u}\Tilde{u},a}_{i,k}=L^{\Tilde{u}\Tilde{u},a}_{i,k}+\operatorname{Blkdiag} [ \sigma^a_i \bm{I}_{m_i}]_{i=1}^n$. The ADMM penalty terms can therefore be separated from the Hamiltonian and Lagrangian Hessians, revealing the standard ADMM augmented Lagrangian structure (similar to~\eqref{eq:augmented lagrangian for our problem}) with $\frac{\mathrm{d}\bm{\nu}^{a-1}_{i,k}}{\mathrm{d}\bm{\theta}}$ and $\frac{\mathrm{d}\bm{\xi}^{a-1}_{i,k}}{\mathrm{d}\bm{\theta}}$ acting as the  dual variables associated with the gradient-matching constraints. Here, $\bar{J}_i$ preserves the quadratic structure of the cost function in~\eqref{eq:same LQR system in lemma1} but replaces the quadratic coefficients $\hat{H}^{(\cdot)(\cdot),a}_{i,k}$ with $H^{(\cdot)(\cdot),a}_{i,k}$~\footnote{It is straightforward to show that $\hat{H}^{xu,a}_{i,k}=H^{xu,a}_{i,k}$.} and the linear coefficients $\hat{H}^{(\cdot)\theta,a}_{i,k}$ with $\nabla_{\bm{\theta}}\hat{H}_{i,k}^{(\cdot),a}$. The same construction applies to $\bar{J}_k$ and $\bar{J}_N$ from~\eqref{eq:matrix-valued static optimization subproblem2} and~\eqref{eq:ternimal matrix-valued static optimization subproblem2}, respectively. 

This centralized interpretation~\eqref{eq:auxiliary centralized qp} provides structural insight into the distributed gradient solver (Algorithm~\ref{alg: ADMM-DDP backward pass}) but does not serve as a computational alternative, for two reasons. First, the interpretation is unidirectional: an ADMM-based solver for~\eqref{eq:auxiliary centralized qp} may employ penalty parameters different from $\rho_i^a$ and $\sigma_i^a$, so the distributed solver cannot in general be recovered from~\eqref{eq:auxiliary centralized qp}. Second, at each ADMM iteration, solving~\eqref{eq:auxiliary centralized qp} directly (eg., in a centralized manner) does not yield the true gradient trajectories consistent with the forward pass, since the gradient-matching constraints are only approximately satisfied under ADMM truncation.

\textit{Convexity Under the Centralized View.} Under two mild assumptions that are typically satisfied in practice, \eqref{eq:auxiliary centralized qp} becomes convex, which enables leveraging classical ADMM theory in convex settings to analyze the convergence properties of Algorithm~\ref{alg: ADMM-DDP backward pass}. The first assumption is that Subproblem 1 is constructed with a convex local cost function and is solved using iLQR. For example, if the cost function is~\eqref{eq:quadratic local cost}, then $H^{\tau\tau,a}_{i,k}$ of the quadratic term in $\bar{J}_{i}$ reduces to the Hessian of~\eqref{eq:quadratic local cost}~\footnote{This follows from the omission of second-order system derivatives in iLQR. For the gradient computation, we reuse the Hessians of the Hamiltonian obtained from the forward DDP pass. Although \cite{dinev2022differentiable} argues that including these second-order derivatives is required for full theoretical consistency, our simulations indicate that their omission does not degrade learning performance.}, i.e., $\operatorname{Blkdiag}(\bm{Q}_i,\bm{R}_i)\succ 0$. This implies that $\bar{J}_{i}$ is convex. The second assumption is that, when constructing Auxiliary Subproblem 2 at each ADMM iteration, we apply the regularization strategy $\hat{L}^{\Tilde{\tau}\Tilde{\tau},a}_{k}+(\epsilon-\lambda_{\min}) \bm{I}$ where $\lambda_{\min}$ is the smallest eigenvalue of ${L}^{\Tilde{\tau}\Tilde{\tau},a}_{k}$ and $\epsilon >0$ is a constant, which ensures that $\bar{J}_k$ is convex (a similar regularization, with $\hat{L}^{\Tilde{\tau}\Tilde{\tau},a}_{k}$ replaced by $\hat{L}^{\Tilde{x}\Tilde{x},a}_{N}$, applies to $\bar{J}_N$). Consequently,~\eqref{eq:auxiliary centralized qp} minimizes a convex cost function subject to affine constraints and is therefore convex.

\textit{Boundedness of Gradient Truncation Errors.} Due to the structure mirroring the forward pass, the distributed gradient solver is truncated to the same number of ADMM iterations to improve computational efficiency. The errors between the distributed solver and its centralized counterpart~\eqref{eq:auxiliary centralized qp} are bounded under mild conditions, which ensures stable training, as certified in the following theorem.
\begin{theorem}
\label{tm: Lipschitz bound}
    Let $\bm{\tau}^*_{\mathrm{f}}\coloneqq \{ \bm{\tau}^*_i,\Tilde{\bm{\tau}}^*_{i} \}_{i=1}^n$ be the fully-coordinated stacked trajectory obtained from the forward pass and $\bm{s}^*\coloneqq\{\Tilde{\bm{X}}^*_{i},\Tilde{\bm{U}}^*_{i},\frac{\mathrm{d}\bm{\nu}_{i}^{*}}{\mathrm{d}\bm{\theta}},\frac{\mathrm{d}\bm{\xi}_{i}^{*}}{\mathrm{d}\bm{\theta}}\}_{i=1}^n$ the corresponding stacked gradient of Auxiliary Subproblem 2 and 3 as $a\to \infty$. Assume that, in a neighborhood of $\bm{\tau}^*_{\mathrm{f}}$, all the gradient update operators in Algorithm~\ref{alg: ADMM-DDP backward pass} are locally bounded and Lipschitz in $\bm{\tau}^a_{\mathrm{f}}$. Then there exist local constants $L_T, L_c>0$ and a local operator factor $\Gamma\ge0$ such that, for any finite ADMM iterations $a_{\mathrm{f}}$ with $\bm{\tau}^{a\le a_{\mathrm{f}}}_{\mathrm{f}}$ staying that neighborhood, we have
    \begin{equation}
        \|\bm{s}^{a_{\mathrm{f}}}-\bm{s}^* \|\le \Gamma^{a_{\mathrm{f}}}\|\bm{s}^{0}-\bm{s}^*  \| + \sum_{j=0}^{a_{\mathrm{f}}-1}\Gamma^{j}(L_{T}\|\bm{s}^*\|+L_{c})\delta{\tau}_{\infty},
        \label{eq: theorem error bound}
    \end{equation}
    where $\delta{\tau}_{\infty}\coloneqq\max_{a}\| \bm{\tau}^{a}_{\mathrm{f}}-\bm{\tau}^*_{\mathrm{f}}\|$. Consequently, since the optimal solution of Auxiliary Subproblem 1 at $a$ depends affinely on $\bm{s}^{a-1}_{i}$ with Lipschitz-bounded coefficients, there exist local constants $K_1, K_2>0$ such that
    \begin{equation}
    \begin{aligned}
        \max_{0\leq k < N}\left ( \|\bm{X}^{a_{\mathrm{f}}}_{k+1}-\bm{X}^*_{k+1}\| + \|\bm{U}^{a_{\mathrm{f}}}_k-\bm{U}^*_k \| \right) &\leq K_1 \|\bm{s}^{a_{\mathrm{f}}-1}-\bm{s}^* \| \\
        &\quad + K_2\delta{\tau}_{\infty},
    \end{aligned}
        \label{eq: theorem gradient error}
    \end{equation}
    where $\bm{X}^a_{k}\coloneqq \{ \bm{X}^a_{i,k}\}^n_{i=1}$ and $\bm{U}^a_{k}\coloneqq \{ \bm{U}^a_{i,k}\}^n_{i=1}$.
\end{theorem}
\begin{proof}
    See Appendix~\ref{appendix:proof of theorem1}.
\end{proof}

\section{Applications to Multilift Systems}\label{section:multilift}
\begin{figure}[h]
    \centering
    \includegraphics[width=0.8\linewidth]{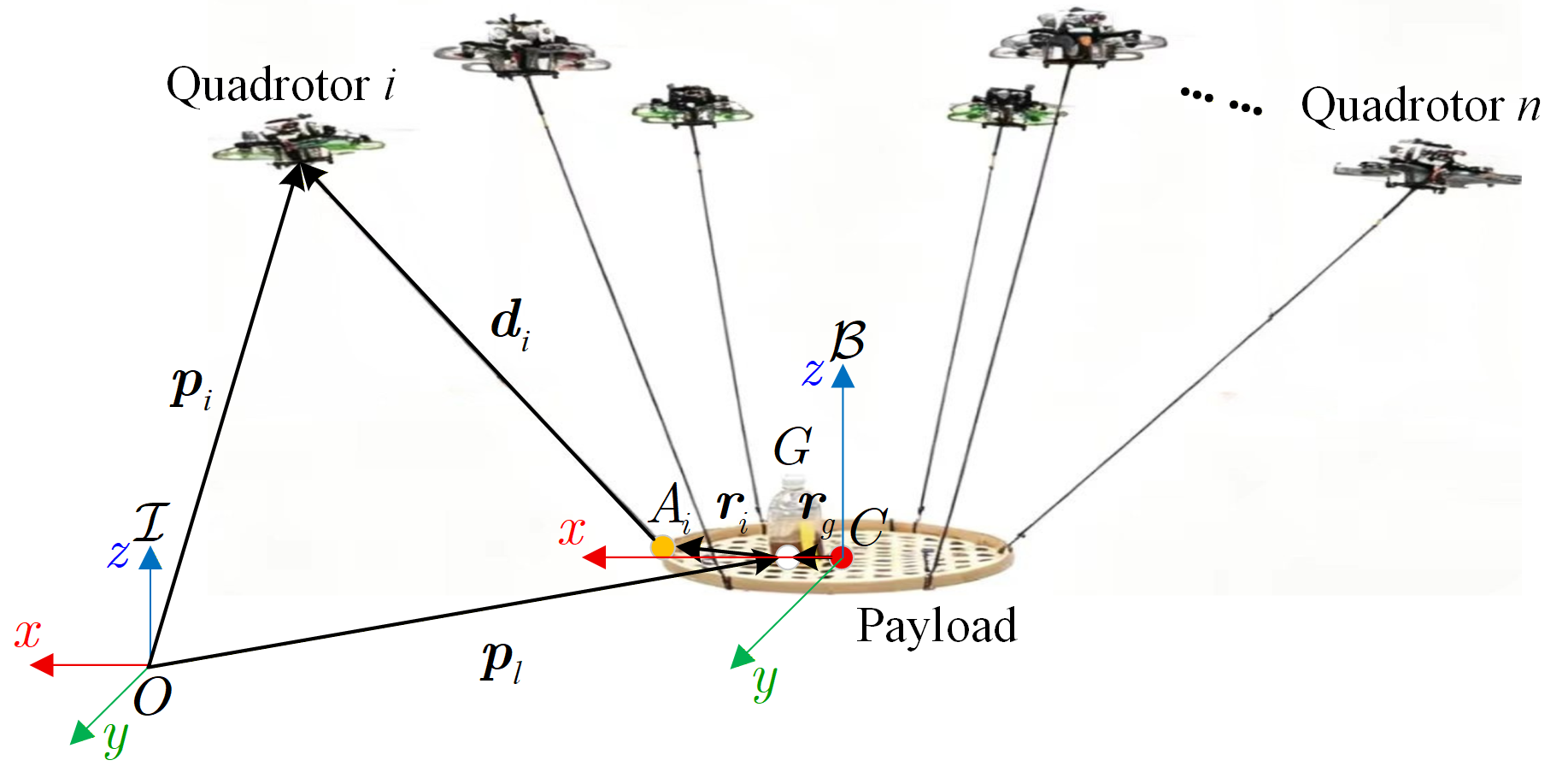}
    \caption{\footnotesize The multilift system with coordinate-frame definitions. The white point marks the center of mass (CoM) of the payload, denoted by $G$, which is offset from the geometric center $C$ by a vector $\bm{r}_g \in \mathbb{R}^{3}$ in the payload body frame ${\mathcal{B}}$. $A_i$ denotes the cable attachment point of cable $i$ on the payload.}
    \label{fig:multilift}
\end{figure}

We apply DiffCoord to meta-learn safe and feasible trajectories for a multilift system. Fig.~\ref{fig:multilift} shows the setup: $n$ quadrotors cooperatively transport a shared payload via $n$ cables. The cable-induced tight kinematic and dynamic couplings distinguish multilift systems from weakly coupled swarm systems, making them a demanding platform for distributed multi-agent trajectory optimization.
% , which DiffCoord is explicitly designed to handle.

Let $m_l \in \mathbb{R}_{+}$ and $\bm{J}_l \in \mathbb{R}^{3\times3}$ denote the payload mass and moment of inertia w.r.t. its CoM, respectively. Let $\bm{p}_{l},\bm{v}_{l}\in \mathbb{R}^{3}$ denote the CoM position and velocity in the world frame $\mathcal{I}$, $\bm{q}_{l}\in \mathbf{S}^{3}$ the payload unit quaternion, and $\bm{\omega}_{l}\in \mathbb{R}^{3}$ the angular velocity in the body frame $\mathcal{B}$. The collective cable force $\bm{F}_{l}\in \mathbb{R}^{3}$ and torque $\bm{M}_{l}\in \mathbb{R}^{3}$ acting on the CoM are expressed in $\mathcal{I}$ and $\mathcal{B}$, respectively. The payload 6-degree-of-freedom (DoF) dynamics are given by

\begin{equation}
    \begin{alignedat}{2}
        \dot{\bm{p}}_{l} &= \bm{v}_{l}, \qquad
        & \dot{\bm{v}}_{l} &= -g\bm{e}_{3} + \frac{\bm{F}_{l}}{m_{l}}, \\
        \dot{\bm{q}}_{l} &= \frac{1}{2}\bm{\Omega}(\bm{\omega}_{l})\bm{q}_{l}, \qquad
        & \dot{\bm{\omega}}_{l} &= \bm{J}_{l}^{-1}
        \left[
        \bm{M}_{l}
        - \bm{\omega}_{l}^{\times}
        \left(\bm{J}_{l}\bm{\omega}_{l}\right)
        \right],
    \end{alignedat}
    \label{eq:load dynamics}
\end{equation}
where $(\bm{a})^{\times}$ denotes the skew-symmetric matrix of $\bm{a}$, $\bm{R}_l(\bm{q}_l)\in\ SO(3)$ maps $\mathcal{B}$ to $\mathcal{I}$, $g$ is the gravitational acceleration, $\bm{e}_3 \coloneqq [0,0,1]^\top$, and $\bm{\Omega}(\bm{\omega}_l) \coloneqq \begin{bmatrix}0&-\bm{\omega}_l^{\top}\\\bm{\omega}_l&-\bm{\omega}_l^\times\end{bmatrix}$.

Following~\cite{sun2023nonlinear,sun2025agile}, we formulate multilift trajectory optimization using cable dynamics rather than quadrotor dynamics for numerical efficiency. Each cable is modeled as massless and inextensible, with the latter justified by the high stiffness of materials such as Nylon~\cite{YEO2013230}. Let $\bm{d}_i \in \mathbf{S}^2$ denote the unit direction of cable $i$ in $\mathcal{I}$, pointing from $A_i$ to quadrotor $i$. Let $\bm{\omega}_i,\bm{\gamma}_i,\bm{a}_i,\bm{j}_i,\bm{s}_i \in \mathbb{R}^3$ denote its angular velocity, acceleration, jerk, and snap in $\mathcal{I}$, and $t_i,v_i,a_{t,i} \in \mathbb{R}$ its tension magnitude, rate, and acceleration. When taut, cable $i$ is governed by
\begin{equation}
        \dot{\bm{d}}_i = \bm{\omega}_i^{\times}\bm{d}_i,\ \dot{\bm{\omega}}_i =\bm{\gamma}_i,\ \dot{\bm{\gamma}}_i= \bm{j}_i, \ \dot{\bm{j}}_i=\bm{s}_i,\
        \dot{t}_i = v_i,\ \dot{v}_i = a_i.
    \label{eq:cable dynamics}
\end{equation}
To keep the cables taut and the payload controllable, we impose
\begin{equation}
    0<t_i\le t_{\max},\ \forall i\in\{1,\cdots,n\},
\end{equation}
where $t_{\max}\in \mathbb{R}_{+}$ is the maximum allowable tension.

The taut cables impose kinodynamic constraints on the multilift system. First, the payload and quadrotor positions are coupled by the cable-length constraint
\begin{equation}
    \bm{p}_i = \bm{p}_l + \bm{R}_l\bm{r}_i + l_i\bm{d}_i, \quad \forall i \in \{1,\cdots,n\},
    \label{eq:length constraint}
\end{equation}
where $\bm{p}_i \in \mathbb{R}^{3}$ is the CoM position of quadrotor $i$ in $\mathcal{I}$, $\bm{r}_i \in \mathbb{R}^{3}$ is the coordinate of $A_i$ in $\mathcal{B}$ relative to $G$, and $l_i \in \mathbb{R}_{+}$ is the cable length. To ensure collision avoidance, we impose
\begin{subequations}
    \begin{align}
        \|\bm{p}_i-\bm{p}_j\| &\geq d_{\min}^{q}, \quad \forall i,j \in \{1,\cdots,n\},\ i\neq j,
    \label{eq:q collision}\\
    \|\bm{p}_i-\bm{p}_o\| &\geq d_{\min}^{o}, \quad \forall i \in \{1,\cdots,n\},
    \label{eq:o collision}
    \end{align}
    \label{eq:safe kinematic constraints}%
\end{subequations}
where $\bm{p}_o\in\mathbb{R}^3$ is the obstacle position, and $d_{\min}^{q},d_{\min}^{o}\in\mathbb{R}_{+}$ are the minimum quadrotor-to-quadrotor and quadrotor-to-obstacle distances. The cable tensions must also generate the payload wrench
\begin{equation}
    \begin{bmatrix}
        \bm{R}^{\top}_l\bm{F}_{l}\\
        \bm{M}_{l}
    \end{bmatrix}
    =
    \underbrace{\begin{bmatrix}
        \bm{I} & \cdots & \bm{I}\\
        \bm{r}_{1}^{\times} & \cdots & \bm{r}_{n}^{\times}
    \end{bmatrix}}_{\coloneqq \bm{P}}
    \begin{bmatrix}
        \bm{R}_{l}^{\top}t_{1}\bm{d}_{1}\\
        \vdots\\
        \bm{R}_{l}^{\top}t_{n}\bm{d}_{n}
    \end{bmatrix}.
    \label{eq:wrench constraint}
\end{equation}
Finally, the quadrotor thrust is bounded by
\begin{equation}
    \|m_i(\ddot{\bm{p}}_i+g\bm{e}_3)+t_i\bm{d}_i\| \leq f_{\max}, \quad \forall i \in \{1,\cdots,n\},
    \label{eq:thrust constraint}
\end{equation}
where $m_i\in\mathbb{R}_{+}$ is the mass of quadrotor $i$ and $f_{\max}\in\mathbb{R}_{+}$ is its maximum collective thrust.

Let $\bm{x}_{l,k}= [\bm{p}_{l,k};\bm{v}_{l,k};\bm{q}_{l,k};\bm{\omega}_{l,k}]$ and $\bm{u}_{l,k}= [\bm{F}_{l,k};\bm{M}_{l,k}]$ denote the load state and control at time step $k$. Similarly, let $\bm{x}_{i,k}= [\bm{d}_{i,k};\bm{\omega}_{i,k};\bm{\gamma}_{i,k};\bm{j}_{i,k};t_{i,k};v_{i,k}]$ and $\bm{u}_{i,k}=[\bm{s}_{i,k};a_{i,k}]$ denote the state and control of cable $i$. We define the load cost $J_l$ and cable costs $\{J_i\}_{i=1}^n$ using the reference-tracking quadratic local cost in~\eqref{eq:quadratic local cost}, as commonly used in multilift trajectory optimization~\cite{9007450,sun2023nonlinear,sun2025agile}. The references can guide the nonconvex optimization and help avoid deadlock. The centralized multilift trajectory optimization problem is then cast as 
\begin{subequations}
    \begin{align}
        \min_{\bm{x}_l,\bm{u}_l,\{\bm{x}_i,\bm{u}_i\}_{i=1}^n} & J_l(\bm{x}_l,\bm{u}_l) +\sum_{i=1}^n J_i(\bm{x}_i,\bm{u}_i) \label{eq:local cost in full multilift system}\\
        \text{s.t.} \quad
        & \bm{x}_{l,k+1} = \bm{f}_{l}(\bm{x}_{l,k},\bm{u}_{l,k}),\ \bm{x}_{l,0}:\text{given},\label{eq:load dynamics in centralized formulation} \\
        & \bm{x}_{i,k+1} = \bm{f}_{i}(\bm{x}_{i,k},\bm{u}_{i,k}),\ \bm{x}_{i,0}:\text{given}, \label{eq:cable dynamics in centralized formulation} \\
        & 0<t_{i,k}\le t_{\max},\label{eq:tension constraint in centralized formulation} \\
        & \text{inter-agent constraints}\ \eqref{eq:safe kinematic constraints},\eqref{eq:wrench constraint},
       \label{eq:interagent kinodynamic constraints}\\
         & \text{quadrotor thrust constraints}\ \eqref{eq:thrust constraint},
    \end{align}
    \label{eq:centralized multilift trajectory optimization}%
\end{subequations}
where $\bm{f}_{l}$ and $\bm{f}_i$ are the discrete-time dynamics corresponding to~\eqref{eq:load dynamics} and~\eqref{eq:cable dynamics}, respectively. 

We solve~\eqref{eq:centralized multilift trajectory optimization} using the ADMM-DDP pipeline in Algorithm~\ref{alg: ADMM-DDP forward pass}. To this end, we introduce safe-copy variables for the load and cable trajectories:
\begin{equation}
        \bm{x}_{l,k}=\Tilde{\bm{x}}_{l,k},\ \bm{u}_{l,k}=\Tilde{\bm{u}}_{l,k}, \ \bm{x}_{i,k}=\Tilde{\bm{x}}_{i,k},\ \bm{u}_{i,k}=\Tilde{\bm{u}}_{i,k}.
    \label{eq:multilift safe copy constraints}
\end{equation}
At each ADMM iteration, Subproblem 1 solves the payload and cable trajectories independently via DDP, Subproblem 2 optimizes the safe-copy variables to satisfy the inter-agent kinodynamic constraints~\eqref{eq:safe kinematic constraints} and~\eqref{eq:wrench constraint}, and the thrust constraints~\eqref{eq:thrust constraint} at each time step, and Subproblem 3 updates the dual variables associated with~\eqref{eq:multilift safe copy constraints}.

In this distributed ADMM-DDP formulation, the problem-level parameters include the payload model parameters, constraint parameters, payload and cable references, and cost weights. The payload model and constraint parameters are determined by the physical platform and obstacle geometry, while the payload reference can be generated using standard single-agent planning methods. The main difficulty lies in constructing cable references $\{t_{i}^{\mathrm{ref}},\bm{d}_{i}^{\mathrm{ref}}\}_{i=1}^n$, which should be compatible with the tension-induced kinodynamic constraints, and tuning the cost weights across tasks. 

We therefore design a two-stage learning procedure. First, we apply DiffCoord to meta-learn kinodynamically feasible cable references. They are generated by solving the following single-agent payload trajectory optimization problem using the ADMM-DDP pipeline:

\begin{subequations}
    \begin{align}
        \min_{\bm{x}_l,\bm{u}_l,\bm{\Pi}} & J_l(\bm{x}_l,\bm{u}_l)+\sum_{k=0}^{N-1}\sum_{i=1}^{n}\frac{1}{2} \left\lVert\bm{t}_{i,k}-\bm{t}^{\mathrm{ref}}_{\mathrm{um}} \right\rVert_{\bm{R}_{t}}^2 \label{eq:single agent cost}\\
        \text{s.t.} \quad
        & \bm{x}_{l,k+1} = \bm{f}_{l}(\bm{x}_{l,k},\bm{u}_{l,k}),\ \bm{x}_{l,0}:\text{given},\\
        & \bm{t}_k = \bm{P}^{\dagger}\begin{bmatrix}
        \bm{R}^{\top}_l\bm{F}_{l}\\
        \bm{M}_{l}
    \end{bmatrix}+\bm{N}\bm{\Pi}_{k}, \label{eq: single agent wrench constraint}\\
        & 0<\|\bm{t}_{i,k}\|\leq t_{\max},\ \text{kinematic constraints}~\eqref{eq:safe kinematic constraints} \label{eq:kinematic constraints in cable reference optimization},
    \end{align}
    \label{eq:single-agent cable reference optimiztion}%
\end{subequations}
Similar to the full multilift problem, this single-agent problem is also implemented using Algorithm~\ref{alg: ADMM-DDP forward pass} with the safe-copy constraints $\bm{x}_{l,k}=\tilde{\bm{x}}_{l,k}$ and $\bm{u}_{l,k}=\tilde{\bm{u}}_{l,k}$. In this pipeline, Subproblem~1 optimizes the payload trajectory via DDP, Subproblem~2 updates the safe-copy variables and $\bm{\Pi}_k$ subject to the kinodynamic constraints~\eqref{eq: single agent wrench constraint} and~\eqref{eq:kinematic constraints in cable reference optimization}, and Subproblem~3 updates the corresponding dual variables. Here, $J_l$ is the same as used in~\eqref{eq:local cost in full multilift system}, $\bm{t}^{\mathrm{ref}}_{\mathrm{um}}=[0;0;m_lg/n]^{\top}$ is the uniform payload gravity allocation per cable, $\bm{R}_t \succ 0 $ is the positive definite weighting matrix of the tension-allocation cost, $\bm{P}^{\dagger}$ is the pseudo-inverse of $\bm{P}$, $\bm{N}\in\mathbb{R}^{3n\times(3n-6)}$ is a basis of the null space of $\bm{P}$, $\bm{\Pi}_k\in\mathbb{R}^{3n-6}$ is the null-space coefficient, $\bm{t}_{i,k}$ is the tension of cable $i$ in $\mathcal{B}$ extracted from $\bm{t}_k$, and $\bm{p}_{i,k}$ in~\eqref{eq:kinematic constraints in cable reference optimization} is given by $\bm{p}_{i,k}=\bm{p}_{l,k}+\bm{R}_{l,k}(\bm{r}_i+l_i\bm{t}_{i,k}/\|\bm{t}_{i,k}\|)$. The meta-learned parameters of this single-agent problem include the cost weights and the open-loop ADMM penalty policies. We parameterize the state and control penalties at ADMM iteration $a$ as
\begin{equation}
    \rho_{ls}^a=\operatorname{Sig} ^a(\alpha_{\rho,ls})\rho_{ls},\qquad
    \sigma_{ls}^a=\operatorname{Sig}^a(\alpha_{\sigma,ls})\sigma_{ls},
    \label{eq:open_loop_penalty_policy}
\end{equation}
where $\operatorname{Sig}^a(\alpha)=\frac{1}{1+\exp[-\alpha(a-a_{\mathrm{offset}})]}$ and $a_{\mathrm{offset}}$ is set to the middle of the ADMM iteration range. Here, $\rho_{ls},\sigma_{ls}\in\mathbb{R}_{+}$ are learnable base penalty parameters, and $\alpha_{\rho,ls},\alpha_{\sigma,ls}\in\mathbb{R}$ are learnable shape parameters. The sign of $\alpha$ determines whether the penalty policy increases ($\alpha >0$), decreases ($\alpha <0$), or remains constant ($\alpha = 0$) over ADMM iterations, enabling flexible open-loop penalty adaptation through meta-learning. Accordingly, we collect the Stage-1 parameters as $\bm{\theta}_{ls}=[\operatorname{vec}(\bm{Q}_{ls}),\operatorname{vec}(\bm{R}_{ls}),\operatorname{vec}(\bm{Q}_{ls,N}),\operatorname{vec}(\bm{R}_{t}),\rho_{ls},\sigma_{ls},\alpha_{\rho,ls},\alpha_{\sigma,ls}]$.

By setting $t^{\mathrm{ref}}_{i,k}\coloneqq \| \bm{t}_{i,k}\|$ and $\bm{d}^{\mathrm{ref}}_{i,k}\coloneqq \bm{R}_{l,k}\bm{t}_{i,k}/\|\bm{t}_{i,k}\|$ as the cable references, we then apply DiffCoord to the full multilift problem to meta-learn the payload and cable cost weights and their open-loop ADMM penalty policies. The penalty policies use the same parameterization as~\eqref{eq:open_loop_penalty_policy}. Since all cables share the same dynamics~\eqref{eq:cable dynamics} and are isomorphic, they use a common set of the parameters to improve the scalability. Accordingly, we collect the Stage-2 payload and cable parameter vectors as $\bm{\theta}_l=[\operatorname{vec}(\bm{Q}_l),\operatorname{vec}(\bm{R}_l),\operatorname{vec}(\bm{Q}_{l,N}),\rho_l,\sigma_l,\alpha_{\rho,l},\alpha_{\sigma,l}]$ and $\bm{\theta}_c=[\operatorname{vec}(\bm{Q}_c),\operatorname{vec}(\bm{R}_c),\operatorname{vec}(\bm{Q}_{c,N}),\rho_c,\sigma_c,\alpha_{\rho,c},\alpha_{\sigma,c}]$, respectively.

These parameters $\bm{\theta}_{ls}$, $\bm{\theta}_l$, and $\bm{\theta}_c$ are generated by three lightweight neural networks, respectively, to enable task adaptation. The latter two serve as agent-wise networks for the payload and cables in the full multilift problem. Following Subsection~\ref{subsec:meta_learning via gradient}, we define the Stage-2 meta-loss over $M$ tasks as
\begin{equation}
    L_{\mathrm{m}}=\frac{1}{M}\sum_{t=1}^M\underbrace{\left(w_p\ell_{l,p}^{t}+w_r\ell_{l,r}^{t}+\sum_{i=1}^n(w_p\ell_{i,p}^{t}+w_r\ell_{i,r}^{t}) \right)}_{\coloneqq L_t},
    \label{eq: multilift meta loss}
\end{equation}
where $\ell_{l,p}^{t}=\|\bm{x}^{a_{\mathrm{f}}}_{l,t}-\bm{x}^{\mathrm{ref}}_{l,t}\|^2_2$ measures the payload tracking error, and $\ell_{l,r}^{t}=\|\bm{\tau}^{a_{\mathrm{f}}}_{l,t}-\tilde{\bm{\tau}}^{a_{\mathrm{f}}}_{l,t}\|^2_2$ penalizes the payload ADMM residual. The cable losses $\ell_{i,p}^{t}$ and $\ell_{i,r}^{t}$ are defined analogously using the cable references generated in Stage-1 and the corresponding ADMM residuals. The positive coefficients $w_p$ and $w_r$ balance task performance and inter-agent coordination~\footnote{In implementation, $w_p$ and $w_r$ can be adjusted according to the relative magnitudes of the tracking and residual losses during training.}. The Stage-1 meta-loss is obtained by applying the same form to the reduced payload problem and removing the cable terms.

% In practice, we adopt an efficient sparse parameterization for the parameters. All the cost weights are set as diagonal matrices to reduce the problem's size. The diagonal entries of $\bm{Q}$, $\bm{R}$, and $\bm{Q}_N$ are parameterized as $Q{.} = w_{\min} + (w_{\max} - w_{\min}) q_{.}$, $R_{.} = w_{\min} + (w_{\max} - w_{\min}) r_{.}$, and $Q_{N,.} = w_{\min} + (w_{\max} - w_{\min}) q_{N,.}$, where $w_{\min}, w_{\max} \in \mathbb{R}{+}$ define the parameter bounds and the subscript “.” denotes the index. The ADMM penalty parameter is similarly parameterized as $\rho_l = w_{\min} + (w_{\max} - w_{\min}) \bar{\rho}_l$. Under this scheme, the neural network output for the load becomes $\bm{\Theta}_l = [q_{:}, q_{N,:}, r_{:}, \bar{\rho}_l]\in \mathbb{R}^{33}$ where ":" denotes the dimension of a vector collecting the diagonal elements like $p_{:}=[p_1,\dots]$. This parameterization strategy applies similarly to the cable and to Problem~\eqref{eq:single-agent cable reference optimiztion}, yielding the network outputs $\bm{\Theta}_c\in \mathbb{R}^{21}$ and $\bm{\Theta}_{ls}\in \mathbb{R}^{36}$, respectively. To reflect this parameterization, the chain rule in the gradient computation~\eqref{eq:chain rule loss gradient2} must additionally account for the gradients $\frac{\partial \bm{\theta}_l}{\partial \bm{\Theta}_l}$, $\frac{\partial \bm{\theta}_c}{\partial \bm{\Theta}_c}$, and $\frac{\partial \bm{\theta}_{ls}}{\partial \bm{\Theta}_{ls}}$.

\section{Experiments}\label{section:experiments}

As an example, DiffCoord is applied to a multilift system, which is detailed in Section~\ref{section:multilift}. In this section, we validate its efficacy through extensive numerical and physical experiments involving multilift flight through constrained spaces across diverse tasks. We consider a practical delivery setting with nonuniform payload mass distributions and uncertain CoM locations, since real payloads are typically dynamically asymmetric. Accordingly, in each task, the payload CoM is offset from its geometric center by $\bm{r}_g$ in the payload body frame, as shown in Fig.~\ref{fig:multilift}. Specifically, we demonstrate the following advantages of DiffCoord:
\begin{enumerate}
\item Efficient training within limited ADMM iteration budgets and strong scalability with both agent count and trajectory horizon length;
\item Improved gradient computation efficiency for the dynamic optimization subproblem over the state-of-the-art full-Jacobian methods;
\item Effective meta-learning of task-adaptive parameters for the ADMM-DDP pipeline to generate feasible cable trajectories, enabling the multilift system to adaptively reconfigure its formation across different tasks;
\item Strong generalization of the meta-learned ADMM-DDP pipeline to unseen tasks, environments, and team sizes in real flight experiments, together with robustness to model uncertainties through DDP feedback gains.
\end{enumerate}

For numerical stability, we constrain the learnable parameters $\bm{\theta}_{ls}$, $\bm{\theta}_l$, and $\bm{\theta}_c$ within prescribed bounds. For example, each element in $\bm{Q}_l$ is parameterized as $Q_{\cdot}=w_{\min}+(w_{\max}-w_{\min})q_{\cdot}$, where $w_{\min}, w_{\max} \in \mathbb{R}_{+}$ define the lower and upper bounds, $q_{\cdot}\in [0,1]$ is the normalized parameter, and the subscript "$\cdot$" denotes the element index. We set $w_{\min}=0.001$ and $w_{\max}=1000$ for all the elements except for the shape parameters $\alpha_{\rho}$ and $\alpha_{\sigma}$, which are bounded within $[-3,3]$ to control the changing pattern of the ADMM penalty parameters over ADMM iterations. We further set all the cost weighting matrices in these parameters to be diagonal to reduce the problem size. The three networks therefore output three sets of the normalized parameters and the bounded shape parameters: $\bm{\Theta}_{ls}\in \mathbb{R}^{39}$, $\bm{\Theta}_{l}\in \mathbb{R}^{36}$, and $\bm{\Theta}_{c}\in \mathbb{R}^{36}$, respectively, and the sigmoid activations are used in their output layers. Each network has two hidden layers with the rectified linear unit (ReLU) activation, forming a standard multilayer perceptron (MLP) architecture. To generate the task-adaptive parameters, the input to the network for $\bm{\Theta}_{ls}$ is the magnitude of the payload CoM offset vector $\|\bm{r}_g \|$, while the networks for $\bm{\Theta}_{l}$ and $\bm{\Theta}_{c}$ take the planar coordinates $[x,y]$ of $\bm{r}_g$ as their inputs. Consequently, the three MLPs have the following layer sizes: $\bm{\Theta}_{ls}: 1\to16\to32\to39$, $\bm{\Theta}_{l}: 2\to16\to32\to36$, and $\bm{\Theta}_{c}: 2\to16\to32\to36$. Their learnable parameters are denoted by $\bm{\varpi}_{ls}$, $\bm{\varpi}_{l}$, and $\bm{\varpi}_{c}$, respectively. To reflect the bounded and sparse parameterization, we additionally include the gradients $\frac{\partial \bm{\theta}_{ls}}{\partial \bm{\Theta}_{ls}}$, $\frac{\partial \bm{\theta}_{l}}{\partial \bm{\Theta}_{l}}$, and $\frac{\partial \bm{\theta}_{c}}{\partial \bm{\Theta}_{c}}$ in the chain rule~\eqref{eq:chain rule loss gradient2} for training the three MLPs, respectively.

We implement DiffCoord in Python. In the forward pass, we solve Subproblem 1 using iLQR and Subproblem 2 using \texttt{ipopt} through CasADi~\cite{andersson2019casadi}. The three MLPs are built using PyTorch~\cite{paszke2019pytorch} and trained with \texttt{Adam}~\cite{kingma2015adam}. In the implementation, we customize the meta-loss~\eqref{eq: multilift meta loss} to align with the standard network training procedure in PyTorch. For example, the customized meta-loss used to train the MLP for $\bm{\Theta}_{ls}$ is defined as $L_{\mathrm{py}}=\frac{\partial L_{\mathrm{m}}}{\partial \bm{\Theta}_{ls}}\bm{\Theta}_{ls}$, where $\frac{\partial L_{\mathrm{m}}}{\partial \bm{\Theta}_{ls}}$ is the gradient of the meta-loss~\eqref{eq: multilift meta loss} w.r.t. $\bm{\Theta}_{ls}$ and is treated as fixed during PyTorch backpropagation. This construction ensures  $\frac{\mathrm{d} L_{\mathrm{py}}}{\mathrm{d} \bm{\varpi}_{ls}}=\frac{\mathrm{d} L_{\mathrm{m}}}{\mathrm{d}\bm{\varpi}_{ls}}$. All the simulations are conducted on a workstation equipped with an 11th Gen Intel Core i7-11700K processor.

\subsection{Efficient Training and Gradient Computation}\label{subsec: efficient training and gradient computation}

In this numerical experiment, we investigate the effect of ADMM iteration budgets on training performance and compare the proposed trajectory gradient solver (defined in Lemma~\ref{lm:DDP gradient solver}) with the state-of-the-art full-Jacobian methods~\cite{NEURIPS2020_5a7b238b} and~\cite{11214470} to show the first and second advantages of DiffCoord.

\begin{figure}[h]
    \centering
    \begin{subfigure}{0.4\columnwidth}
        \centering
        \includegraphics[width=\linewidth]{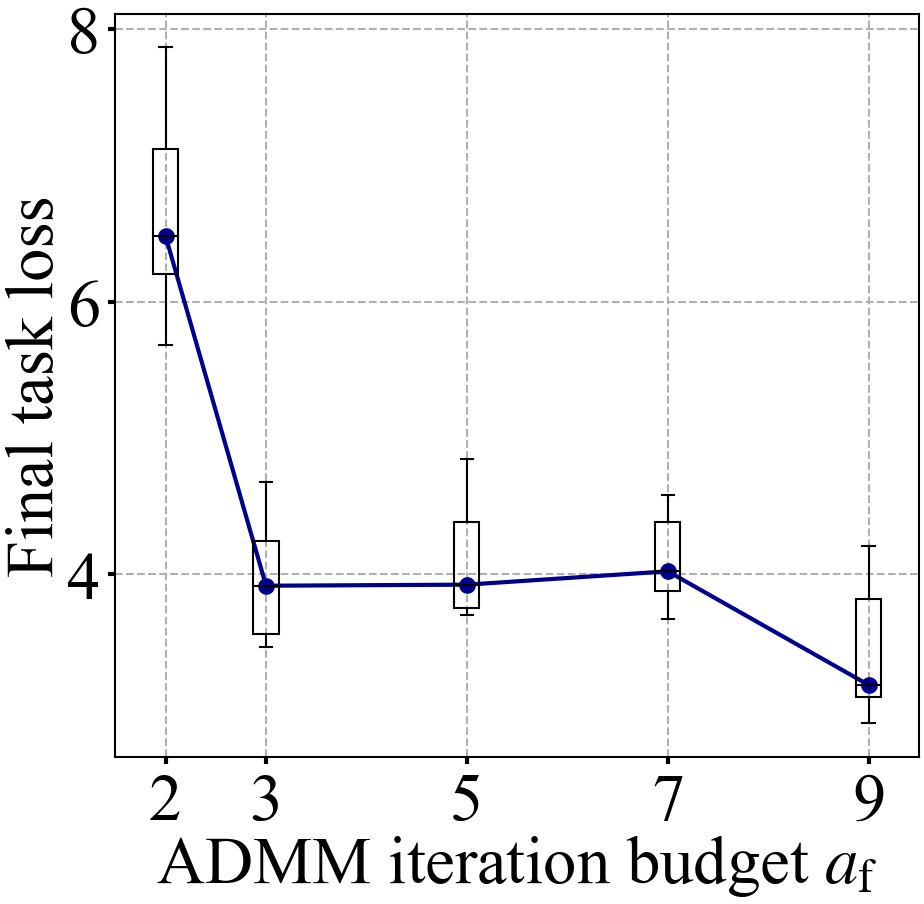}
        \caption{}
        \label{fig:admm_iteration_effect}
    \end{subfigure}
    \quad
    \begin{subfigure}{0.4\columnwidth}
        \centering
        \includegraphics[width=\linewidth]{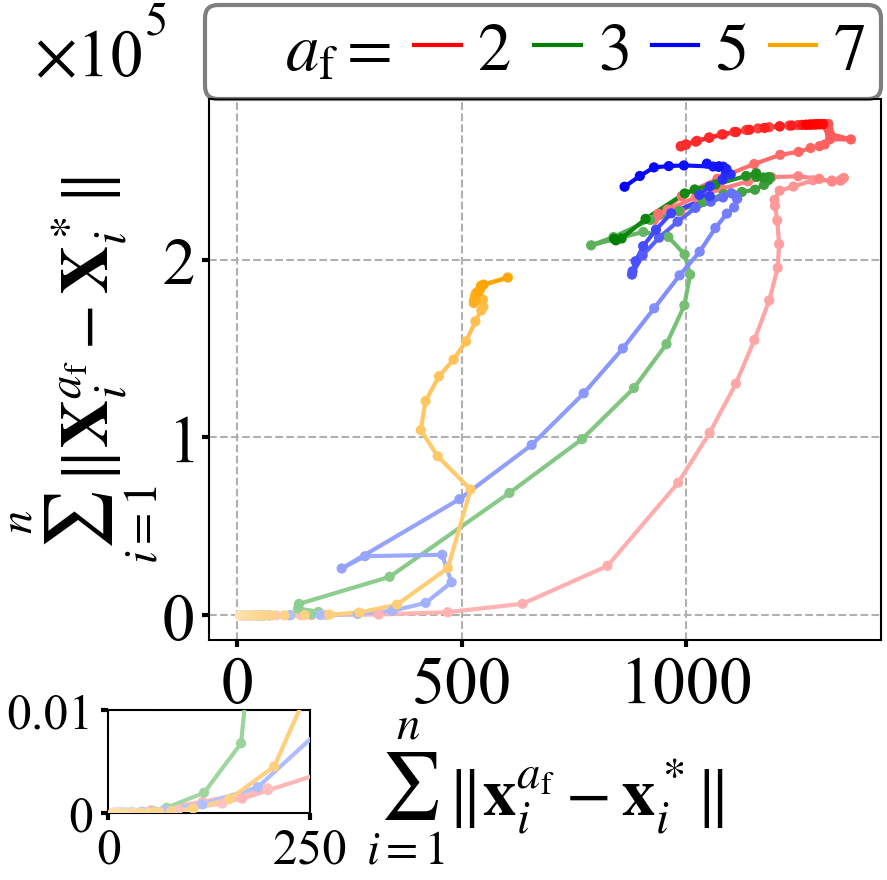}
        \caption{}
        \label{fig:gradient_error_admm_iteration}
    \end{subfigure}
   \caption{Effect of ADMM iteration budgets on training. (a) Task losses over 10 tasks at the final training episode under different ADMM iteration budgets. (b) Truncation-induced gradient and trajectory errors, where the color of each curve becomes lighter as the training episode increases. We compare the trajectories and gradients obtained at different $a_{\mathrm{f}}$ values with those obtained at $a_{\mathrm{f}}=9$, which are assumed to be close enough to those produced when the ADMM pipelines are nearly converged.}
    \label{fig:admm_iteration_effect_gradient}
\end{figure}

Fig.~\ref{fig:admm_iteration_effect} shows that the task loss $L_t$, defined in~\eqref{eq: multilift meta loss}, at the final training episode decreases dramatically from $a_{\mathrm{f}}=2$ to $a_{\mathrm{f}}=3$ and becomes relatively stable after $a_{\mathrm{f}}=3$. This confirms efficient training under very limited ADMM iteration budgets and allows us to set $a_{\mathrm{f}}=3$ hereafter in training. Fig.~\ref{fig:gradient_error_admm_iteration} further shows that both the truncation-induced trajectory and gradient errors decrease as the training episode and $a_{\mathrm{f}}$ increase. This indicates that the training process can effectively improve the consistency of both the truncated forward and backward passes with their nearly converged counterparts. Notably, these gradient errors almost vanish around the trajectories generated by the nearly converged ADMM-DDP pipeline (see the zoom-in plot in Fig.~\ref{fig:gradient_error_admm_iteration}). This numerically validates Theorem~\ref{tm: Lipschitz bound}.

\begin{figure}[h]
    \centering
    \includegraphics[width=0.8\columnwidth]{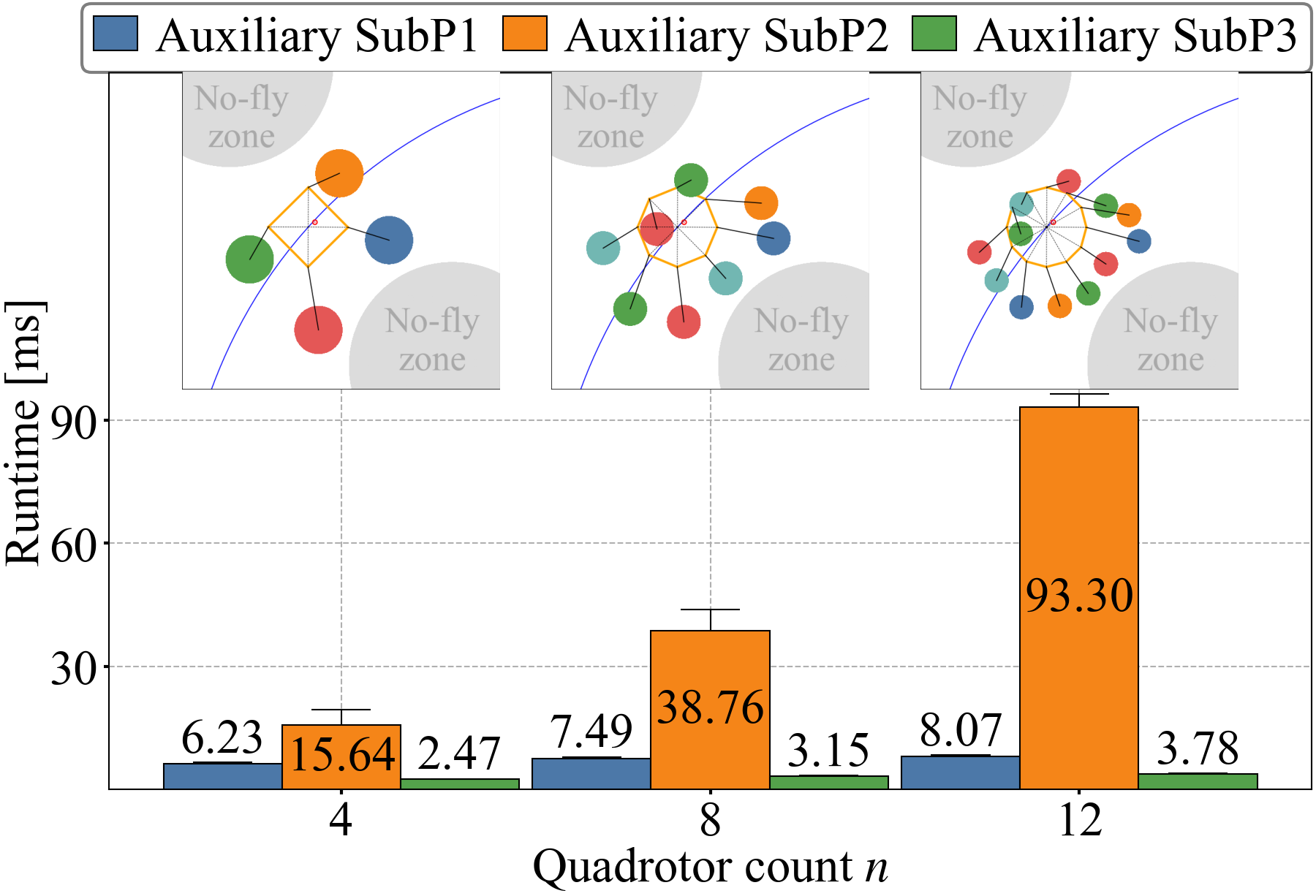}
    \caption{Computational times for solving the three auxiliary subproblems under different quadrotor counts with $N=100$. The numbers denote the median values. The shown computational times are averaged across ADMM iterations.}
    \label{fig:gradient_comparison_agent_count}
\end{figure}

\begin{table}[h]
\centering
\caption{Comparisons of computational times for solving the three auxiliary subproblems under different horizon lengths with $n=4$.}
\label{tab:gradient_horizon_length_comparison}
\begin{tabular}{c !{\vrule width 0.25pt} c !{\vrule width 0.25pt} c !{\vrule width 0.25pt} c !{\vrule width 0.25pt} c}
\toprule
Runtime [ms] & $N=50$ & $N=100$ & $N=150$ & $N=200$ \\
\midrule
Aux. SubP1 & 3.74  & 6.23  & 8.65  & 11.21 \\
Aux. SubP2 & 15.39 & 15.64 & 15.79 & 16.08 \\
Aux. SubP3 & 1.99  & 2.47  & 2.90  & 3.46  \\
\bottomrule
\end{tabular}
\end{table}

\begin{figure}[h]
    \centering
    \begin{subfigure}{0.4\columnwidth}
        \centering
        \includegraphics[width=\linewidth]{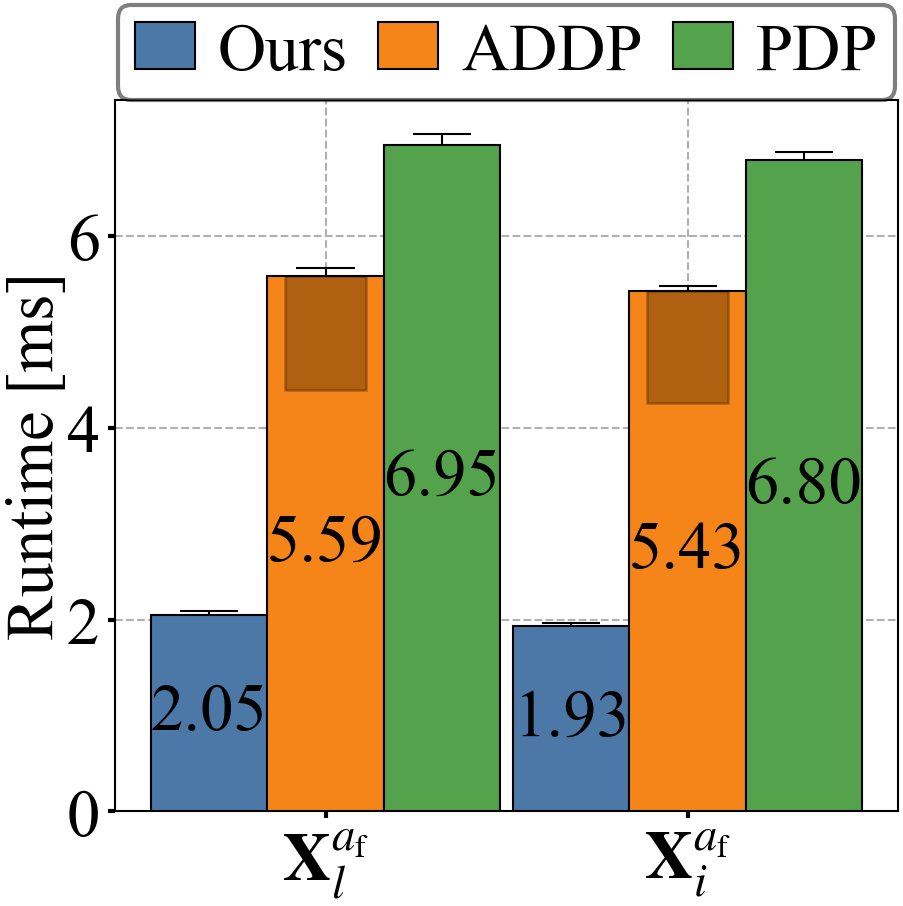}
        \caption{}
        \label{fig:gradient_runtime_comparison_with_PDP}
    \end{subfigure}
    \quad
    \begin{subfigure}{0.4\columnwidth}
        \centering
        \includegraphics[width=\linewidth]{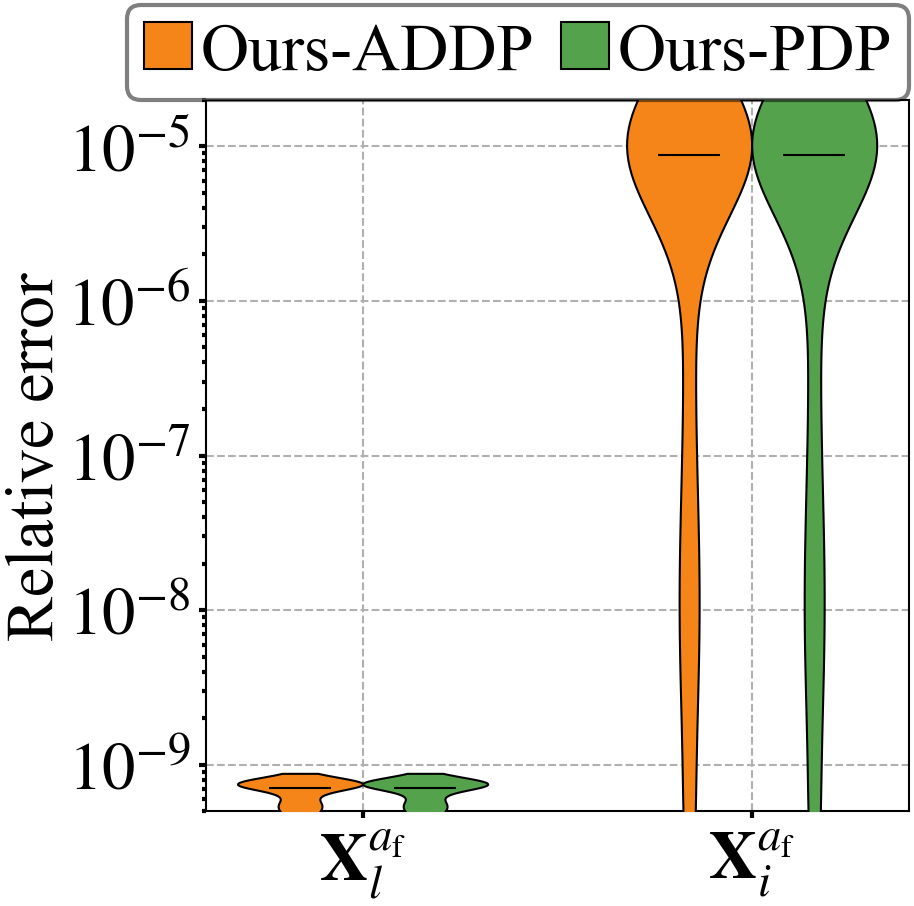}
        \caption{}
        \label{fig:gradient_error_comparison_with_PDP}
    \end{subfigure}
   \caption{Comparisons of computational times and relative errors between our trajectory gradient solver and the state-of-the-art full-Jacobian methods with $N=100$. (a) Comparison of computational times. (b) Comparison of gradient relative errors. For $\bm{X}_l^{a_{\mathrm{f}}}$, its relative error w.r.t. the state-of-the-art methods is defined by $\frac{1}{N}\sum_{k=1}^N\| \bm{X}_{l,k}^{a_{\mathrm{f}}}-\bm{X}_{l,k}^{a_{\mathrm{f}},\mathrm{m}}\|_{F}/\|\bm{X}_{l,k}^{a_{\mathrm{f}},\mathrm{m}}\|_{F}$ where $\mathrm{m}$ denotes ADDP or PDP. The same definition applies to the relative error of $\bm{X}_i^{a_{\mathrm{f}}}$.}
    \label{fig:gradient_comparison_with_PDP}
\end{figure}

Next, we test the scalability of DiffCoord by measuring the computational times of its three auxiliary subproblems under different quadrotor counts $n$ and trajectory horizon lengths $N$. Fig.~\ref{fig:gradient_comparison_agent_count} shows that the computational times of Auxiliary Subproblems 1 and 3 increase only mildly with $n$. However, the computational time of Auxiliary Subproblem 2 increases geometrically with $n$. This comes from the dense tension-induced coupling constraints between the quadrotors and the payload, which grow rapidly as more quadrotors are used. For robot swarm systems without such dense inter-agent couplings, this sharp increase can be eliminated by fixing the neighborhood set of each agent, as detailed in Remark~\ref{remark:decentralized auxiliary subproblem2}. Table~\ref{tab:gradient_horizon_length_comparison} compares the computational times of the auxiliary subproblems over different horizon lengths. Auxiliary Subproblem 1 scales approximately linearly with $N$, confirming the linear computational complexity $\mathcal{O}(N)$ of our method (see Lemma~\ref{lm:DDP gradient solver}). The runtime of Auxiliary Subproblem 2 remains almost constant, while that of Auxiliary Subproblem 3 increases mildly as $N$ increases. This demonstrates the strong scalability of our method for long-horizon planning.

For computing the trajectory gradients of Subproblem 1 (defined in~\ref{subsubp1}), we further compare the proposed DBP-based solver against two state-of-the-art full-Jacobian methods, PDP~\cite{NEURIPS2020_5a7b238b} and augmented DDP (ADDP)~\cite{11214470}, in Fig.~\ref{fig:gradient_comparison_with_PDP}. PDP computes the trajectory gradients by solving the same auxiliary matrix-valued LQR problem as our method (Lemma~\ref{lm:DBP and PDP equivalence}), but through the PMP conditions (Lemma 5.2 in~\cite{NEURIPS2020_5a7b238b}). ADDP augments Subproblem 1 with $\bm{\theta}$ to form the augmented state $\bm{y}\coloneqq[\bm{x};\bm{\theta}]$ and computes the gradients $\frac{\delta \bm{y}}{\delta \bm{\theta}}$ by running one DDP iteration on the augmented problem (Theorem III.3 in~\cite{11214470}). Fig.~\ref{fig:gradient_runtime_comparison_with_PDP} shows that our approach outperforms the other two methods in both the payload and cable gradients, achieving up to $70\%$ faster computation\footnote{The Hamiltonian Hessians are computed from iLQR during the solve of Subproblem 1 in the forward pass and reused by all the three methods for gradient computation, and thus are excluded from the CPU runtime. This allows us to compare the computational time differences arising solely from their algorithmic structures.}. The improvement stems from two features of our method. First, it can reuse many key DDP results from the forward pass due to its structural overlap with the DDP solve. Second, relative to ADDP, it further benefits from the more compact Riccati recursion over $V^{x\theta}_k$, which is a block of the value-function Hessian on the augmented state $V^{yy}_k$. The redundant block $V^{\theta\theta}_k$ in ADDP does not affect the gradients but introduces additional computation (see Appendix~\ref{appendix:proof of lemma2}), whose runtime is roughly $50\%$ of that of our method, as shown in the shaded orange area in Fig.~\ref{fig:gradient_runtime_comparison_with_PDP}. Our method is theoretically equivalent to PDP and ADDP in computing the trajectory gradients (see Appendices~\ref{appendix:proof of lemma1} and~\ref{appendix:proof of lemma2}). To validate this equivalence numerically, we compute two sets of relative gradient errors by comparing our method with PDP and ADDP, respectively. Fig.~\ref{fig:gradient_error_comparison_with_PDP} shows that the maximum error across all the gradients is below $0.01\%$, confirming the theoretical equivalence. 

\subsection{Effective Meta-Learning for Multilift Planning}\label{subsec: meta_learning for multilift}
\begin{figure}[h]
    \centering
    \begin{subfigure}{0.4\columnwidth}
        \centering
        \includegraphics[width=\linewidth]{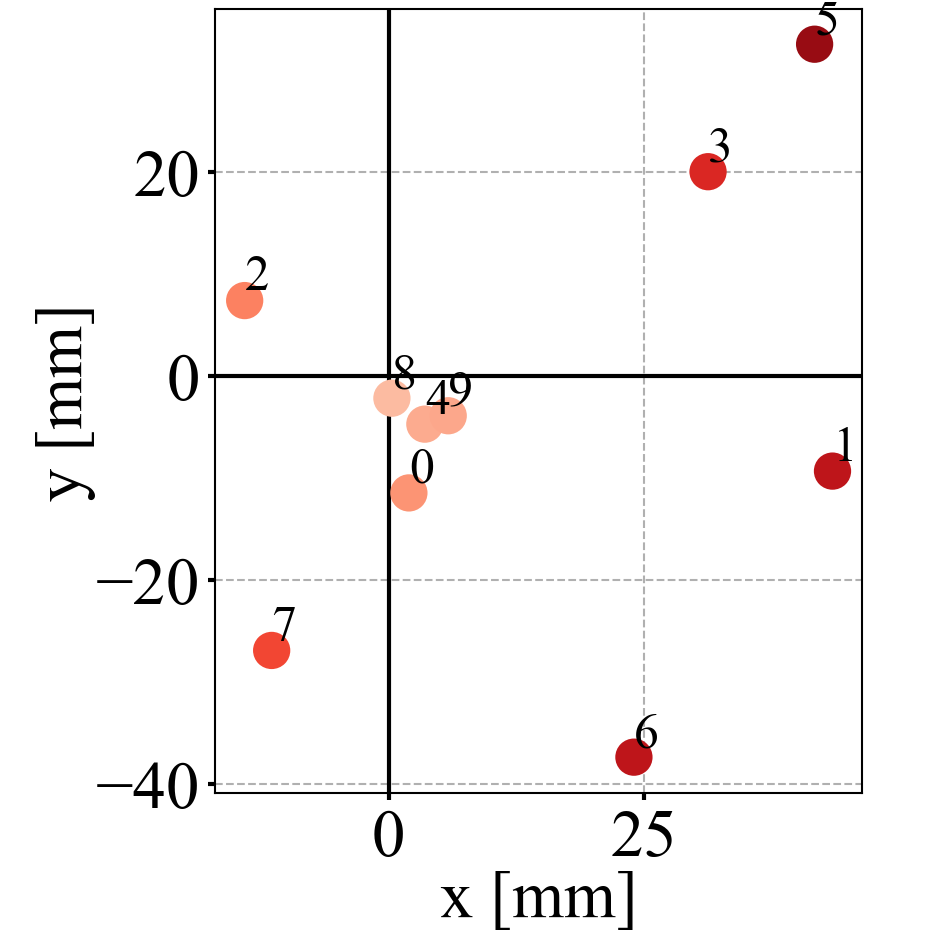}
        \caption{}
        \label{fig:load_coordinate}
    \end{subfigure}
    \quad
    \begin{subfigure}{0.4\columnwidth}
        \centering
        \includegraphics[width=\linewidth]{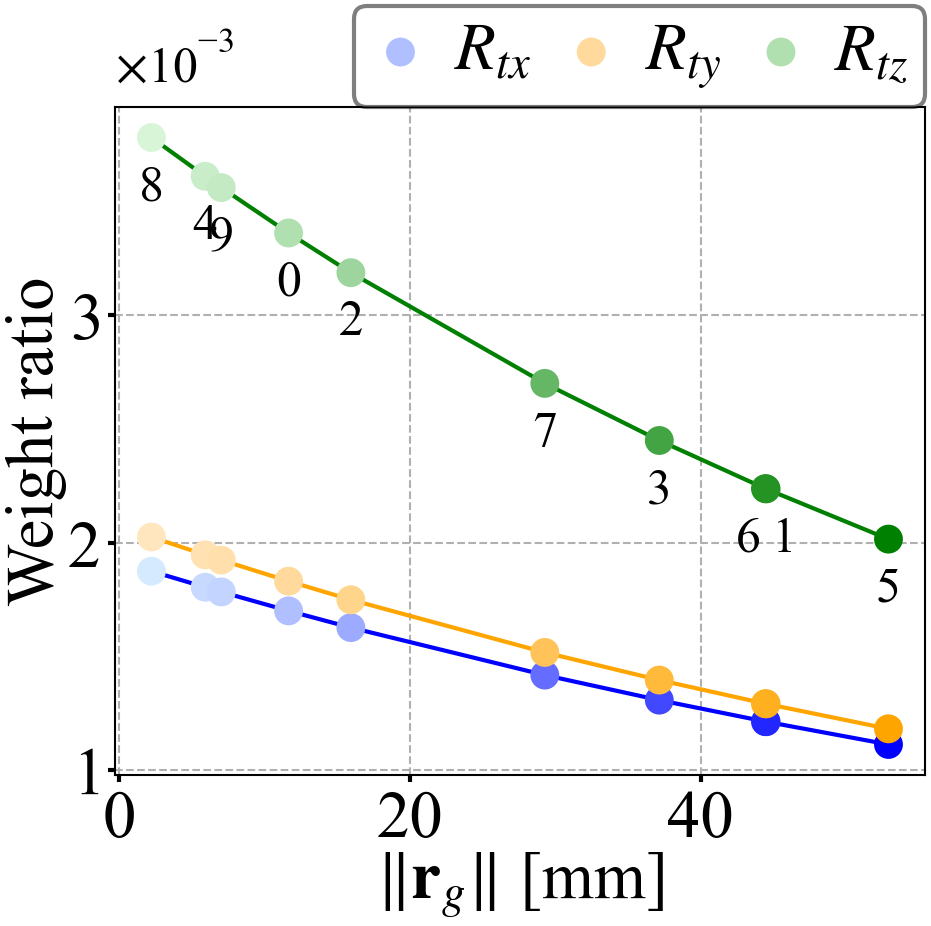}
        \caption{}
        \label{fig:weight_ratio_rg_stage1}
    \end{subfigure}
   \caption{Task settings and task-adaptive parameters. (a) Payload CoM coordinates $\bm{r}_g$ in the body frame across 10 tasks. The numbers denote the task indices, and the color darkness is proportional to the magnitude $\|\bm{r}_g\| $. (b) Ratios between the meta-learned parameters of Subproblem 2 in Stage 1.}
    \label{fig:rg_coordinate_and_weight_ratio}
\end{figure}

To show the third advantage of DiffCoord, we meta-learn the ADMM-DDP pipeline to plan feasible cable trajectories for a 4-quadrotor multilift system in constrained-space flight over $M=10$ tasks. In each task, the planar coordinates of $\bm{r}_g$ are parameterized as $[\| \bm{r}_g\|\cos\alpha_g;\|\bm{r}_g\|\sin\alpha_g]$, where $\| \bm{r}_g\|$ and $\alpha_g$ are randomly sampled from two uniform distributions, $\mathcal{U}(0,0.054 \ \mathrm{m})$ and $\mathcal{U}(0,2\pi \ \mathrm{rad})$, respectively, as shown in Fig.~\ref{fig:load_coordinate}. 

\begin{figure}[h]
    \centering
    \begin{subfigure}{0.4\columnwidth}
        \centering
        \includegraphics[width=\linewidth]{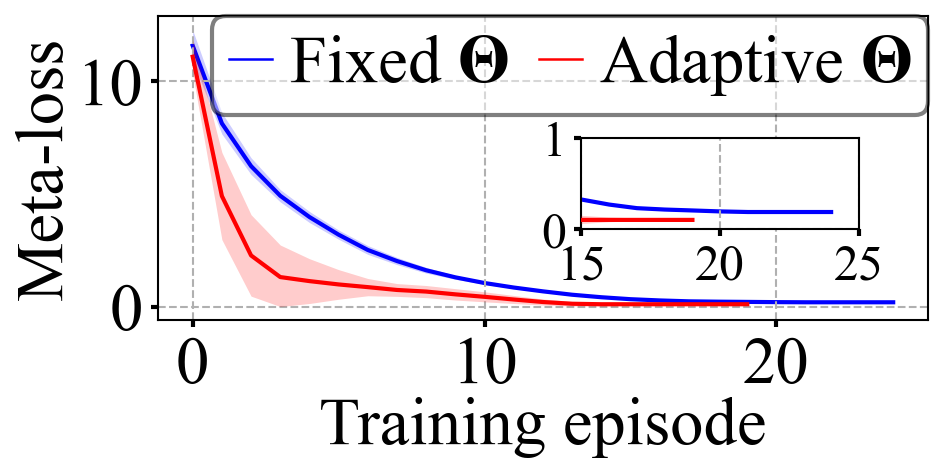}
        \caption{}
        \label{fig:stage_1_loss}
    \end{subfigure}
    \quad
    \begin{subfigure}{0.4\columnwidth}
        \centering
        \includegraphics[width=\linewidth]{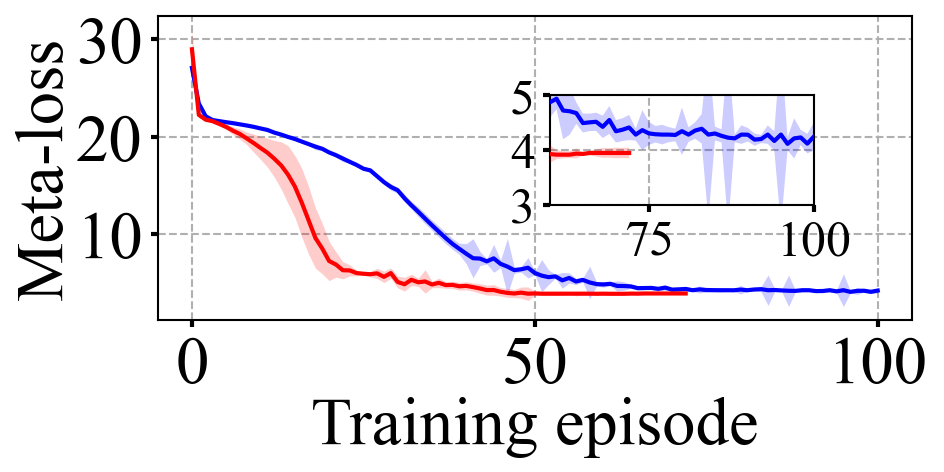}
        \caption{}
        \label{fig:stage_2_loss}
    \end{subfigure}
   \caption{Meta-loss comparisons during training under task-fixed and task-adaptive parameter settings. (a) Stage-1 meta-loss. (b) Stage-2 meta-loss. In each setting, training is repeated five times with randomly initialized networks or task-fixed parameters. The solid curves denote the median values, while the shaded regions indicate the standard deviations.}
    \label{fig:meta_loss_in_training}
\end{figure}

\begin{figure}[h]
    \centering
    \includegraphics[width=0.8\linewidth]{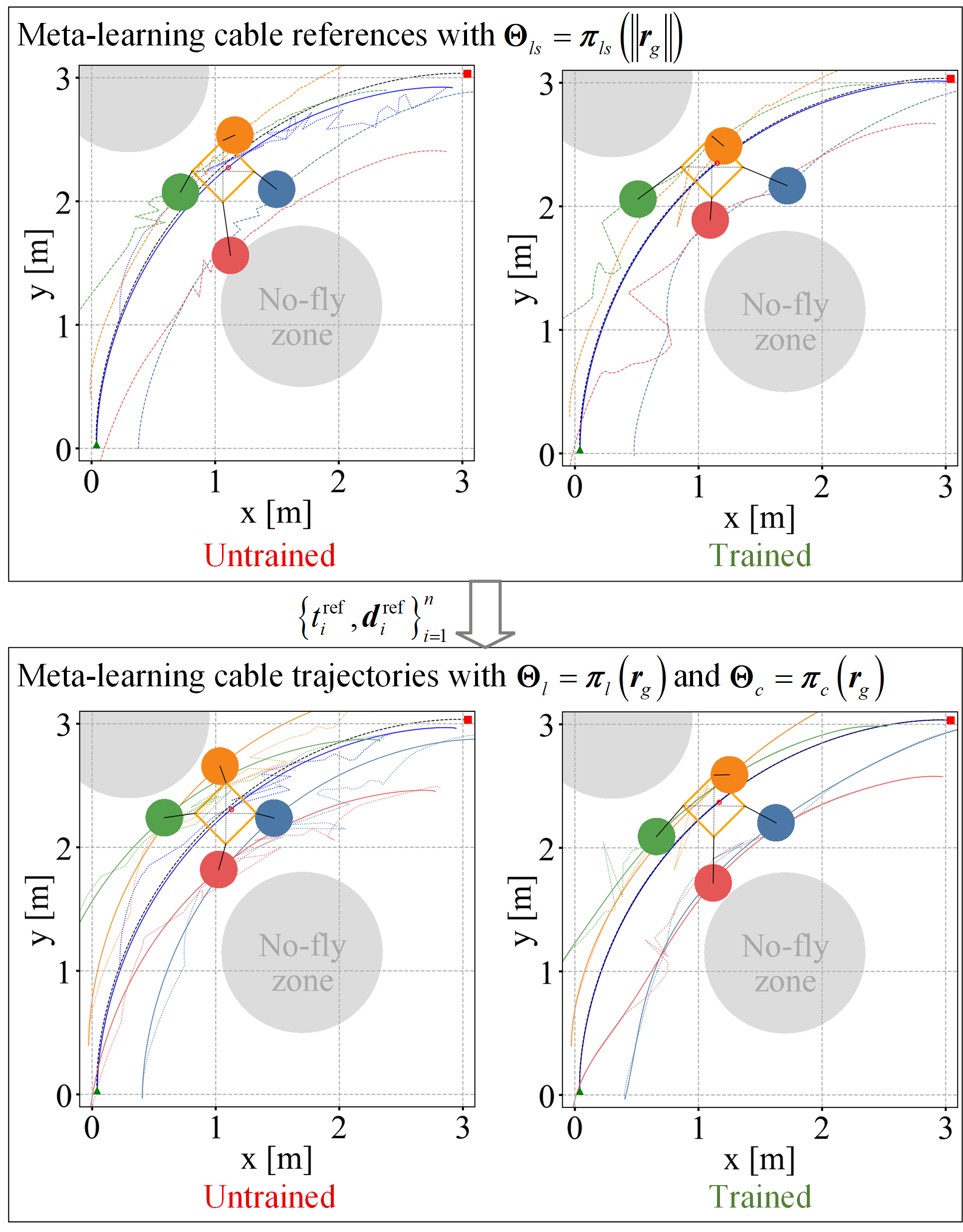}
   \caption{\footnotesize Two-stage training process in Task 5 (top view). A 4-s payload reference trajectory is planned using the minimum-snap algorithm~\cite{mellinger2011minimum} to guide the multilift system toward the target, marked by the red square. The solid curves denote agent trajectories, the dashed curves denote reference trajectories, and the dotted curves denote safe-copy trajectories.}
    \label{fig:learning_process_two_stage}
\end{figure}

\begin{figure}[h]
    \centering
    \includegraphics[width=0.8\linewidth]{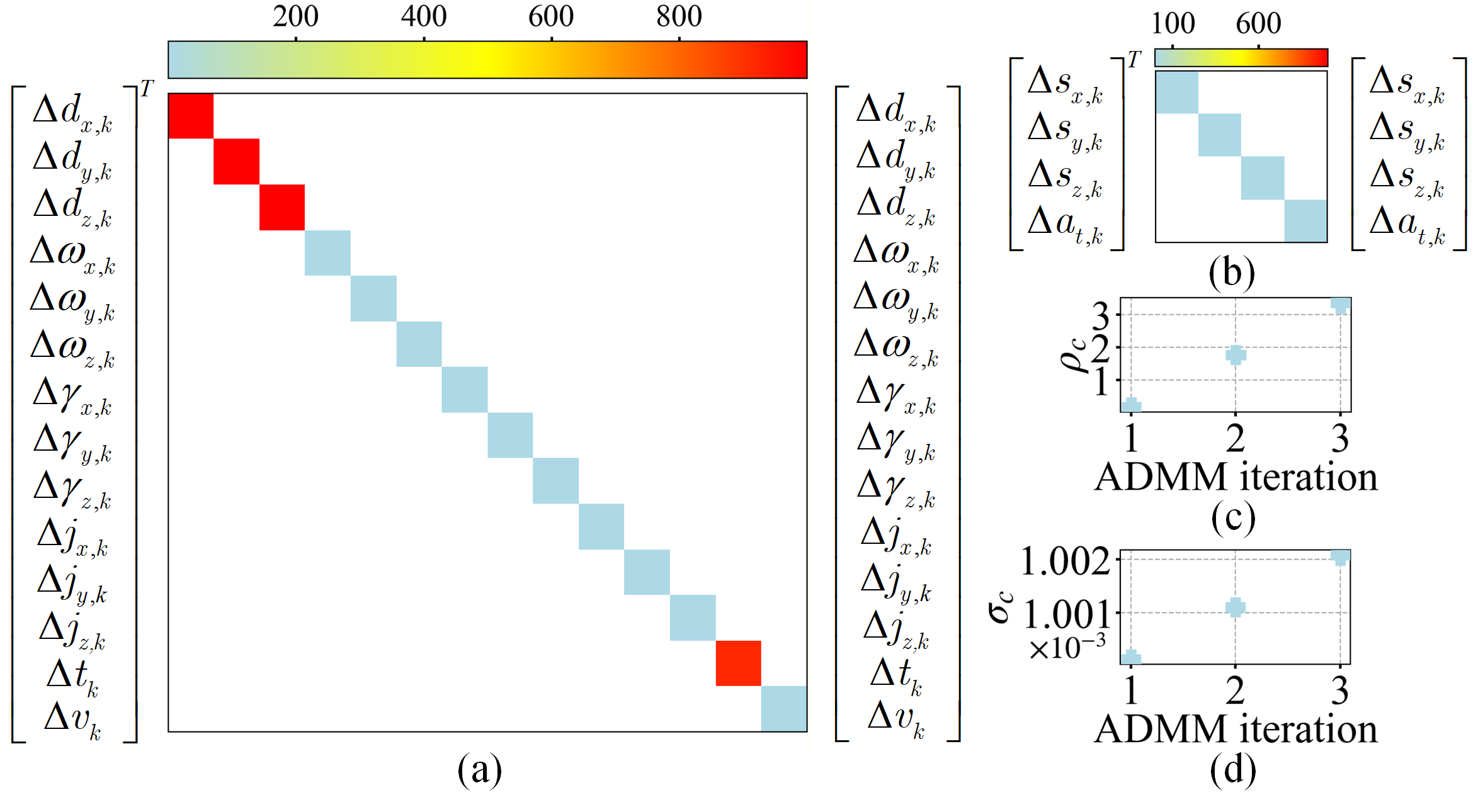}
   \caption{\footnotesize Stage-2 meta-learned parameters for the cables in Task 5. (a) State weights $\bm{Q}_c$. (b) Control weights $\bm{R}_c$. (c) State ADMM penalty parameters $\rho_{c}$ over ADMM iterations. (d) Control ADMM penalty parameters $\sigma_c$ over ADMM iterations. $\Delta$ denotes the error between a variable and its reference.}
    \label{fig:stage_2_weights}
\end{figure}

Fig.~\ref{fig:weight_ratio_rg_stage1} illustrates that the meta-learned parameters adapt to diverse tasks through the neural network modeling in~\eqref{eq:agent_wise network}. The shown ratio is computed by dividing the diagonal tension-allocation weight $\bm{R}_t$ by the ADMM penalty parameter $\rho_{ls}$ at the last ADMM iteration. These parameters appear in the step-wise cost function of Subproblem 2 in Stage 1:
\begin{equation}
    \sum_{i=1}^n\frac{1}{2}\|\bm{t}_{i,k}-\bm{t}^{\mathrm{ref}}_{\mathrm{um}} \|_{\bm{R}_t}^2+\text{ADMM-Residuals}(\rho_{ls},\sigma_{ls})
\end{equation}
The ratio decreases as $\| \bm{r}_g \|$ increases, indicating a smaller $\bm{R}_t$ along with a larger $\rho_{ls}$. A similar ratio change is also observed w.r.t. $\sigma_{ls}$. This variation pattern encourages the cable tensions to effectively deviate from the uniformly distributed tension reference $\bm{t}^{\mathrm{ref}}_{\mathrm{um}}$, enabling non-uniform tension sharing to balance the payload attitude dynamics when its CoM is off-centered. Such off-centered CoM cases (i.e., $\| \bm{r}_g\| >0$) are common in practical transportation.

The task-adaptive parameters enable DiffCoord to achieve better learning performance across tasks than the task-fixed parameters, as reflected by the smaller steady-state meta-losses in Fig.~\ref{fig:meta_loss_in_training}. In addition, the meta-losses associated with the task-adaptive parameters decay more rapidly, indicating more efficient training. Fig.~\ref{fig:learning_process_two_stage} shows the two-stage training process in Task 5 as an example. In all the tasks, the 4-quadrotor multilift system flies through a constrained space formed by two vertical columns. In both stages, the meta-learned parameters balance tracking-error reduction with ADMM residual reduction. Specifically, in Stage 1, the cable references $\{t_{i}^{\mathrm{ref}},\bm{d}_{i}^{\mathrm{ref}}\}_{i=1}^n$ are meta-learned to enable the payload to track its reference trajectory while avoiding collisions with obstacles. Although these cable references satisfy the kinodynamic constraints~\eqref{eq: single agent wrench constraint} and~\eqref{eq:kinematic constraints in cable reference optimization}, they may not satisfy the cable dynamics~\eqref{eq:cable dynamics} and can be zig-zag. In Stage 2, feasible cable trajectories are meta-learned to respect the cable dynamics while closely tracking the Stage-1 cable references, leading to collision-free and dynamically feasible motion.

\begin{figure}[h]
    \centering
    \begin{subfigure}{0.4\columnwidth}
        \centering
        \includegraphics[width=\linewidth]{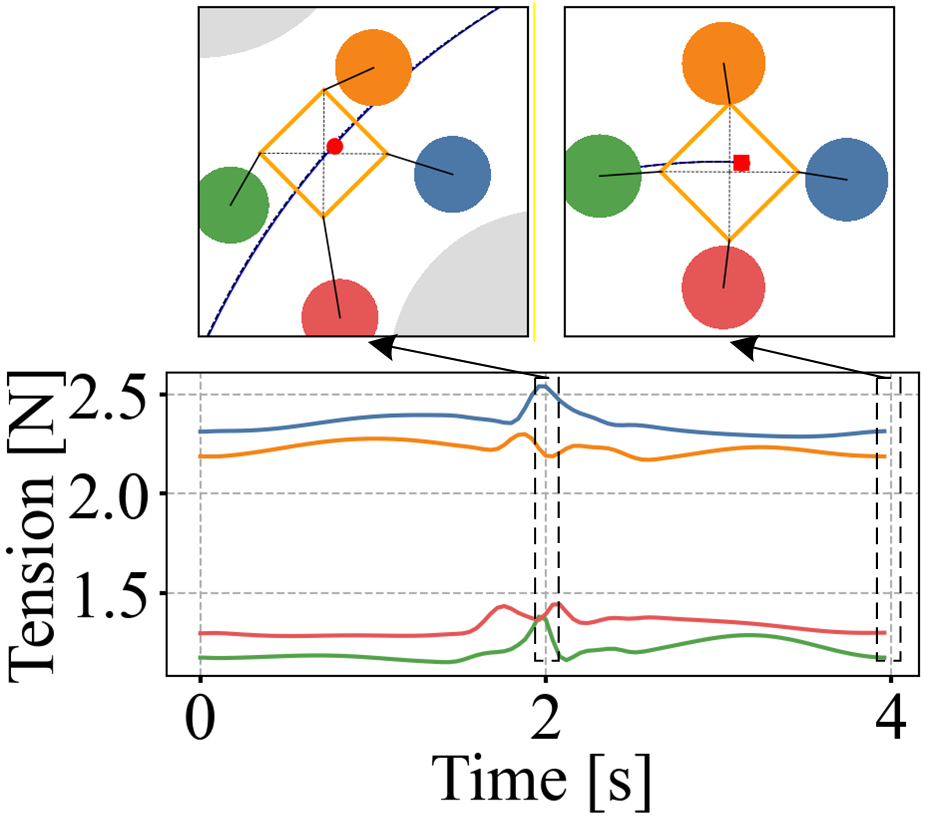}
        \caption{}
        \label{fig:4lift_tension_5}
    \end{subfigure}
    \quad
    \begin{subfigure}{0.4\columnwidth}
        \centering
        \includegraphics[width=\linewidth]{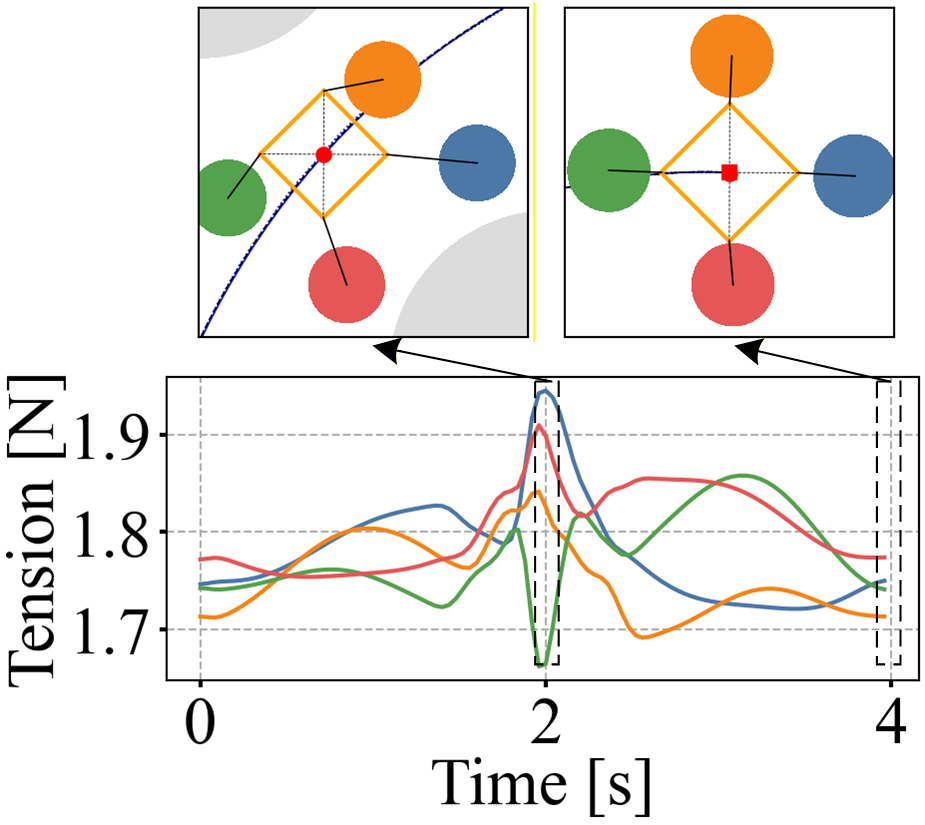}
        \caption{}
        \label{fig:4lift_tension_8}
    \end{subfigure}
   \caption{Comparisons of cable tension allocation and multilift formation shape across different tasks. (a) Cable tension allocation and formation shape during flight in Task 5. (b) Cable tension allocation and formation shape during flight in Task 8. The red point on the payload denotes its CoM, and the insets show the formation shapes at the marked time instants.}
    \label{fig:tension_configuration_multi_tasks}
\end{figure}

Fig.~\ref{fig:stage_2_weights} visualizes the Stage-2 meta-learned parameters for the cables in Task 5. The weights on the errors related to the cable references are significantly larger than those on the other errors. Since explicit references are unavailable for the latter, their references are set to zero to avoid introducing manually specified tracking objectives. The learned weight distribution therefore shows that DiffCoord can identify and emphasize the task-relevant terms. Meanwhile, the meta-learned ADMM penalty parameters exhibit an overall increasing trend over ADMM iterations, progressively enforcing the consensus constraints. These meta-learned parameters balance reference tracking with ADMM residual reduction, confirming effective learning. Fig.~\ref{fig:tension_configuration_multi_tasks} compares cable tension allocation and multilift formation shape during flight between Task 5 and Task 8, which have the largest and smallest $\|\bm{r}_g \|$, respectively. To stabilize the payload attitude during flight, a larger $\|\bm{r}_g \|$ leads to more non-uniform cable tension allocation among $\{t_i\}_{i=1}^n$ and more tilted cable directions $\{\bm{d}_i\}_{i=1}^n$, while the opposite trend is observed for a smaller $\|\bm{r}_g \|$. This demonstrates the effective adaptability of our method to diverse tasks.

\subsection{Real-Flight Experiments: Scalability and Robustness}\label{subsec: real flight experiments}

To show the fourth advantage of DiffCoord, we conduct real-flight experiments using two multilift systems with three and six quadrotors, respectively. The meta-learned networks, trained on the 4-quadrotor multilift system in Subsection~\ref{subsec: meta_learning for multilift}, are directly deployed to the two-stage ADMM-DDP pipelines for both systems without extra tuning.

\begin{figure}[h]
    \centering
    \includegraphics[width=0.8\linewidth]{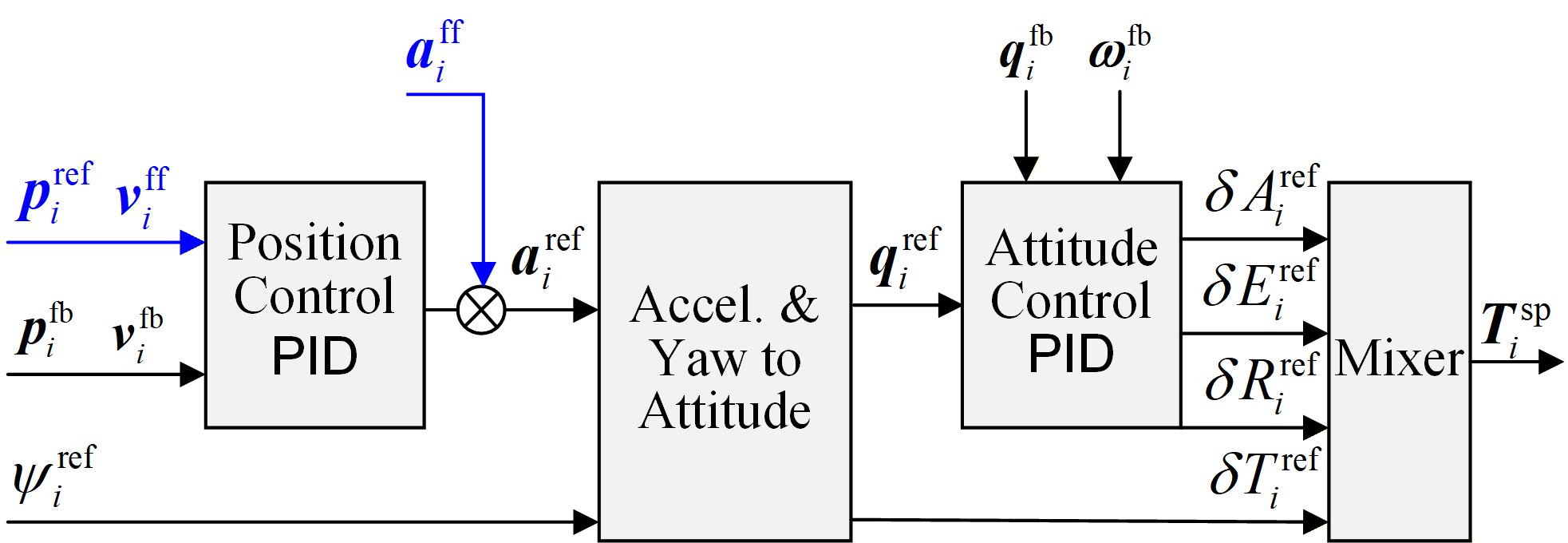}
   \caption{\footnotesize Closed-loop quadrotor PX4 flight control diagram.}
    \label{fig:px4control}
\end{figure}

\begin{figure}[h]
    \centering
    \includegraphics[width=0.8\linewidth]{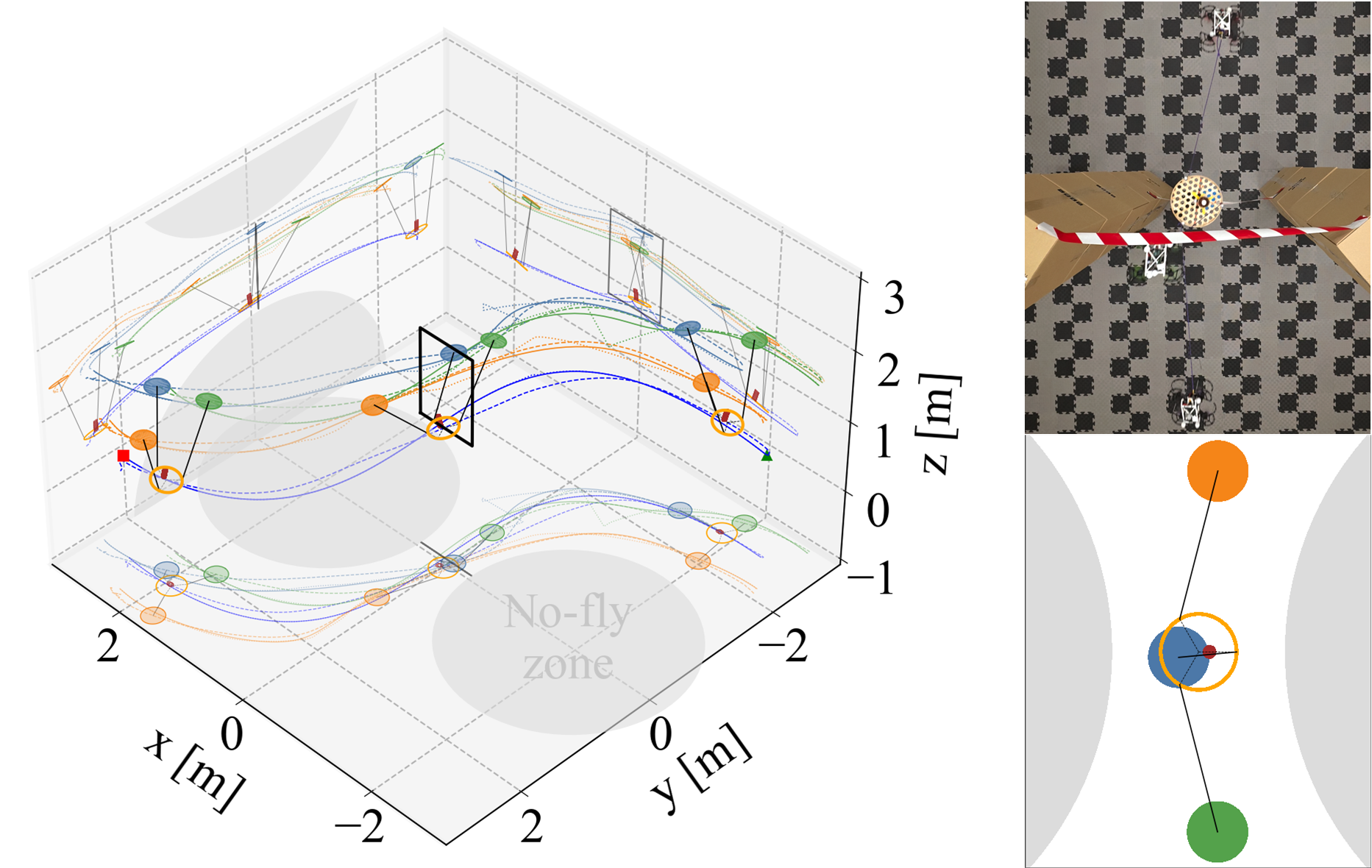}
   \caption{\footnotesize 3D illustration of the 3-quadrotor multilift trajectories in the real-flight experiment with the $x$-biased payload CoM. The right column compares the real-flight formation with the planned formation at the instant of passing through the narrow window. The solid curves denote the planned trajectories, the dotted curves denote the safe-copy trajectories, and the dashed curves denote the real-flight trajectories.}
    \label{fig:hv_3lift_3d_trajectories}
\end{figure}

\begin{figure}[h]
    \centering

    \begin{subfigure}{0.4\columnwidth}
        \centering
        \includegraphics[width=\linewidth]{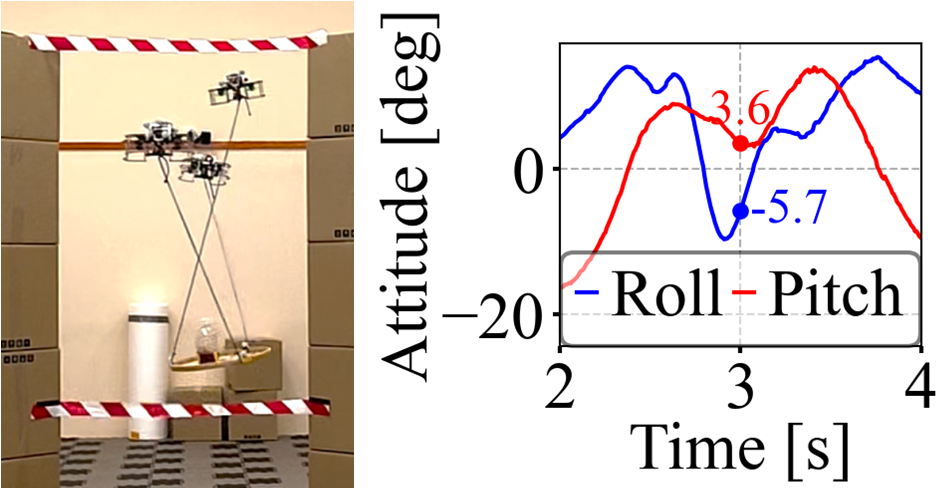}
        \caption{}
        \label{fig:3lift_rpx5_closedloop}
    \end{subfigure}
    \quad
    \begin{subfigure}{0.4\columnwidth}
        \centering
        \includegraphics[width=\linewidth]{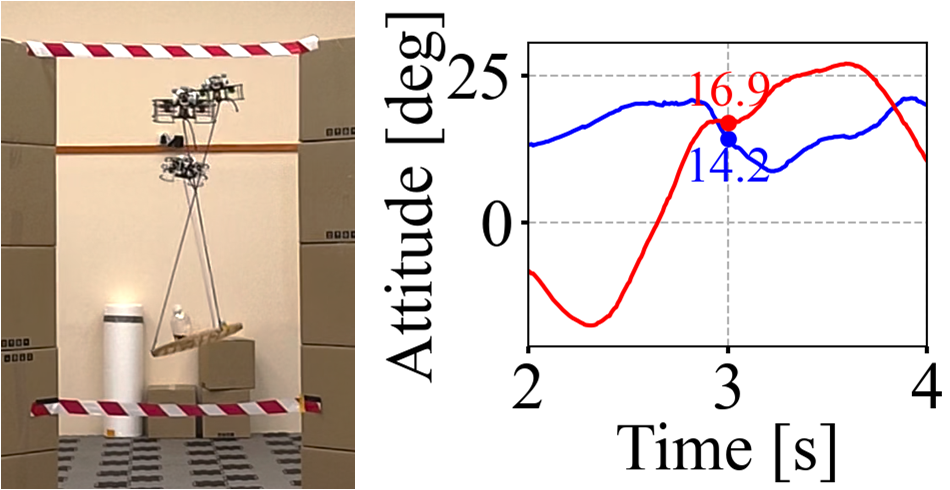}
        \caption{}
        \label{fig:3lift_rpx5_openloop}
    \end{subfigure}

    \vspace{1mm}

    \begin{subfigure}{0.4\columnwidth}
        \centering
        \includegraphics[width=\linewidth]{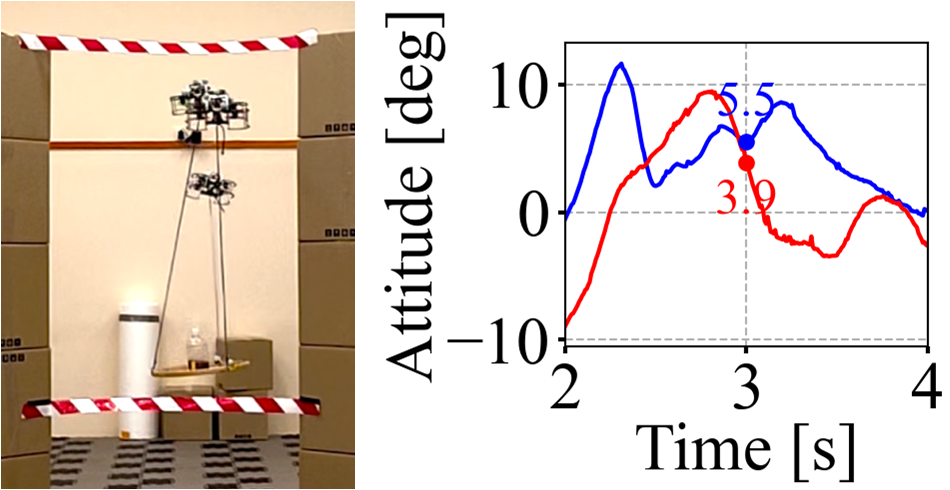}
        \caption{}
        \label{fig:3lift_rpy5_closedloop}
    \end{subfigure}
    \quad
    \begin{subfigure}{0.4\columnwidth}
        \centering
        \includegraphics[width=\linewidth]{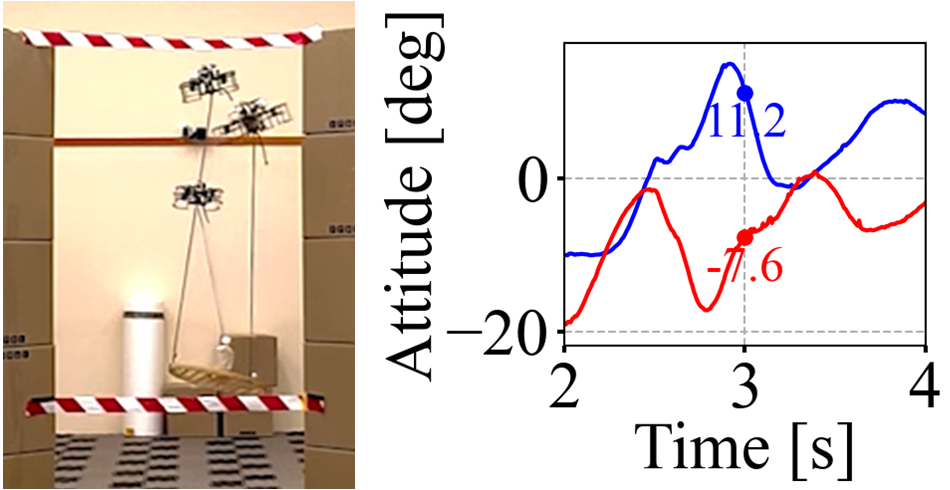}
        \caption{}
        \label{fig:3lift_rpy5_openloop}
    \end{subfigure}

    \caption{Comparison of payload attitude performance under the closed-loop and open-loop control strategies at the instant of passing through the narrow window. (a), (b) Payload attitude with the $x$-biased CoM under the closed-loop and open-loop strategies, respectively. (c), (d) Payload attitude with the $y$-biased CoM under the closed-loop and open-loop strategies, respectively.}

    \label{fig:3lift_payload_control_comparison}
\end{figure}

\begin{figure*}[t]
    \centering
    \includegraphics[width=0.8\linewidth]{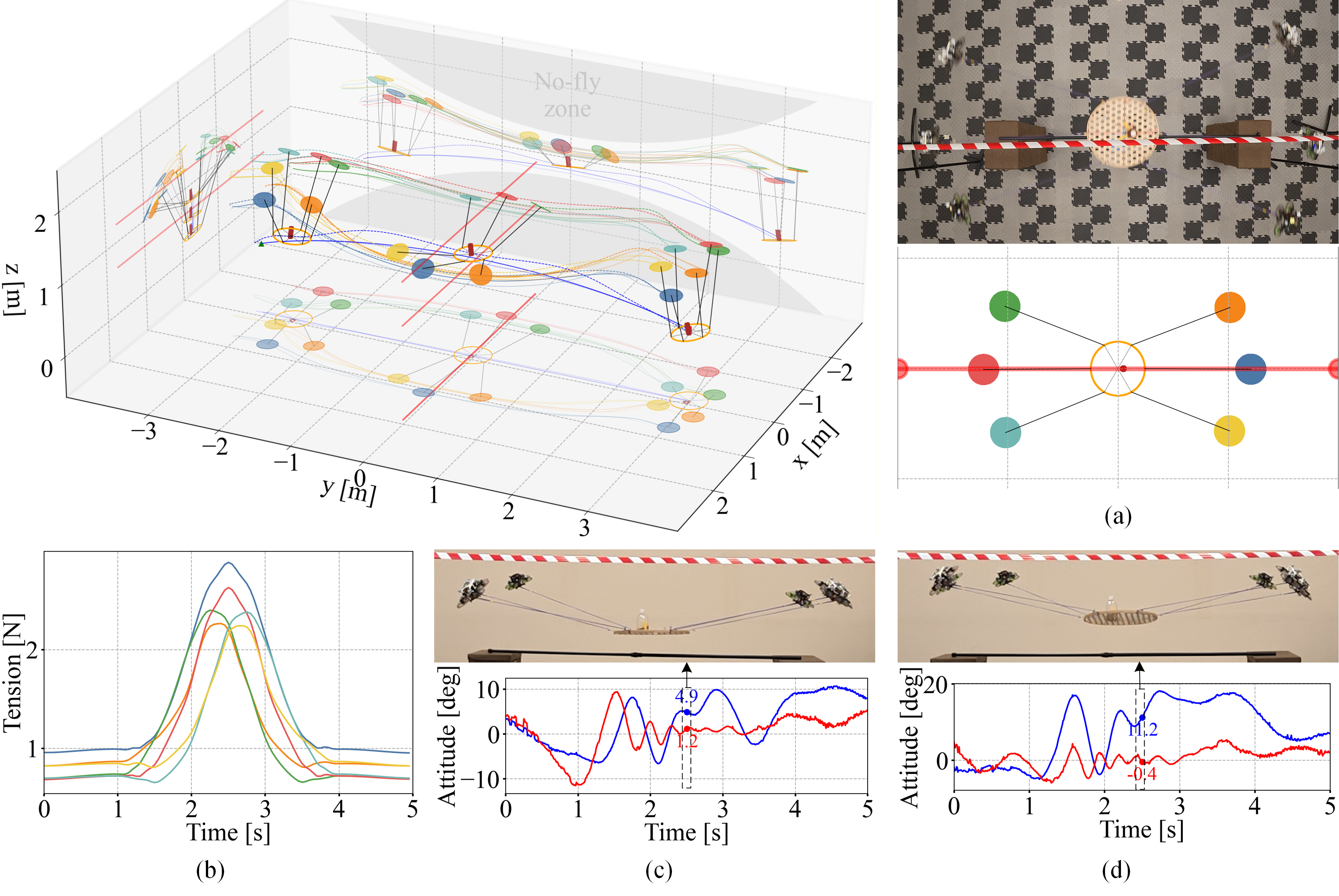}
   \caption{\footnotesize 3D illustration of the 6-quadrotor multilift trajectories in the real-flight experiment with the $x$-biased payload CoM. (a) Comparison of the real-flight formation with the planned formation at the instant of passing through the horizontal narrow gap. (b) Planned cable tensions used in the feedforward signals $\{\bm{a}^{\mathrm{ff}}_i\}_{i=1}^6$. (c), (d) Payload attitude under the closed-loop and open-loop strategies, respectively. }
    \label{fig:6lift_3d_trajectories}
\end{figure*}

\begin{figure}[h]
    \centering
    \begin{subfigure}{0.8\columnwidth}
        \centering
        \includegraphics[width=\linewidth]{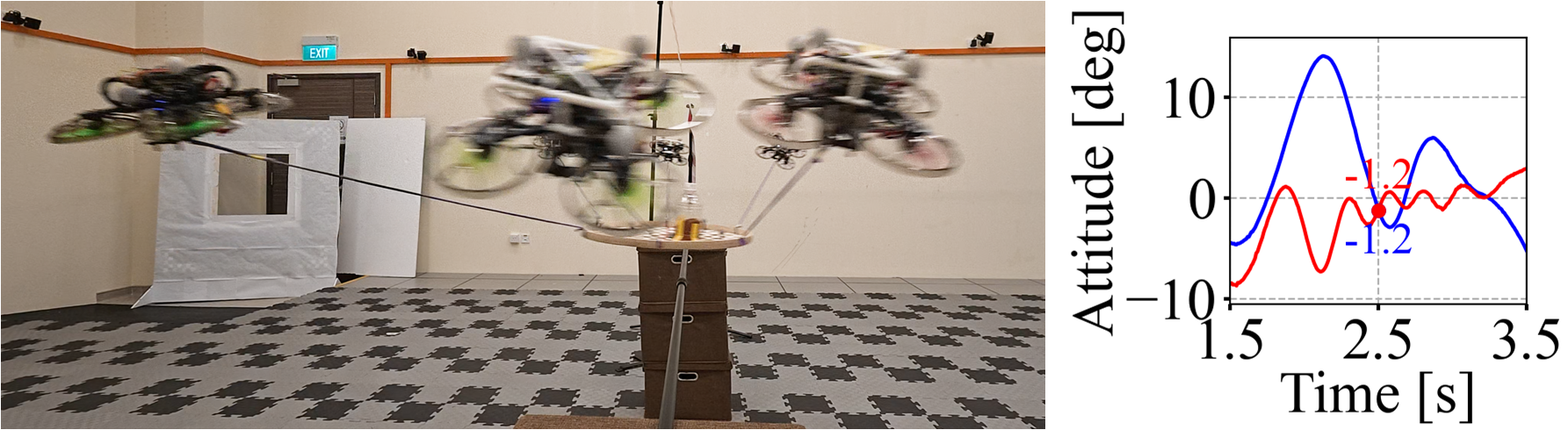}
        \caption{}
        \label{fig:6lift_rpy5_closedloop}
    \end{subfigure}

    \vspace{1mm}

    \begin{subfigure}{0.8\columnwidth}
        \centering
        \includegraphics[width=\linewidth]{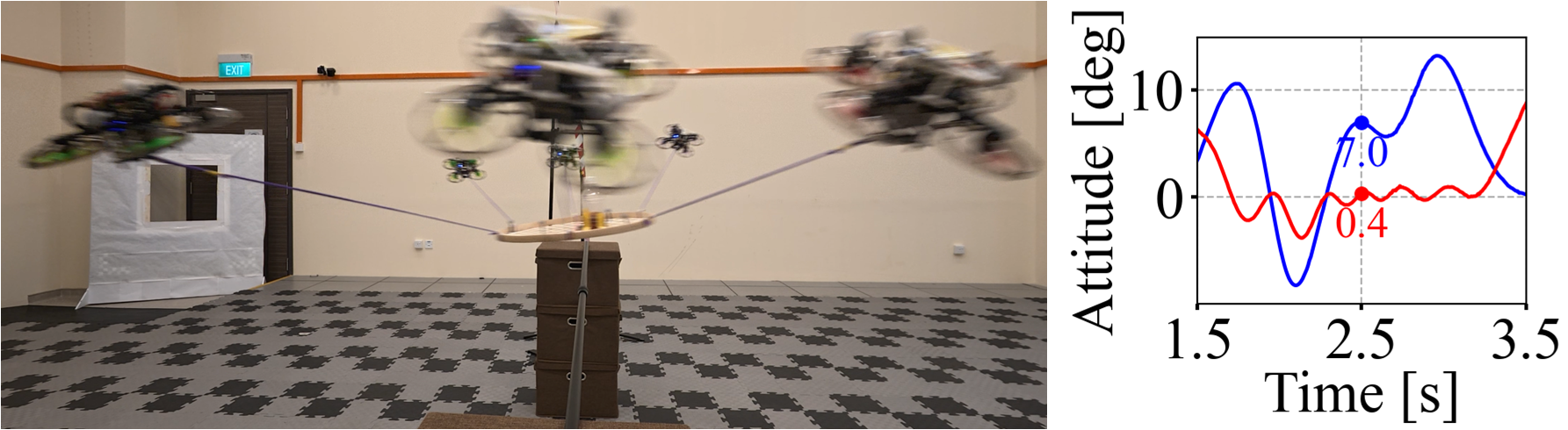}
        \caption{}
        \label{fig:6lift_rpy5_openloop}
    \end{subfigure}

    \caption{Comparison of payload attitude performance with the $y$-biased CoM under the closed-loop and open-loop control strategies at the instant of passing through the horizontally oriented gap. (a) Closed-loop. (b) Open-loop.}
    \label{fig:6lift_payload_rpy5_comparison}
\end{figure}

\begin{table}[h]
\centering
\caption{Comparisons of Payload Attitude RMSEs During Flights.}
\label{tab:comparison_payload_rmse_flight}
\begin{threeparttable}
\begin{tabular}{c !{\vrule width 0.25pt} c !{\vrule width 0.25pt} c !{\vrule width 0.25pt} c !{\vrule width 0.25pt} c}
\toprule
\multirow{2}{*}{RMSE [deg]}
& \multicolumn{2}{c !{\vrule width 0.25pt}}{3-quadrotor system} 
& \multicolumn{2}{c}{6-quadrotor system} \\
& $x$-biased & $y$-biased & $x$-biased & $y$-biased \\
\midrule
Closed-loop & $\bm{14.21}$ & $\bm{11.57}$ & $\bm{7.9}$ & $\bm{9.46}$ \\
Open-loop & $18.26$ & $16.11$ & $11.39$ & $9.78$ \\
\bottomrule
\end{tabular}
\begin{tablenotes}[flushleft]
\footnotesize
\item[] The RMSE for each trajectory is computed from the corresponding payload attitude trajectory in terms of Euler angles, defined by $\text{RMSE}=\sqrt{N^{-1}\sum_{k=1}^{N}(\phi_{l,k}^2+\theta_{l,k}^2+\psi_{l,k}^2)}$, where $\phi_l$, $\theta_l$, and $\psi_l$ denote the payload roll, pitch, and yaw angles, respectively.
\end{tablenotes}
\end{threeparttable}
\end{table}

For each multilift system, we place a 200-g bottle of water 5 cm away from the geometric center of a bamboo tray along either its $x$- or $y$-axis (see Fig.~\ref{fig:multilift}), generating $x$-biased and $y$-biased payload CoMs, respectively. The tray masses are 160 g and 240 g for the 3-quadrotor and 6-quadrotor systems, respectively. Compared with the training tasks in Fig.~\ref{fig:rg_coordinate_and_weight_ratio}, this setup has unseen $\|\bm{r}_g\|$ values, planar coordinates of $\bm{r}_g$, and payload dynamics.

Fig.~\ref{fig:px4control} shows the closed-loop quadrotor control diagram. The planned quadrotor positions, velocities, and accelerations, marked in blue, are sent to the off-the-shelf PX4 controller through its standard setpoint interfaces for trajectory tracking. To further improve robustness to payload model uncertainties, such as liquid sloshing and disturbances caused by quadrotor downwash flow, we implement a closed-loop payload control strategy using the payload DDP feedback gain obtained during planning. Specifically, at each time step $k$, we compute the quadrotor position corrections\footnote{They are bounded by a manually tuned threshold $\|\Delta\bm{p} \|_{\max}$ so that the corrected quadrotor positions still satisfy the collision-free constraints.} $\Delta \bm{p}_k$ in real time based on the cable tension corrections:
\begin{equation}
    \Delta \bm{t}_k
=
\bm{P}^{\dagger}
\begin{bmatrix}
\bm{R}_{l,k}^{\top} & \bm{0}\\
\bm{0} & \bm{I}
\end{bmatrix}
\hat{\bm{K}}_{l,k}^{a_{\mathrm{f}}}
\left(\bm{x}_{l,k}^{\mathrm{fb}}-\bm{x}_{l,k}^{a_{\mathrm{f}}}\right),
\end{equation}
where $\bm{x}_{l,k}^{\mathrm{fb}}$ denotes the payload feedback state. The closed-loop strategy applies these corrections online, while the open-loop strategy tracks the planned quadrotor trajectories without payload feedback.

Fig.~\ref{fig:hv_3lift_3d_trajectories} shows the 3D real-flight trajectories of the 3-quadrotor multilift system passing through a narrow window. The system reconfigures its formation into a line when passing through the window, roughly matching the planned formation and confirming the feasibility of the planned trajectories. Fig.~\ref{fig:3lift_payload_control_comparison} further compares the payload attitude under the closed-loop and open-loop strategies at the instant of passing through the window. The closed-loop strategy keeps the payload attitude closer to the horizontal plane for both the $x$-biased and $y$-biased CoMs by up to $69.5\%$, showing the robustness provided by the DDP feedback gains.

Fig.~\ref{fig:6lift_3d_trajectories} illustrates the 3D real-flight trajectories of the 6-quadrotor multilift system passing through a horizontally oriented narrow gap. This flight scenario is more challenging because it requires stronger cooperation among quadrotors to reconfigure the multilift formation. The system spreads the quadrotors apart to pass through the gap and then gathers them after passing through it, closely following the planned formation change. For the $x$-biased CoM, the closed-loop strategy improves the payload attitude performance over the open-loop strategy by $55\%$ at the instant of passing through the gap. A greater improvement of $75.8\%$ at the same instant is observed for the $y$-biased CoM in Fig.~\ref{fig:6lift_payload_rpy5_comparison}. The advantage of the closed-loop strategy is further supported by comparing the payload Euler angles under the two strategies over the entire flights using the Root Mean Square Errors (RMSEs), as summarized in Table~\ref{tab:comparison_payload_rmse_flight}. The closed-loop strategy can reduce the attitude RMSEs by up to $28.2\%$.

Overall, the experiment results demonstrate:
\begin{enumerate}
\item Generalization of the meta-learned networks to tasks outside the training set, as tested by new $\bm{r}_g$ values;
\item Adaptation to unseen narrow-window and horizontal-gap environments through the inherent constraint-handling capability of the model-based ADMM-DDP planner;
\item Scalability to different team sizes by deploying networks trained with a 4-quadrotor system to 3- and 6-quadrotor systems through the shared agent-wise networks and distributed ADMM-DDP structure;
\item Robustness to model uncertainties through DDP feedback gains, as shown by the improved payload attitude stabilization under the closed-loop control.
\end{enumerate}

\section{Discussion and Conclusion}\label{section:conclusion}
This paper proposed DiffCoord, a unified meta-learning framework for distributed multi-agent trajectory optimization based on the truncated ADMM-DDP pipeline. The central insight is that end-to-end differentiation of this pipeline reveals a two-level structural mirroring. At the dynamic optimization level, trajectory gradients are computed by auxiliary LQR problems that reuse DDP results. At the coordination level, these gradient problems are coupled by an auxiliary ADMM pipeline that inherits the dynamic-static structure, penalty parameters, and iteration budget of the forward pass. This structure-exploiting gradient solver enables efficient joint meta-learning of problem-level and solver-level parameters through shared agent-wise neural networks. Through extensive experiments on multilift systems, we have shown that DiffCoord achieves efficient training under limited ADMM iterations, faster gradient computation than the state-of-the-art full-Jacobian methods, task-adaptive formation reconfiguration, and generalization to different task features, environments, and team sizes.

However, several limitations remain. First, the ADMM iteration budget is not included in the learnable parameters and is instead chosen heuristically by comparing the steady-state meta-losses under different budgets against their computational costs. Once selected, this budget is fixed after training, which may limit generalization to unseen planning environments that are much more complex than the training settings and require more ADMM iterations for satisfactory trajectories. Second, the efficiency gain of the proposed trajectory gradient solver relies on reusing DDP quantities from the forward pass, such as Riccati recursions and feedback gains. Although the gradient solver can still be used when the trajectory subproblem is solved by another numerical solver, these quantities may then need to be recomputed, reducing the efficiency advantage.

Future work will investigate adaptive ADMM iteration budgets to further improve the generalization of DiffCoord across unseen planning scenarios. We will also extend DiffCoord to distributed multi-agent trajectory optimization in dynamic environments.


% or
\appendix % for no appendix heading

\subsection{Proof of Lemma~\ref{lm:DBP and PDP equivalence}}\label{appendix:proof of lemma1}

We begin by establishing the equivalence between the DBP conditions and the differential PMP conditions. Expanding the coefficient matrices on the right-hand side (R.H.S.) of~\eqref{eq:differential condition on x} yields the following expression:
    \begin{equation}
        \begin{split}
            \frac{\mathrm{d}\hat{V}_{i,k}^{x,a}}{\mathrm{d}\bm{\theta}}&=\Bigl[\hat{\ell}_{i,k}^{xx,a}+\hat{V}_{i,k+1}^{x,a}\cdot f_{i,k}^{xx,a} +(f_{i,k}^{x,a})^{\top}\hat{V}_{i,k+1}^{xx,a}f_{i,k}^{x,a} \Bigl]\frac{\mathrm{d} {\bm{x}_{k}^{  a}}}{\mathrm{d} \bm{\theta}}\\
            & + \Bigl[\hat{\ell}_{i,k}^{xu,a}+\hat{V}_{i,k+1}^{x,a}\cdot f_{i,k}^{xu,a} +(f_{i,k}^{x,a})^{\top}\hat{V}_{i,k+1}^{xx,a}f_{i,k}^{u,a} \Bigl]\frac{\mathrm{d} {\bm{u}_{k}^{a}}}{\mathrm{d} \bm{\theta}}\\
            & + \Bigl[\hat{\ell}_{i,k}^{x\theta,a}+\hat{V}_{i,k+1}^{x,a}\cdot f_{i,k}^{x\theta,a}+(f_{i,k}^{x,a})^{\top}\hat{V}_{i,k+1}^{xx,a}f_{i,k}^{\theta,a} \\
            & +f_{i,k}^{x,a}\hat{V}_{i,k+1}^{x\theta,a}\Bigl].
        \end{split}
        \label{eq:proof lemma1 expand RHS of 36}
    \end{equation}
By comparing the first two terms in each square bracket with the Hamiltonian $\hat{\mathcal{H}}_{i,k}^a$, and using the key relation $\hat{\bm{\lambda}}_{i,k}^{a}=\hat{V}^{x,a}_{i,k}$, we observe that these terms correspond exactly to the Hessians $\hat{H}^{xx,a}_{i,k}$, $\hat{H}^{xu,a}_{i,k}$, and $\hat{H}^{x\theta,a}_{i,k}$. With \eqref{eq:differential dynamics} applied, \eqref{eq:proof lemma1 expand RHS of 36} becomes:
    \begin{equation}
        \begin{split}
            \frac{\mathrm{d}\hat{V}^{x,a}_{i,k}}{\mathrm{d}\bm{\theta}}&=\hat{H}^{xx,a}_{i,k}\frac{\mathrm{d} {\bm{x}_{i,k}^{a}}}{\mathrm{d} \bm{\theta}}+\hat{H}^{xu,a}_{i,k}\frac{\mathrm{d} {\bm{u}_{i,k}^{a}}}{\mathrm{d} \bm{\theta}}+\hat{H}^{x\theta,a}_{i,k}\\
            &\quad + (f^{x,a}_{i,k})^{\top}\Bigl[ \hat{V}^{xx,a}_{i,k+1}\frac{\mathrm{d}\bm{x}_{i,k+1}^a}{\mathrm{d}\bm{\theta}}+\hat{V}^{x\theta,a}_{k+1}\Bigl].
        \end{split}
        \label{eq:proof lemma1 expand RHS of 36 after submission}
    \end{equation}
The terms within the square bracket in~\eqref{eq:proof lemma1 expand RHS of 36 after submission} satisfy
    \begin{equation}
        \frac{\mathrm{d}\hat{V}^{x,a}_{i,k+1}}{\mathrm{d}\bm{\theta}}=\hat{V}^{xx,a}_{i,k+1}\frac{\mathrm{d}\bm{x}_{i,k+1}^a}{\mathrm{d}\bm{\theta}}+\hat{V}^{x\theta,a}_{i,k+1}.
        \label{eq:proof lemma1 partial derivative of Vx}
    \end{equation}
By substituting \eqref{eq:proof lemma1 partial derivative of Vx} and using $\hat{\bm{\lambda}}_{i,k}^{a}=\hat{V}^{x,a}_{i,k}$ again, the resulting~\eqref{eq:proof lemma1 expand RHS of 36 after submission} matches exactly \eqref{eq:differential PMP on x lemma1}. Applying the same reasoning to \eqref{eq:differential condition on u} yields its equivalence to \eqref{eq:differential PMP on u lemma1}. Next, we show that the LQR system~\eqref{eq:same LQR system in lemma1} can also be constructed from the DBP conditions. At the optimal solutions, the Bellman equation of~\eqref{eq:same LQR system in lemma1} takes the following form:
    \begin{equation}
    \begin{split}
        \bar{\hat{V}}^a_{i,k}(\bm {X}^a_{i,k})&=\min_{\bm {U}^a_{i,k}}\bar{\hat{Q}}^a_{i,k}(\bm{X}^a_{i,k},\bm{U}^a_{i,k})\\
        &= \operatorname{tr} \left( \frac{1}{2}\begin{bmatrix}
                \bm{X}_{i,k}^a\\
                \bm{U}_{i,k}^a
            \end{bmatrix}^{\top}
            \begin{bmatrix}
                \hat{H}^{xx,a}_{i,k},\hat{H}^{xu,a}_{i,k}\\
                \hat{H}^{ux,a}_{i,k},\hat{H}^{uu,a}_{i,k}
            \end{bmatrix}
            \begin{bmatrix}
                \bm{X}_{i,k}^a\\
                \bm{U}_{i,k}^a
            \end{bmatrix} \right.\\
            &\qquad \quad  \left.+\begin{bmatrix}
                \hat{H}^{x\theta,a}_{i,k}\\
                \hat{H}^{u\theta,a}_{i,k}
            \end{bmatrix}^{\top}\begin{bmatrix}
                \bm{X}_{i,k}^a\\
                \bm{U}_{i,k}^a
            \end{bmatrix} \right)+\bar{\hat{V}}^a_{i,k+1}(\bm {X}^a_{i,k+1}).
    \end{split}
        \label{eq:Bellman's principle of matrix_valued LQR}
    \end{equation}
    Inspired by the terminal cost, we define the value function at step $k$ as
    \begin{equation}
    \begin{split}
        \bar{\hat{V}}_{i,k}^a(\bm{X}_{i,k}^{a})&=\operatorname{tr}\left(\frac{1}{2}\left(\bm{X}_{i,k}^a\right)^{\top}\hat{V}^{xx,a}_{i,k}\bm{X}_{i,k}^{a}\right.\\
        &\quad \quad \ \left. +\left(\hat{V}^{x\theta,a}_{i,k}\right)^{\top}\bm{X}_{i,k}^{a} +\hat{V}^{\theta\theta,a}_{i,k}\right),
    \end{split}
        \label{eq:expression of the value function lemma1}
    \end{equation}
    where $\hat{V}^{\theta\theta,a}_{i,k}$ is the accumulated, state-independent matrix and satisfies $\hat{V}^{\theta\theta,a}_{i,N}=\bm{0}$. Differentiating both sides of~\eqref{eq:Bellman's principle of matrix_valued LQR} w.r.t. $\bm{X}_{i,k}^a$ and $\bm{U}_{i,k}^a$, respectively, using matrix calculus and~\eqref{eq:expression of the value function lemma1}, we obtain the following first-order optimality conditions:
    \begin{subequations}
        \begin{align}
            \begin{split}
                \frac{\mathrm{d}\bar{\hat{V}}_{i,k}^a}{\mathrm{d}\bm{X}_{i,k}^a}&=\hat{H}^{xx,a}_{i,k}\bm{X}_{i,k}^{a}+\hat{H}^{xu,a}_{i,k}\bm{U}_{i,k}^{a}+\hat{H}^{x\theta,a}_{i,k}+ (f^{x,a}_{i,k})^{\top}\\
                &\quad \times \Bigl[ \hat{V}^{xx,a}_{k+1}\left(f^{x,a}_{i,k}\bm{X}_{i,k}^{a}+f^{u,a}_{i,k}\bm{U}_{i,k}^{a} + f^{\theta,a}_{i,k}\right) +\hat{V}^{x\theta,a}_{i,k+1}\Bigl],
            \end{split}\label{eq:partial derivative of Bellman on X lemma1}\\
            \begin{split}
                \bm{0}&=\hat{H}^{ux,a}_{i,k}\bm{X}_{i,k}^{a}+\hat{H}^{uu,a}_{i,k}\bm{U}_{i,k}^{a}+\hat{H}^{u\theta,a}_{i,k}+ (f^{u,a}_{i,k})^{\top}\\
                &\quad \times \Bigl[ \hat{V}^{xx,a}_{k+1}\left(f^{x,a}_{i,k}\bm{X}_{i,k}^{a}+f^{u,a}_{i,k}\bm{U}_{i,k}^{a} + f^{\theta,a}_{i,k}\right) +\hat{V}^{x\theta,a}_{i,k+1}\Bigl].
            \end{split}\label{eq:partial derivative of Bellman on U lemma1}
            % \\
            % \bm{X}_{k+1}^{\ast}&=f_{\bm{x},k}^{\ast}\bm{X}_{k}^{\ast}+f_{\bm{u},k}^{\ast}\bm{U}_{k}^{\ast}+f_{\bm{\theta},k}^{\ast}\label{eq:auxiliary dynamics}.
        \end{align}
        \label{eq:first_order optimality conditions of matrix_valued LQR}%
    \end{subequations}
Compared with~\eqref{eq:proof lemma1 expand RHS of 36 after submission}, \eqref{eq:partial derivative of Bellman on X lemma1} matches \eqref{eq:differential condition on x}. The same reasoning applies to \eqref{eq:partial derivative of Bellman on U lemma1}, establishing its equivalence to \eqref{eq:differential condition on u}. We conclude that Bellman's principle of~\eqref{eq:same LQR system in lemma1} coincides with the DBP conditions~\eqref{eq:differential optimality conditions} and thus~\eqref{eq:equivalence between the optimal solutions and the gradients} holds.

\subsection{Proof of Lemma~\ref{lm:DDP gradient solver}}\label{appendix:proof of lemma2}
We prove the analytical gradients in Lemma~\ref{lm:DDP gradient solver} by solving the auxiliary matrix-valued LQR problem~\eqref{eq:same LQR system in lemma1} using Bellman's principle. From~\eqref{eq:partial derivative of Bellman on U lemma1}, we can solve for the matrix-valued optimal control:
\begin{equation}
    \bm{U}_{i,k}^a =-(\hat{Q}^{uu,a}_{i,k})^{-1}\hat{Q}^{ux,a}_{i,k}\bm{X}_{i,k}^a  -(\hat{Q}^{uu,a}_{i,k})^{-1}\hat{Q}^{u\theta,a}_{i,k},
    \label{eq:optimal matrix-valued control}
\end{equation}
where $\hat{Q}^{uu,a}_{i,k}=\hat{H}^{uu,a}_{i,k}+(f^{u,a}_{i,k})^{\top}\hat{V}^{xx,a}_{i,k+1}f^{u,a}_{i,k}$, $\hat{Q}^{ux,a}_{i,k}=\hat{H}^{ux,a}_{i,k}+(f^{u,a}_{i,k})^{\top}\hat{V}^{xx,a}_{i,k+1}f^{x,a}_{i,k}$, and $\hat{Q}^{u\theta,a}_{i,k}=\hat{H}^{u\theta,a}_{i,k}+(f^{u,a}_{i,k})^{\top}\hat{V}^{x\theta,a}_{i,k+1}+(f^{u,a}_{i,k})^{\top}\hat{V}^{xx,a}_{i,k+1}f^{\theta,a}_{i,k}$.
Plugging $\bm{U}_{i,k}^{a}$ into~\eqref{eq:expression of the value function lemma1} and comparing the resulting $\bar{Q}_k$ with \eqref{eq:expression of the value function lemma1} yields the following two Riccati recursions:
\begin{subequations}
    \begin{align}
        \begin{split}
            \hat{V}^{xx,a}_{i,k}&=\hat{Q}^{xx,a}_{i,k}-(\hat{Q}^{ux,a}_{i,k})^{\top}
            (\hat{Q}^{uu,a}_{i,k})^{-1}\hat{Q}^{ux,a}_{i,k},
        \end{split}\label{eq:Vxx Riccati matrix-valued LQR}\\
        \begin{split}
            \hat{V}^{x\theta,a}_{i,k}&=\hat{Q}^{x\theta,a}_{i,k}-(\hat{Q}^{ux,a}_{i,k})^{\top}(\hat{Q}^{uu,a}_{i,k})^{-1}\hat{Q}^{u\theta,a}_{i,k},
        \end{split}\label{eq:Vxtheta Riccati matrix-valued LQR}
    \end{align}
    \label{eq:Riccati recursions matrix-valued LQR}%
\end{subequations}
where $\hat{Q}^{xx,a}_{i,k}=\hat{H}^{xx,a}_{i,k}+(f^{x,a}_{i,k})^{\top}\hat{V}^{xx,a}_{i,k+1}f^{x,a}_{i,k}$, $\hat{Q}^{x\theta,a}_{i,k}=\hat{H}^{x\theta,a}_{i,k}+(f^{x,a}_{i,k})^{\top}\hat{V}^{x\theta,a}_{i,k+1}+(f^{x,a}_{i,k})^{\top}\hat{V}^{xx,a}_{i,k+1}f^{\theta,a}_{i,k}$, $\hat{V}^{xx,a}_{i,N}=\hat{\ell}^{xx,a}_{i,N}$ and $\hat{V}^{x\theta,a}_{i,N}=\hat{\ell}^{x\theta,a}_{i,N}$ are the terminal conditions. We observe that the two recursions \eqref{eq:Vxx Riccati matrix-valued LQR} and \eqref{eq:V_xx} are identical. This justifies the reuse of $\{\hat{V}^{xx,a}_{i,k+1}\}^{N-1}_{k=0}$ . The matrix-valued optimal control law~\eqref{eq:optimal matrix-valued control} comprises a feedback gain and a feedforward gain, and has a similar structure to the DDP control law given in~\eqref{eq:DDP control law}. In particular, the feedback gain can stabilize the gradient computation and coincides with the DDP gain $\hat{\bm{K}}_{i,k}^a = - (\hat{Q}^{uu,a}_{i,k})^{-1} (\hat{Q}^{xu,a}_{i,k})^{\top}$. This confirms the reuse of $\{\hat{\bm{K}}^a_{i,k}\}^{N-1}_{k=0}$ and completes the proof.

Note that it is unnecessary to compute the Riccati recursion of $\hat{V}^{\theta\theta,a}_{i,k}$, as it plays no role in computing $\bm{U}_{i,k}^{a}$. This, together with the aforementioned reuse, leads to significantly higher computational efficiency compared to the DDP-based gradient solver in~\cite{11214470}, which additionally requires computing the trajectory of $V^{\theta\theta}_{k}$ and recomputing the trajectories of $\bm {K}_{k}$ and $V^{xx}_k$ (following the single-agent notation in~\cite{11214470}).

\subsection{Proof of Theorem~\ref{tm: Lipschitz bound}}\label{appendix:proof of theorem1}
We prove the Lipschitz boundedness certified in Theorem~\ref{tm: Lipschitz bound} by observing that the stacked optimal solution $\bm{s}^a$ of Auxiliary Subproblems 2 and 3 admits the following affine, iteration-varying recursion:
\begin{equation}
    \bm{s}^a=\mathcal{T}(\bm{\tau}^a_{\mathrm{f}})\bm{s}^{a-1}+c(\bm{\tau}^a_{\mathrm{f}}),
    \label{eq: tm proof affine iteration varying s}
\end{equation}
which is obtained by substituting the optimal solution of Auxiliary Subproblem 1 into Auxiliary Subproblems 2 and 3. Here, $\mathcal{T}(\bm{\tau}^a_{\mathrm{f}})$ and $c(\bm{\tau}^a_{\mathrm{f}})$ are constructed using the terms from the three auxiliary subsystems, such as the coefficient matrices in the cost functions and the Jacobians of the system dynamics model. They are locally Lipschitz in $\bm{\tau}^a_{\mathrm{f}}$ near $\bm{\tau}^*_{\mathrm{f}}$:
\begin{equation}
\begin{aligned}
    \| \mathcal{T}(\bm{\tau}^a_{\mathrm{f}}) -\mathcal{T}(\bm{\tau}^*_{\mathrm{f}}) \| &\leq L_{T}\| \bm{\tau}^a_{\mathrm{f}} - \bm{\tau}^*_{\mathrm{f}} \|,\\
    \| c(\bm{\tau}^a_{\mathrm{f}}) -c(\bm{\tau}^*_{\mathrm{f}}) \| &\leq L_{c}\| \bm{\tau}^a_{\mathrm{f}} - \bm{\tau}^*_{\mathrm{f}} \|.
\end{aligned}
\label{eq: tm proof lipschitz bounded coefficients}
\end{equation}
By the local boundedness, there exists a bounded nonnegative constant~\footnote{For iteration-invariant convex quadratic cases, the corresponding ADMM updates can converge as $a \to \infty$, thereby yielding $\Gamma \leq 1$~\cite{6892987}. In our iteration-varying setting, however, we only require $\Gamma$ to be bounded, even when the auxiliary ADMM-LQR system is made strictly convex under the requirements detailed in Subsection~\ref{subsec theoretical guarantee}.} $0\leq \Gamma < \infty$ such that $\| \mathcal{T}(\bm{\tau}^a_{\mathrm{f}})\|\leq \Gamma$ for all $\bm{\tau}^a_{\mathrm{f}}$ near $\bm{\tau}^*_{\mathrm{f}}$. Let $\bm{e}^a\coloneqq \bm{s}^a - \bm{s}^*$ denote the truncation-induced error, where $\bm{s}^*$ is a fixed-point solution of $\bm{s}^*=\mathcal{T}(\bm{\tau}^*_{\mathrm{f}})\bm{s}^{*}+c(\bm{\tau}^*_{\mathrm{f}})$. Then, using~\eqref{eq: tm proof lipschitz bounded coefficients}, we have the following recursion on $\bm{e}^a$:
\begin{equation}
    \|\bm{e}^a \|\leq \Gamma \| \bm{e}^{a-1}\| + (L_T\| \bm{s}^*\| + L_c)\|\bm{\tau}^a_{\mathrm{f}} -\bm{\tau}^*_{\mathrm{f}}  \|.
    \label{eq: tm error recursion}
\end{equation}
Unrolling it over $a_{\mathrm{f}}$ steps yields:
\begin{equation}
    \|\bm{e}^{a_{\mathrm{f}}} \|\le \Gamma^{a_{\mathrm{f}}}\| \bm{e}^0 \| + \sum_{j=0}^{a_{\mathrm{f}}-1}\Gamma^{j}(L_{T}\|\bm{s}^*\|+L_{c})\delta{\tau}_{\infty},
    \label{eq: tm unrolliing errors}
\end{equation}
which completes the proof of~\eqref{eq: theorem error bound}. 

The stacked optimal solution $\{ \bm{X}^a_k, \bm{U}^a_k\}$ of Auxiliary Subproblem 1 across $n$ agents also admits the following iteration-varying affine form in terms of $\bm{s}^{a-1}$:
\begin{equation}
    \bm{X}^a_k = \bm{A}^a_k\bm{s}^{a-1}+\bm{\beta}^a_k, \ \bm{U}^a_k=\bm{B}^a_k\bm{s}^{a-1}+\bm{\phi}^a_k.
    \label{eq: tm affine form of auxiliary subsystem1}
\end{equation}
The agent-wise component of~\eqref{eq: tm affine form of auxiliary subsystem1} is obtained by unrolling the recursion~\eqref{eq:gradient dynamics lemma2} from step $0$ to $k$. Correspondingly, the unrolling generates the coefficients $\bm{A}^a_k$, $\bm{\beta}^a_k$, $\bm{B}^a_k$, and $\bm{\phi}^a_k$, which depend on $\bm{\tau}^a_{\mathrm{f}}$ through the cost weights and system Jacobians and are locally Lipschitz:
\begin{equation}
    \begin{aligned}
        \|\bm{A}^a_k - \bm{A}^*_k \|&\leq L_A\|\bm{\tau}^a_{\mathrm{f}} - \bm{\tau}^*_{\mathrm{f}} \|,\ \|\bm{\beta}^a_k - \bm{\beta}^*_k \|\leq L_{\beta}\|\bm{\tau}^a_{\mathrm{f}} - \bm{\tau}^*_{\mathrm{f}} \|,\\
        \|\bm{B}^a_k - \bm{B}^*_k \|&\leq L_B\|\bm{\tau}^a_{\mathrm{f}} - \bm{\tau}^*_{\mathrm{f}} \|,\ \|\bm{\phi}^a_k - \bm{\phi}^*_k \|\leq L_{\phi}\|\bm{\tau}^a_{\mathrm{f}} - \bm{\tau}^*_{\mathrm{f}} \|,
    \end{aligned}
    \label{eq: tm bounded coefficients in unrolled subsystem1}
\end{equation}
where $L_A, L_{\beta}, L_B, L_{\phi}>0$. Combining~\eqref{eq: tm affine form of auxiliary subsystem1} with~\eqref{eq: tm bounded coefficients in unrolled subsystem1} directly gives~\eqref{eq: theorem gradient error}. This completes the proof.

\bibliographystyle{IEEEtran}
\bibliography{reference}

\end{document}